\documentclass{article}

\PassOptionsToPackage{numbers, compress}{natbib}
\usepackage[preprint]{neurips_2026}

\usepackage[utf8]{inputenc} 
\usepackage[T1]{fontenc}    
\usepackage{hyperref}       
\usepackage{url}            
\usepackage{booktabs}       
\usepackage{amsfonts}       
\usepackage{nicefrac}       
\usepackage{microtype}      
\usepackage{xcolor}         
\usepackage{subfigure}
\usepackage{graphicx}
\usepackage{mwe} 
\usepackage{wrapfig}
\usepackage{threeparttable}
\usepackage{multirow}
\usepackage{amsmath}
\usepackage{algorithm}
\usepackage{algpseudocode}
\usepackage{amssymb}

\usepackage{siunitx} 
\usepackage{tabularx}
\usepackage{caption}
\usepackage{subcaption}

\title{Set Prediction for Next-Day Active Fire Forecasting}

\author{%
\textbf{Yuchen Bai}$^{1}$ \quad
\textbf{Georgios Athanasiou}$^{2}$ \quad
\textbf{Xin Yu}$^{3}$ \quad
\textbf{Diogenis Antonopoulos}$^{2}$\\
\textbf{Ioannis Papoutsis}$^{2,4}$ \quad
\textbf{Stijn Hantson}$^{1,5}$ \quad
\textbf{Nuno Carvalhais}$^{1,6,7}$\thanks{Corresponding author.}\\
$^1$Max Planck Institute for Biogeochemistry, Jena, Germany\\
$^2$Orion Lab, National Technical University of Athens\\
$^3$University of Utah, Salt Lake City, USA\\
$^4$National Observatory of Athens\\
$^5$Earth System Science Program, School of Sciences and Engineering,\\
Universidad del Rosario, Bogotá, Colombia\\
$^6$Departamento de Ciências e Engenharia do Ambiente,\\
Faculdade de Ciências e Tecnologia, Universidade Nova Lisboa,\\
Costa da Caparica, Portugal\\
$^7$ELLIS Unit Jena, Jena, Germany\\
\texttt{\{ybai,shantson,ncarval\}@bgc-jena.mpg.de}\\
\texttt{\{georgios\_athanasiou,ipapoutsis\}@mail.ntua.gr}\\
\texttt{diogenis.antonopoulos@gmail.com}\\
\texttt{xin.yu@utah.edu}
}

\begin{document}

\maketitle

\begin{abstract}
Accurate next-day active fire forecasts can support early warning, disaster response, forest risk assessment, and downstream estimation of fire-related carbon emissions. Existing machine learning approaches to wildfire forecasting typically predict wildfire danger or fire probability on kilometre-scale daily grids, which is useful for regional warning but does not directly represent localized fire events. We propose Wildfire Ignition Set Predictor (WISP), a query-based model that reformulates next-day active fire forecasting as point-set prediction. From 48 hours of covariates including meteorology, satellite vegetation products, static land, and fire history, WISP predicts a fixed-size ranked set of future active fire cluster centres on a \qty{375}{\meter} grid across globally distributed regions. The model is trained end-to-end with Hungarian matching; to address the conflicting roles of the classification score in assignment, ranking, and query activation, we use asymmetric classification–localization weighting in matching and loss. We further construct a globally distributed, hourly, multi-source benchmark for this task. On a held-out test set spanning fire regions worldwide, the best WISP variant achieves 38.2\% average precision (AP) for ranked fire-centre detections, covers 53.4\% of fire cluster mass weighted by fire radiative power (FRP), and localizes 54.1\% of observed clusters within \qty{5}{km}. These results establish sparse set prediction as a viable formulation for high-resolution wildfire forecasting and provide a benchmark for future work in this regime.
\end{abstract}



\section{Introduction}
\label{Sec1:intro}

Wildfires displace communities, devastate ecosystems, and impose growing economic and human costs each year, with the frequency of the most extreme events more than doubling globally between 2003 and 2023~\cite{cunningham2024increasing}. As fire weather seasons lengthen and extreme wildfire activity becomes more frequent, the burden on emergency response shifts upstream: anticipating where fire is likely to emerge over the next day at fine spatial scale --- not merely whether risk is elevated in a region --- is therefore an operationally important prediction problem in Earth-system science~\cite{cunningham2024increasing}. It is also intrinsically difficult: ignitions are sparse, stochastic, geographically diffuse, and conditioned on complex interactions between weather, fuel, topography, and human activity, many of which are inferred or observed only at coarser spatial scales~\cite{Di_Giuseppe_2025}. Existing global wildfire forecasting models address this by operating at the resolution where harmonised data is available rather than the resolution where ignitions and fires occur, producing a probability of fire for each cell in a regular grid at one to nine kilometres and daily cadence~\cite{Di_Giuseppe_2025,kondylatos2023mesogeos,Kondylatos2022,Eddin2023,Kondylatos2025}. These models produce fire-probability fields, which are useful for broad risk assessment but do not represent the structure of fire occurrence. The mismatch is structural: a coarse probability raster does not directly represent discrete future fire occurrences~\cite{Lee_2013}.

This motivates a different output representation. At sub-kilometre resolution, next-day active fire forecasting is a highly sparse-event problem: the prediction domain is dominated by background, while the relevant learning signal is concentrated in a small number of fire pixels. A dense segmentation view therefore makes training dominated by foreground--background imbalance~\cite{Zhong2023, Lin2020}. We present Wildfire Ignition Set Predictor (WISP), which casts active fire forecasting as sparse set prediction. Following DETR-type set prediction~\cite{Carion2020, song2021rethinking}, WISP predicts a fixed-size set of query hypotheses, each with a fire probability and a normalized location; unmatched queries are trained as no-fire through bipartite Hungarian assignment~\cite{Kuhn1955}. 

WISP uses active fire labels derived from the VIIRS active fire product, whose imaging-band observations provide nominal \qty{375}{\meter} spatial resolution~\cite{SCHROEDER2014_VIIRS}. We use this product to define the target reference grid and aggregate future detections into active fire cluster centres. WISP forecasts the next 24 hours of these cluster centres from 48 hours of multi-source Earth observation and reanalysis covariates, including meteorology~\cite{Hersbach2020}, satellite-derived vegetation~\cite{CLMS}, static land descriptors~\cite{Hersbach2020, JASIEWICZ2013, NOBRE2011_hand}, and VIIRS active fire history~\cite{SCHROEDER2014_VIIRS} (see \autoref{Sec3:method}). Rather than predicting a scalar probability for every cell of a coarse forecast grid, WISP outputs a fixed budget of fire query hypotheses, each with a fire score and a normalized location.

The contributions of this work are three-fold. First, we reformulate next-day active-fire forecasting as point-set prediction and introduce WISP, a query-based model that predicts a ranked set of localized future fire-cluster centres. Second, we identify a score-role conflict specific to sparse wildfire set prediction, where the same score affects assignment, ranking, and query activation, and mitigate it through asymmetric classification–localization weighting in matching and loss. Third, we build a globally distributed, hourly, multi-source benchmark and evaluation protocol for active fire forecasting using VIIRS-derived targets~\cite{SCHROEDER2014_VIIRS}.

\section{Related work}
\label{Sec2:related_work}

\textbf{Machine learning for wildfire forecasting.} Machine learning has been applied to several wildfire-related tasks, including danger forecasting~\cite{Huot2022,Apostolakis2022, McNorton2024,Rosch2024}, spread prediction~\cite{Shadrin2024}, burned-area probability and mapping~\cite{Son2024}, and post-fire assessment. The line most relevant here, however, is short-term wildfire forecasting. Prior work in this setting typically predicts a cell-wise danger score or fire probability on a regular spatio-temporal grid, using either tree-based models on engineered covariates or neural networks trained on a wildfire datacube~\cite{Di_Giuseppe_2025,kondylatos2023mesogeos,Kondylatos2022,Kondylatos2025}. This literature has evolved from regional case studies to benchmark datasets and, more recently, to global forecasting systems. Kondylatos et al.~\cite{Kondylatos2022} forecast next-day wildfire danger in the Eastern Mediterranean and use xAI to analyse the contribution of key drivers. Mesogeos~\cite{kondylatos2023mesogeos} later standardized this setting with a 17-year Mediterranean datacube at 1 km $\times$ 1 day resolution, together with benchmark tracks for short-term danger forecasting and burned-area estimation given an ignition point (similar to Shadrin et al.~\cite{Shadrin2024}). More recently, Di Giuseppe et al.~\cite{Di_Giuseppe_2025} demonstrated global data-driven prediction of fire activity by combining weather, fuel, and ignition information. These studies establish the value of data-driven wildfire forecasting and treat the prediction primarily as dense gridded field estimation. 

\textbf{Set prediction for sparse localization.} Set prediction is a natural framework when the target is a sparse, unordered set of entities. DETR~\cite{Carion2020} introduced set prediction with learned object queries and bipartite Hungarian matching~\cite{Kuhn1955} for object detection. Subsequent DETR-style work improved convergence, multi-scale feature use, query design, and matching stability~\cite{zhu2021deformable,liu2022dabdetr,Liu2023_stable_matching}. This formulation has since been used beyond standard box detection. P2PNet~\cite{song2021rethinking} directly predicts crowd points and matches them to head annotations with Hungarian assignment; Oriented DETR extends query-based set prediction to oriented objects in aerial imagery through a point-axis representation~\cite{Zhao2025pta}; and MapTR~\cite{Liao2023} represents vectorized map elements as structured point sets with hierarchical bipartite matching. These examples show that set prediction is well suited to sparse spatial outputs. WISP brings this sparse localization paradigm to next-day active fire forecasting.

\section{WISP: Wildfire Ignition Set Predictor}
\label{Sec3:method}

\subsection{Problem Formulation}
\label{Sec3:1_problem_formulation}

We formulate next-day wildfire active fire prediction as conditional fire cluster set prediction (see~\autoref{fig:1_model_structure}). Given historical Earth-system states $\mathbf{X}^{h}$ and future-known covariates $\mathbf{X}^{f}$, the goal is to predict a variable-length set of future fire-cluster centres over the next $T_p$ hours. Rather than extrapolating past fire observations alone, the objective is to estimate the conditional distribution of future fire-cluster locations given historical Earth-system states and future-known features: 
\begin{equation}
\label{eq:0_conditional_distribution}
    P_{\theta}\left(\mathcal{G}_{T_p}\mid\mathbf{X}^{h},\mathbf{X}^{f}\right),
\end{equation}
where $\theta$ denotes the learnable parameters; $\mathcal{G}_{T_p}$ denotes the set of active fire clusters over the next $T_p$ hours; $\mathbf{X}^{h}$ is the historical context up to time $t$, and $\mathbf{X}^{f}$ contains future-known covariates over the prediction horizon. The future fire field is not determined by historical fire observations alone, but by their interaction with meteorology, vegetation state, static land properties, and expected future weather.

\subsection{Model Architecture}
\label{Sec3:2_architecture}

\begin{figure}[t]
    \centering
    \includegraphics[width=1\linewidth]{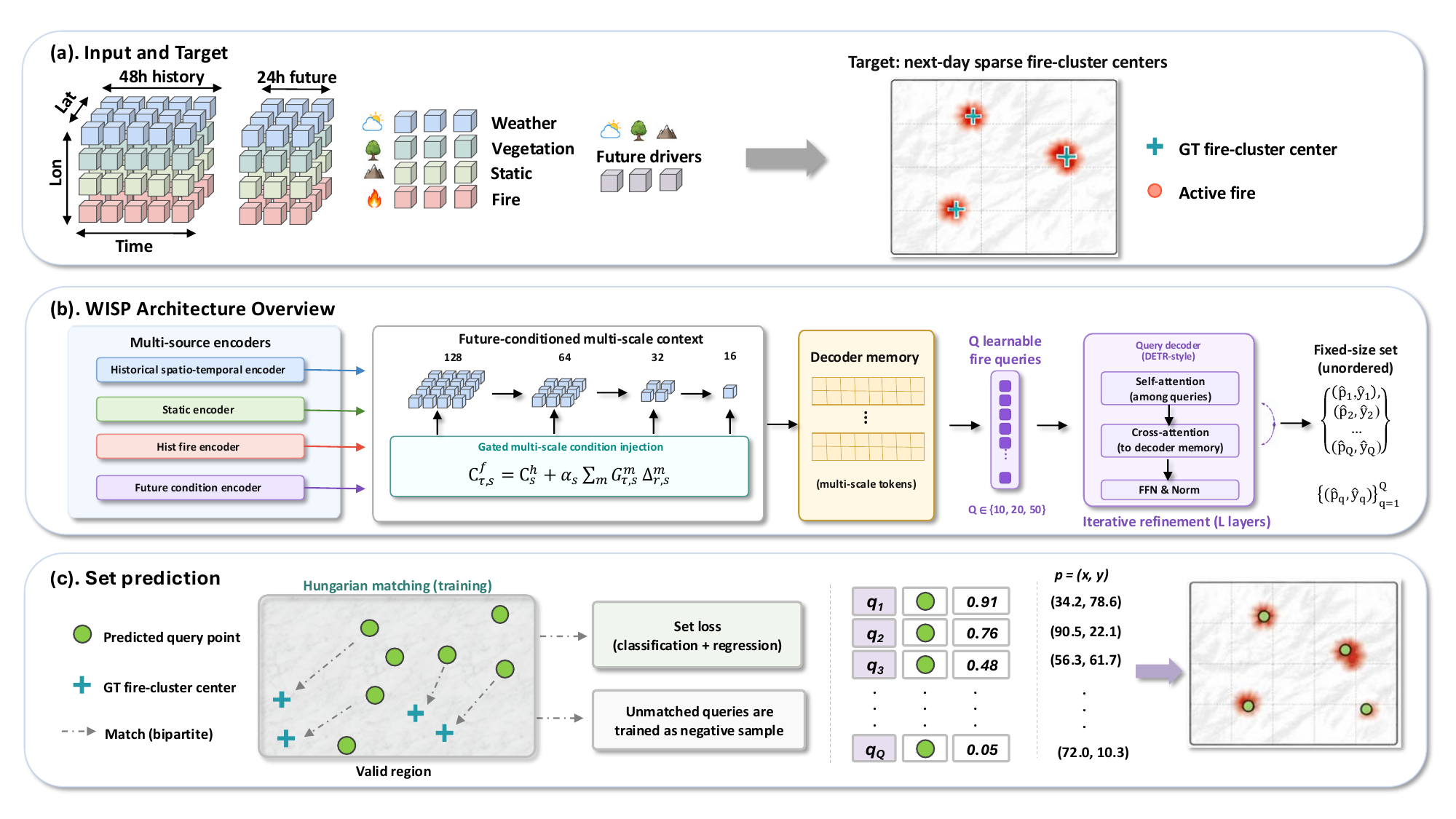}
    \caption{Overview of WISP. \label{fig:1_model_structure}}
\end{figure}

\textbf{Input Entity.} The input to the model is a preprocessed multi-source entity (see (a) in~\autoref{fig:1_model_structure}):
\begin{equation}
\label{eq:1_entity}
\mathbf{X}\in\mathbb{R}^{F\times T\times H\times W}, \quad \mathbf{X} = \left\{\mathbf{W}, \mathbf{V}, \mathbf{S}, \mathbf{A} \right\},
\end{equation}
where $F=55$ denotes the number of input channels, $T=72$ is the temporal length, and $H=W=128$ is the spatial context size. The entity contains four groups of variables: $\mathbf{W}$ contains 34 weather variables (e.g., temperature), $\mathbf{V}$ contains 5 vegetation-related features (e.g., leaf area index), $\mathbf{S}$ contains 14 static variables (e.g., elevation), and $\mathbf{A}$ contains 2 fire observations variables: active fire and fire radiative power (FRP). The full feature list is in \autoref{App_feature_list}. Each entity is divided in to hist $T_h=48$ input $\mathbf{X}^{h}$ and future $T_p=24$ input $\mathbf{X}^{f}$:
\begin{equation}
\label{eq:3_historical_inputs}
    \mathbf{X}^{h} = \left\{\mathbf{W}_{t-T_h+1:t}, \mathbf{V}_{t-T_h+1:t}, \mathbf{S}, \mathbf{A}_{t-T_h+1:t}  \right\},
    \quad
    \mathbf{X}^{f} = \left\{\mathbf{W}_{t+1:t+T_p}, \mathbf{V}_{t+1:t+T_p}, \mathbf{S} \right\},
\end{equation}
where $\mathbf{W}$ and $\mathbf{V}$ are used in the historical and future windows, but $\mathbf{V}$ is quasi-static within 3 days; $\mathbf{S}$ is encoded once as a time-independent land prior. For encoding fire history, WISP derives a fire mask, a fire-recency feature, and two summary maps indicating historical fire occurrence and recency. These channels form the historical fire representation $\mathbf{A}^{h}_{\mathrm{hist}}$ and summary map $\mathbf{B}^{h}_{\mathrm{sum}}$ (see \autoref{App_hist_fire_feature}).

\textbf{Target.} The target is a variable length set of future fire cluster centres. We aggregate the future active fire observations $\mathbf{A}_{t+1:t+T_p}$ into a temporal union mask $\tilde{\mathbf{A}}$. Nearby fire pixels are then grouped into connected components using a $7\times7$ footprint, equivalent to a Chebyshev radius of 3 pixels. Since the VIIRS active fire product has a nominal spatial resolution of \qty{375}{\meter} \cite{SCHROEDER2014_VIIRS}, this radius corresponds to approximately \qty{1}{\kilo\meter}. This grouping prevents adjacent detections from the same fire region from being treated as independent fire-cluster targets:
\begin{equation}
\label{eq:7_target}
    \mathcal{G}_{K} = \{\mathbf{g}_k\}_{k=1}^{K},
    \qquad
    \mathbf{g}_k\in[0,1]^2,
\end{equation}
where $\mathbf{g}_k$ is the FRP-weighted normalized centre location of active fire cluster and $K$ is the number of connected components in $\tilde{\mathbf{A}}$. The model outputs a fixed-size query set:
\begin{equation}
\label{eq:8_output}
    \hat{\mathcal{Y}} = \{(\hat{p}_q,\hat{\mathbf{y}}_q)\}_{q=1}^{Q},
\end{equation}
where $Q=10$ is the default setting; $\hat{p}_q$ is the fire probability and $\hat{\mathbf{y}}_q\in[0,1]^2$ is the predicted cluster location. Unused queries are trained as no-fire queries (see (c) in~\autoref{fig:1_model_structure}). 

\textbf{Spatio-Temporal Encoder.} After adding the derived fire-history channels, the historical spatio-temporal encoder in \autoref{fig:1_model_structure}(b) encodes $\mathbf{X}^{h}$ into a historical context $\mathbf{Z}^{h}$:
\begin{equation}
\label{eq:9_target}
\mathbf{Z}^{h} = \left[ \mathbf{W}^{h},
\mathbf{V}^{h}, \mathbf{A}^{h}_{\mathrm{hist}}, \mathbf{S}^{seq}, \mathbf{B}^{h,seq}_{\mathrm{sum}} \right] \in
    \mathbb{R}^{T_h\times C_{in}\times H\times W}, \quad C_{in} = 59
\end{equation}
where $\mathbf{S}^{\mathrm{seq}}$ and $\mathbf{B}^{h,\mathrm{seq}}_{\mathrm{sum}}$ denote $\mathbf{S}$ and $\mathbf{B}^{h}_{\mathrm{sum}}$ broadcast along the temporal dimension $T_h$, respectively. The historical tensor $\mathbf{Z}^{h}$ is processed by a spatio-temporal backbone encoder~\cite{Gao2022_Earthformer}. A stride-4 patch embedding reduces each $128\times128$ frame to a $32\times32$ latent grid and projects the input channels to an embedding dimension $d_e=128$: 
\begin{equation}
\label{eq:patch_embedding}
\begin{aligned}
    \mathbf{E}^{h}
    &=
    \operatorname{PatchEmbed}(\mathbf{Z}^{h}),\\
    \widetilde{\mathbf{E}}^{h}
    &=
    \mathbf{E}^{h}
    +
    \mathbf{P}^{h}_{1:T_h},
\end{aligned}
\qquad
\mathbf{E}^{h},\mathbf{P}^{h}_{1:T_h},\widetilde{\mathbf{E}}^{h}
\in
\mathbb{R}^{T_h\times d_e\times H'\times W'},
\end{equation}
where $\mathbf{P}^{h}_{1:T_h}$ denotes a learnable spatio-temporal positional embedding added before attention and $\widetilde{\mathbf{E}}^{h}$ is then processed by space-time self-attention blocks, using cuboids of size $4\times8\times8$ and global vectors for cross-cuboid communication~\cite{Gao2022_Earthformer}. Instead of collapsing the historical window into a single feature map, the backbone preserves a compact $T_h$-step historical sequence at $32\times32$ resolution. After the attention blocks, an output projection produces a 64-channel temporal representation, which is compressed by a history context projector. The projector first reduces each temporal feature from 64 to 16 channels and then folds the temporal dimension into the channel dimension:
\begin{equation}
\label{eq:history_projector_trace}
[T_h,64,32,32]
\;\rightarrow\;
[T_h,16,32,32]
\;\rightarrow\;
[T_h\!\cdot\!16,32,32].
\end{equation}
The folded representation is further converted into a four-level historical context pyramid:
\begin{equation}
\label{eq:hist_context_pyramid}
\mathcal{C}^{h} = \left\{\mathbf{C}^{h}_{s}\right\}, 
\quad
s\in\{128,64,32,16\}
\end{equation}
where the subscripts indicate the spatial resolution of each context level. These multi-scale historical features are used by the subsequent condition-injection modules.

\textbf{Land and Fire Feature Encoding.} In parallel to the historical spatio-temporal encoder, WISP encodes three auxiliary sources into scale-aligned spatial pyramids: the static land prior
$\mathbf{S}$, the historical fire sequence $\mathbf{A}^{h}_{\mathrm{hist}}$, and the historical fire-summary map $\mathbf{B}^{h}_{\mathrm{sum}}$:
\begin{equation}
\label{eq:prior_pyramids}
\Phi_{\mathrm{land}}\left(\mathbf{S}\right)
=
\left\{ \mathbf{S}_{s} \right\},
\;
\Phi_{\mathrm{af}}\left(\mathbf{A}^{h}_{\mathrm{hist}}\right)
=
\left\{\mathbf{A}^{h}_{s} \right\},
\;
\Phi_{\mathrm{fire}}\left(\mathbf{B}^{h}_{\mathrm{sum}}\right)
=
\left\{\mathbf{B}^{h}_{s}\right\},
\;
s\in\{128,64,32,16\}.
\end{equation}
The three encoders do not share parameters, and their outputs are aligned with the historical context pyramid $\mathcal{C}^{h}$. These pyramids provide scale-wise land and fire priors for the subsequent gated condition-injection module, as shown in~\autoref{fig:1_model_structure}(b). Architectural details are provided in \autoref{App_auxiliary_encoders}.

\textbf{Future Condition Encoding.} WISP then encodes the future drivers over the prediction horizon, as shown in~\autoref{fig:1_model_structure}(b). For each future step $\tau\in\{1,\ldots,T_p\}$, the model separately encodes future weather and future vegetation variables in $\mathbf{X}^{f}$ into multi-scale condition features:
\begin{equation}
\label{eq:step_future_conditions}
\mathcal{W}^{f}_{\tau} = \left\{ \mathbf{W}^{f}_{\tau,s} \right\},
\quad
\mathcal{V}^{f}_{\tau} = \left\{ \mathbf{V}^{f}_{\tau,s} \right\}, 
\quad
s\in\{128,64,32,16\}.
\end{equation}
To provide each future step with horizon-level context, WISP also encodes the $T_p$-step future features into one future-sequence context. The step-wise and future-sequence context features are added at each scale:
\begin{equation}
\label{eq:future_condition_sum}
\bar{\mathbf{W}}^{f}_{\tau,s} = \mathbf{W}^{f}_{\tau,s} + \mathbf{W}^{f}_{\mathrm{seq},s},
\quad
\bar{\mathbf{V}}^{f}_{\tau,s} = \mathbf{V}^{f}_{\tau,s} + \mathbf{V}^{f}_{\mathrm{seq},s},
\quad
s\in\{128,64,32,16\}.
\end{equation}
Thus, each prediction step receives both its local future feature and the global future trajectory.

\textbf{Gated Multi-Scale Condition Injection.} WISP constructs a conditioned context pyramid for each future step (\autoref{fig:1_model_structure}(b)). For each scale $s$ and forecast step $\tau$, the model starts from the historical context feature $\mathbf{C}^{h}_{s}$ and injects five condition features: $\mathcal{M} = \{\mathcal{W}^{f}_{\tau}, \mathcal{V}^{f}_{\tau}, \Phi_{\mathrm{land}}\left(\mathbf{S}\right), \Phi_{\mathrm{af}}\left(\mathbf{A}^{h}_{\mathrm{hist}}\right), \Phi_{\mathrm{fire}}\left(\mathbf{B}^{h}_{\mathrm{sum}}\right)\}$.

Rather than concatenating all sources directly, WISP uses gated residual injection~\cite{Srivastava2015, Perez2018_FiLM}. At each scale, every source is first projected into the same channel space as the historical context. A source-specific gate then controls how much that source modifies the historical feature:
\begin{equation}
\label{eq:gated_condition_injection}
\mathbf{C}^{f}_{\tau,s} = \mathbf{C}^{h}_{s} + \alpha_s \sum_{m\in\mathcal{M}} \mathbf{G}^{m}_{\tau,s}
\odot \Delta^{m}_{\tau,s}.
\end{equation}
Here, $\Delta^{m}_{\tau,s}$ is the residual update generated from source $m$; $\mathbf{G}^{m}_{\tau,s}\in[0,1]$ is its learned gate, and $\alpha_s$ is a scale-dependent injection coefficient. In WISP, the high-resolution features are updated more conservatively, while lower-resolution semantic features receive stronger conditioning~\cite{Tan2020}: $\alpha_{128}=0.25, \alpha_{64}=0.50, \alpha_{32}=1.00, \alpha_{16}=1.00.$ The output is a future-conditioned context pyramid for each forecast step:
\begin{equation}
\label{eq:future_conditioned_context}
\mathcal{C}^{f}_{\tau} = \left\{ \mathbf{C}^{f}_{\tau,s}\right\}, 
\quad
s\in\{128,64,32,16\}.
\end{equation}
This design preserves the multi-scale structure of the historical representation while allowing future forcing and land/fire priors to modulate it in a source-aware and scale-aware manner. The resulting conditioned pyramid $\mathcal{C}^{f}_{\tau}=\{\mathbf{C}^{f}_{\tau,s}\}$
is then converted into a fixed-resolution decoder memory:
\begin{equation}
\label{eq:future_memory}
\mathbf{M}_{\tau} = \operatorname{Flatten} \left( \Gamma_{\mathrm{mem}}(\mathcal{C}^{f}_{\tau})
\right) \in \mathbb{R}^{1024\times128},
\mathbf{M} = \left[ \mathbf{M}_{1},\ldots,\mathbf{M}_{T_p} \right].
\end{equation}
Here, $\Gamma_{\mathrm{mem}}$ projects the four pyramid levels to a common $32\times32$ memory grid with 128 channels, and $\operatorname{Flatten}(\cdot)$ converts it into $1024$ spatial tokens. The
stacked memory $\mathbf{M}$ is used as the future-conditioned spatio-temporal context for the query decoder.

\textbf{Query Decoding.} Given the future-conditioned memory $\mathbf{M}$, WISP predicts a fixed-size
set of future fire-cluster hypotheses using a query decoder (\autoref{fig:1_model_structure}). The memory is first
flattened over the prediction horizon and spatial tokens:
\begin{equation}
\label{eq:decoder_memory_flatten}
\mathbf{M}
\in
\mathbb{R}^{T_p\times1024\times128}
\;\longrightarrow\;
\mathbf{M}^{\flat}
\in
\mathbb{R}^{(T_p\cdot1024)\times128}.
\end{equation}
A positional embedding is added to each memory token using its forecast step, row index, and column index. This allows the decoder to distinguish not only where a memory token is located, but also which future hour it represents. The decoder uses $Q$ learnable queries. Each query has a content embedding and a learnable reference point:
\begin{equation}
\label{eq:query_initialization}
\mathbf{q}^{0}_{q}\in\mathbb{R}^{128},
\qquad
\mathbf{r}^{0}_{q}\in[0,1]^2,
\qquad
q=1,\ldots,Q.
\end{equation}
The reference point $\mathbf{r}^{0}_{q}$ is defined in the normalized valid prediction region (see \autoref{Sec4:experiments}). WISP maps the reference point back to the full context coordinates when constructing query positional embeddings. Following DETR-style query decoding and bipartite matching design~\cite{Carion2020, liu2022dabdetr}, each decoder layer performs self-attention among queries, cross-attention to the memory tokens, and a feed-forward update. The query reference point is then refined by an additive offset in inverse-sigmoid space:
\begin{equation}
\label{eq:query_refinement}
\mathbf{r}^{\ell}_{q} = \sigma \left( \sigma^{-1}(\mathbf{r}^{\ell-1}_{q}) + \Delta^{\ell}_{q}
\right),
\quad
\ell=1,\ldots,L.
\end{equation}
In the current implementation, $L=2$. This iterative update gives each query an explicit spatial hypothesis and allows the decoder to progressively move it toward a future fire-cluster centre. The final query states are mapped to fire classification logits and normalized point coordinates:
\begin{equation}
\label{eq:query_outputs}
\hat{\mathcal{Y}} = \left\{ (\hat{p}_{q},\hat{\mathbf{y}}_{q}) \right\}_{q=1}^{Q},
\quad
\hat{p}_{q} = \operatorname{softmax}(\mathbf{z}_{q})_{1}, \quad \hat{\mathbf{y}}_{q} = \mathbf{r}^{L}_{q}.
\end{equation}
Here, $\hat{p}_{q}$ is the predicted fire probability and $\hat{\mathbf{y}}_{q}\in[0,1]^2$ is the predicted fire-cluster location in the valid region (see~\autoref{eq:7_target}). 

\textbf{Hungarian Set Objective.} WISP is trained as a fixed-size set predictor. For each entity, the decoder outputs $Q$ fire-cluster predictions, while the target is a variable-size set $\mathcal{G}_{K}=\{\mathbf{g}_{k}\}_{k=1}^{K}$. Following DETR-style set prediction~\cite{Carion2020} and the Hungarian matching algorithm~\cite{Kuhn1955}, we first compute a one-to-one assignment between predicted queries and ground-truth fire clusters (see~\autoref{fig:1_model_structure}(c)). For query $q$ and target cluster $k$, the matching cost is defined as
\begin{equation}
\label{eq:hungarian_cost}
\mathcal{D}_{qk} = \lambda_{\mathrm{m,cls}} \left(-\hat{p}_{q}\right) + \lambda_{\mathrm{m,loc}} \left\|
\hat{\mathbf{y}}_{q} - \mathbf{g}_{k} \right\|_{1},
\end{equation}
where $\hat{p}_{q}$ is the predicted fire probability and $\hat{\mathbf{y}}_{q}\in[0,1]^2$ is the predicted cluster location. The first term encourages a ground-truth cluster to be assigned to a confident fire query, whereas the second term favours spatially close predictions. The assignment is then obtained by solving
\begin{equation}
\label{eq:hungarian_assignment}
\pi^{\star} = \arg\min_{\pi} \sum_{(q,k)\in\pi} \mathcal{D}_{qk},
\end{equation}
where $\pi$ ranges over one-to-one matchings between predicted queries and target clusters. This matching step determines which queries are treated as positive fire predictions and queries not selected by $\pi^{\star}$ are assigned to the no-fire class. An illustrative query-to-cluster cost matrix is provided in \autoref{App_Hungarian_cost_matrix}. After matching, WISP applies classification supervision to all queries and localization supervision only to matched queries. Let $c_q^\star=1$ if query $q$ is matched to a target cluster and $c_q^\star=0$ otherwise (see Hungarian Matching algorithm in~\autoref{App_algo_Hungarian}). The classification loss is
\begin{equation}
\label{eq:set_cls_loss}
\mathcal{L}_{\mathrm{cls}} = \frac{1}{Q} \sum_{q=1}^{Q} \mathrm{CE} \left( \mathbf{z}_{q}, c_q^\star;
\mathbf{w}_{\mathrm{cls}} \right),
\quad
\mathbf{w}_{\mathrm{cls}} = \{w_{\mathrm{eos}}, 1\},
\end{equation}
where $w_{\mathrm{eos}}$ down-weights the no-fire class. We use $w_{\mathrm{eos}}=0.1$, which reduces the dominance of unmatched queries in the loss. The localization loss is computed only for matched query-target pairs:
\begin{equation}
\label{eq:set_loc_loss}
\mathcal{L}_{\mathrm{loc}} = \frac{1}{2|\pi^{\star}|} \sum_{(q,k)\in\pi^{\star}} \left\| \hat{\mathbf{y}}_{q} - \mathbf{g}_{k} \right\|_{1},
\end{equation}
where the factor $2$ normalizes over the two coordinate dimensions; and with $\lambda_{\mathrm{loc}}=5.0$, the loss function is
\begin{equation}
\label{eq:set_total_loss}
\mathcal{L} = \mathcal{L}_{\mathrm{cls}} + \lambda_{\mathrm{loc}} \mathcal{L}_{\mathrm{loc}}.
\end{equation}

\section{Experiments}
\label{Sec4:experiments}

\subsection{Data and Settings}
\label{Sec4_1:data}
\textbf{Data.} We build a global wildfire ignition dataset covering 11 years, 2014--2024, by integrating multiple Earth observation and reanalysis products with heterogeneous spatial and temporal resolutions (see feature list in \autoref{App_feature_list}). All variables are harmonized onto a common \qty{375}{\meter} spatial grid and hourly temporal resolution. For each year, we sample three $3^\circ\times3^\circ\times 15 \;\text{day}$ region cubes from each of 14 GFED regions~\cite{GFED5}, giving 42 region cubes per year, from which we extract 10 entities per region. Each entity $x_i \in \mathbf{X}$ contains 34 weather variables~\cite{Hersbach2020}, 5 vegetation variables~\cite{CLMS}, 14 static land variables~\cite{JASIEWICZ2013,Hersbach2020,CLMS}, and 2 fire-observation variables~\cite{SCHROEDER2014_VIIRS}. The 72 hourly frames are divided into $x^h_i \in \mathbf{X}^{h}$ and $x^f_i \in \mathbf{X}^{f}$. After removing entities with excessive missing values in high-latitude regions, the dataset contains 2863 training entities, 406 validation entities, and 816 test entities. Approximately 20\% of the entities contain no future fire, while about 25\% contain more than 20 active fire pixels with confidence level at least 2. Preprocessing, augmentation, training, and hyperparameter details are provided in \autoref{App_data_preprocessing_model_hyperparameter}. 

\subsection{Results}
\label{Sec4_2:results}

\begin{table*}[!h]
\centering
\caption{Detection and coverage results on the test dataset in \%.}
\label{tab:1_main_detection_coverage_results}
\resizebox{\textwidth}{!}{
\begin{tabular}{lccccc cccc ccc}
\toprule
WISP & $Q$ & $\lambda_{\mathrm{m,cls}}$ & $\lambda_{\mathrm{m,loc}}$ & $\lambda_{\mathrm{loc}}$ 
& AP@7 & AP@14 & AP@21 & mAP 
& MassCov@14 & Hit@14 & Union AUROC \\
\midrule
v1 & 10 & 0 & 1 & 5
& 6.4 & 23.0 & 33.4 & 20.9
& 42.8 & 44.6 & 74.9 \\

v2 & 10 & 1 & 1 & 5
& 4.4 & 21.3 & 36.6 & 20.8
& 38.6 & 41.3 & 64.6 \\

v3 & 10 & 1 & 2 & 5
& 5.5 & 22.4 & 36.1 & 21.3
& 37.4 & 41.5 & 72.8 \\

v4 & 10 & 1 & 5 & 5
& 5.7 & 22.7 & 34.8 & 21.1
& 41.6 & 44.0 & 73.9 \\

v5 & 20 & 1 & 2 & 5
& 8.7 & 30.0 & 42.0 & 26.9
& 42.3 & 45.4 & \textbf{76.4} \\

v6 & 50 & 1 & 2 & 5
& \textbf{12.8} & \textbf{38.2} & \textbf{50.0} & \textbf{33.7}
& \textbf{53.4} & \textbf{54.1} & 73.5 \\
\bottomrule
\end{tabular}
}
\end{table*}

\textbf{Quantitative results.} In this section, we report the test-set results of six WISP variants with
different query budgets $Q$, Hungarian matching weights for classification and localization $(\lambda_{\mathrm{m,cls}}, \lambda_{\mathrm{m,loc}})$, and localization loss weights $\lambda_{\mathrm{loc}}$. \autoref{tab:1_main_detection_coverage_results} reports the main detection and coverage results on the test dataset. AP@$r$ are confidence-ranked event average precision scores: a predicted query is counted as a true positive if it falls within $r \in \{7,14,21\}$ pixels of an unmatched ground-truth fire-cluster centre, respectively. As the resolution is \qty{375}{\meter}, 14 pixels correspond to approximately \qty{5}{km}. MassCov@14 measures the FRP-weighted fire-cluster mass covered by the matched query set, while Hit@14 measures the fraction of matched query --- cluster pairs localized within the same radius. Union AUROC measures the threshold-free pixel-level separability of the rendered 24-hour union map, i.e., whether ground-truth fire pixels receive higher scores than background pixels. The exact definitions are provided in \autoref{App_eval_metrics}.

Table~\ref{tab:2_main_query_diagnostic_results} summarizes query-level diagnostics at a fixed operating point. ClusPrec@14 and ClusRec@14 measure the precision and recall of the selected query set, Top10Rec@14 measures whether the highest-ranked queries cover the true clusters (10 is the default set in WISP), and Localization Recall Precision (LRP)@14 summarizes localization, false-positive, and false-negative errors~\cite{Oksuz2018}. Card. err., Dup., and Avg. pred. further quantify whether the model predicts the right number of clusters or wastes queries on redundant detections. 

\begin{table*}[!t]
\centering
\caption{Operating-point and query-set diagnostics on the test dataset in \%.}
\label{tab:2_main_query_diagnostic_results}
\resizebox{\textwidth}{!}{
\begin{tabular}{lccccc ccccccc}
\toprule
WISP & $Q$ &
$\lambda_{\mathrm{m,cls}}$ &
$\lambda_{\mathrm{m,loc}}$ &
$\lambda_{\mathrm{loc}}$ &
ClusPrec@14 & ClusRec@14 & Top10Rec@14 & LRP@14 $\downarrow$ &
Card. err.\textsuperscript{*} $\downarrow$ & Dup. $\downarrow$ & Avg. pred.\textsuperscript{*} $\downarrow$ \\
\midrule
v1 & 10 & 0 & 1 & 5
& 18.0 & 36.2 & 48.8 & \textbf{42.3}
& 2.81 & 31.5 & 4.82 \\

v2 & 10 & 1 & 1 & 5
& 18.1 & 32.4 & 55.4 & 49.9
& 1.86 & 25.0 & 3.75 \\

v3 & 10 & 1 & 2 & 5
& \textbf{24.8} & 32.1 & 57.4 & 48.2
& \textbf{1.57} & \textbf{20.5} & \textbf{3.14} \\

v4 & 10 & 1 & 5 & 5
& 21.4 & 35.9 & \textbf{58.2} & 48.8
& 2.00 & 25.0 & 3.81 \\

v5 & 20 & 1 & 2 & 5
& 22.4 & 40.7 & 53.1 & 49.8
& 2.58 & 27.7 & 4.64 \\

v6 & 50 & 1 & 2 & 5
& 21.7 & \textbf{49.6} & 57.0 & 45.7
& 3.94 & 31.6 & 6.34 \\
\bottomrule
\end{tabular}
}
\parbox{\textwidth}{\footnotesize \textsuperscript{*} Card. err. and Avg. pred. are reported in query-count units, not percentages.}
\end{table*}


\textbf{Query budget controls coverage and redundancy.}
Comparing WISPv3, WISPv5, and WISPv6 shows that the query budget is a structural capacity factor. Increasing $Q$ reduces ground-truth truncation from $8.58\%$ at $Q=10$ to $1.96\%$ at $Q=20$ and $0.25\%$ at $Q=50$, which explains the gains in AP, MassCov@14, Hit@14, and ClusRec@14. WISPv6 achieves the strongest coverage metrics, with $38.2\%$ AP@14 and $53.4\%$ MassCov@14. This recall gain, however, comes with a larger warning burden: compared with WISPv3, cardinality error increases from $1.57$ to $3.94$, duplicate rate from $20.5\%$ to $31.6\%$, and Avg. pred. from $3.14$ to $6.34$. Thus, larger query budgets improve coverage, while smaller budgets provide more compact predictions.

\textbf{Matching cost controls query compactness.} Among the $Q=10$ variants, WISPv1 (location-only matching) gives the highest AP@14 and relatively strong ClusRec@14, suggesting that geometry-dominated assignment encourages spatial coverage. However, its prediction set is less controlled: compared with WISPv3, cardinality error increases from $1.57$ to $2.81$, duplicate rate from $20.5\%$ to $31.5\%$, and the average number of predictions from $3.14$ to $4.82$. Adding the classification term makes the query set more compact and less redundant (see~\autoref{tab:2_main_query_diagnostic_results}). The $\lambda_{\mathrm{m,cls}}=1,\lambda_{\mathrm{m,loc}}=2$ configuration gives the best operational trade-off among the $Q=10$ models, with the highest ClusPrec@14, lowest cardinality error, lowest duplicate rate, and fewest average predictions, while also achieving the best mAP within this group.

\begin{figure}[!h]
    \centering
    \includegraphics[width=1\linewidth]{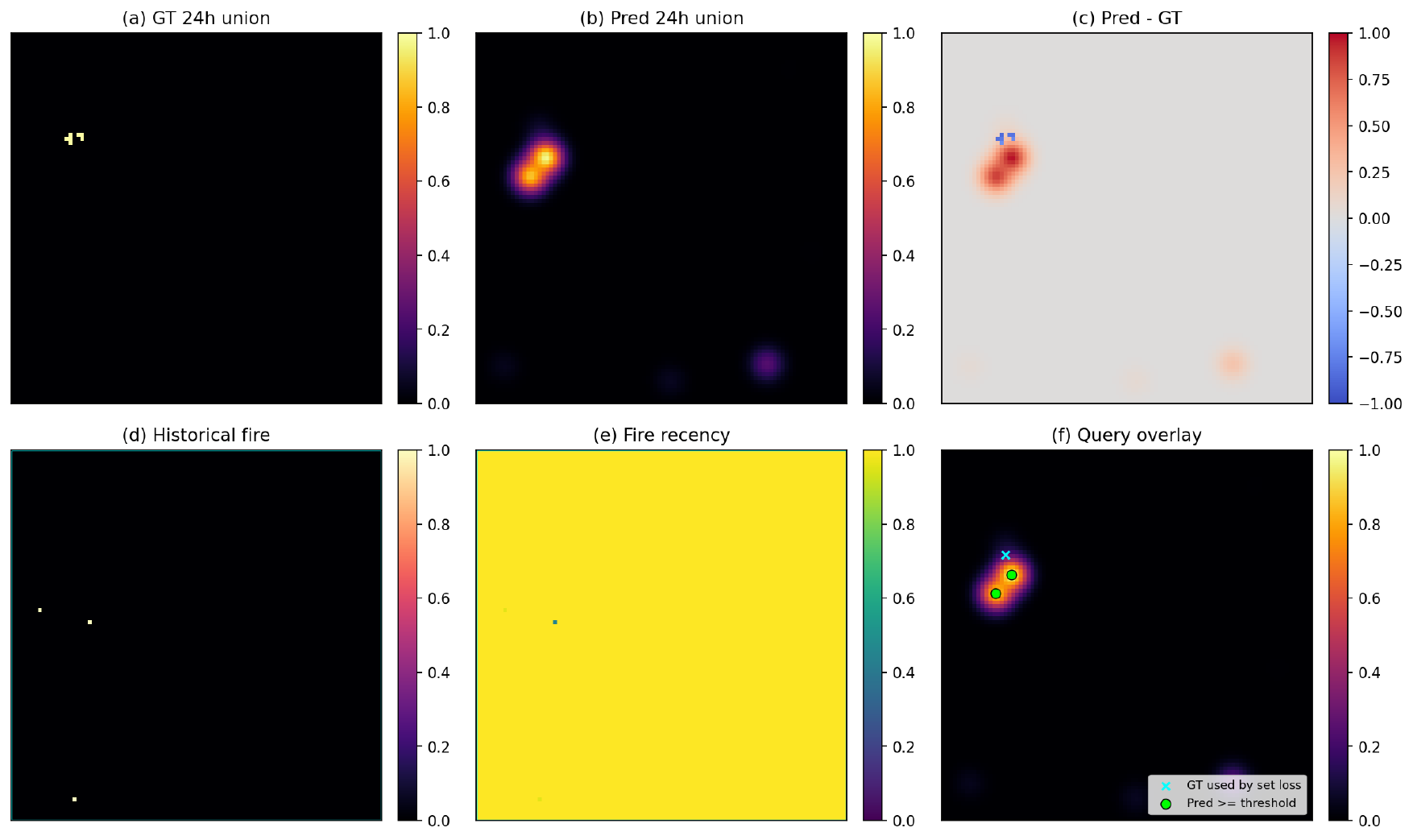}
    \caption{Qualitative set-prediction example on one test entity. (a) Ground-truth (GT) 24-hour active fire union, (b) rendered WISP prediction, (c) residual map computed as prediction minus GT, (d) historical fire evidence, (e) fire recency map, and (f) query overlay. In (f), cyan crosses denote GT centres, and green circles denote predicted queries above the operating threshold of 0.5.
    \label{fig:qualitative_set_prediction}}
\end{figure}

\textbf{Qualitative results.} Figure~\ref{fig:qualitative_set_prediction} visualizes a representative test entity using WISPv3. The ground-truth future fire activity is sparse, and WISPv3 produces a compact set of query predictions near the observed cluster. The query overlay shows that predicted centres are close to the GT centre, supporting the intended set-prediction behaviour: WISP represents the future fire region with point hypotheses rather than a dense segmentation mask. The predictions are not simply anchored to historical fire pixels or low-recency locations, suggesting that the model combines historical context, future forcing, and static/fire priors. Additional good, median, and failure cases are provided in \autoref{App_qualitative_gallery}; training-time query evolution is shown in \autoref{fig:app_validation_query_evolution}.

\textbf{Query-level evaluation.} The prediction map illustrates why pixel-level metrics are not used as the primary evaluation. WISP predicts a finite set of fire-cluster queries, and the displayed 24-hour union is only a Gaussian rendering of these points. Thus, a correctly localized query can appear smoother and spatially larger than the binary ground-truth union, producing positive residuals around the true fire cluster. Pixel precision, IoU, Dice/F1, and raster AP/AUPRC would penalize this rendering-induced over-coverage as false-positive area, even when the query centre is close to the target. We therefore treat Union AUROC as an auxiliary raster diagnostic and use event AP, cluster coverage, and query-set diagnostics as the primary metrics. Weak low-probability blobs away from the target region still reveal off-target activation, which motivates reporting duplicate and query-set diagnostics in~\autoref{tab:2_main_query_diagnostic_results}. 

\subsection{Regime-wise Analysis}
\label{Sec4_4:regime}

We further divide the test dataset by past-fire presence ($\mathrm{PF}=1/0$) and future-fire presence ($\mathrm{FF}=1/0$). This yields four active fire regimes: \textbf{New ignition} $(\mathrm{PF}=0,\mathrm{FF}=1)$, \textbf{Continued fire} $(\mathrm{PF}=1,\mathrm{FF}=1)$, \textbf{Extinguished fire} $(\mathrm{PF}=1,\mathrm{FF}=0)$, and \textbf{Quiescent cases} $(\mathrm{PF}=0,\mathrm{FF}=0)$. \autoref{tab:regime_wise_operational} reports regime-wise diagnostics for these four cases. For future-no-fire regimes, Mean prob. denotes the mean predicted probability of the rendered union fire-probability map over the valid region; Avg. pred. is the average number of predicted fire queries above the operating threshold.

\begin{table*}[!h]
\centering
\caption{Regime-wise operational behaviour on the test dataset. Hit@14, AP@14, Rec@14, and Mean prob. are reported in \%.}
\label{tab:regime_wise_operational}
\resizebox{\textwidth}{!}{
\begin{tabular}{l ccc ccc cc cc}
\toprule
& \multicolumn{3}{c}{New ignition $(n=138)$}
& \multicolumn{3}{c}{Continued fire $(n=386)$}
& \multicolumn{2}{c}{Extinguished $(n=59)$}
& \multicolumn{2}{c}{Quiescent $(n=233)$} \\
\cmidrule(lr){2-4}\cmidrule(lr){5-7}\cmidrule(lr){8-9}\cmidrule(lr){10-11}
WISP & Hit@14 & AP@14 & Rec@14 & Hit@14 & AP@14 & Rec@14 
& Avg. pred. $\downarrow$& Mean prob. $\downarrow$& Avg. pred. $\downarrow$& Mean prob. $\downarrow$\\
\midrule
v1 & 12.9 & 5.9 & 9.6 & 56.0 & 25.2 & 45.7 & 5.71 & 2.91 & 0.74 & 0.61 \\
v2 & 15.8 & 7.1 & 10.5 & 50.4 & 23.7 & 40.2 & 3.75 & 2.36 & 0.82 & 0.73 \\
v3 & 12.8 & 10.0 & 10.8 & 51.7 & 24.8 & 39.7 & \textbf{2.97} & \textbf{1.79} & \textbf{0.34} & \textbf{0.24} \\
v4 & 18.8 & 12.0 & 13.9 & 53.0 & 25.0 & 43.8 & 3.80 & 2.27 & 0.55 & 0.38 \\
v5 & 15.7 & 9.3 & 13.9 & 56.0 & 34.8 & 50.2 & 4.10 & 2.46 & 0.59 & 0.43 \\
v6 & \textbf{19.1} & \textbf{12.5} & \textbf{14.2} & \textbf{66.7} & \textbf{43.8} & \textbf{62.2} & 5.59 & 3.93 & 0.88 & 1.14 \\
\bottomrule
\end{tabular}
}
\end{table*}

The corresponding training-set diagnostics are provided in \autoref{App_train_regime}. We use them only to assess model fit and the train--test gap; all conclusions below are based on the test dataset.

\textbf{Historical fire is a strong predictive cue.} \textbf{Continued fire} is substantially easier than \textbf{New ignition}. For WISPv3, AP@14 increases from $10.0\%$ in new ignition cases to $24.8\%$ in continued fire cases, and Hit@14 increases from $12.8\%$ to $51.7\%$. This gap confirms that recent active fire evidence provides a strong cue for future fire activity. 

\textbf{New ignition remains the hardest regime.} When no historical fire is observed, the model must localize future active fire without a local fire-memory cue. All variants have much lower AP@14 in the new ignition regime than in the continued-fire regime. Among the $Q=10$ variants, increasing the localization weight in the matching cost improves new ignition performance: WISPv4 reaches $12.0\%$ AP@14 and $18.8\%$ Hit@14, compared with $10.0\%$ and $12.8\%$ for WISPv3. However, this gain comes with a higher false-alarm burden in future-no-fire regimes.

\textbf{Fire-memory persistence.} For entities with no future fire, there is no target cluster to match; the desired behaviour is therefore to suppress fire-query activations. \textbf{Quiescent cases} is relatively well controlled: WISPv3 predicts only $0.34$ fire queries above the operating threshold on average, with a mean rendered fire probability of $0.24\%$. \textbf{Extinguished fire} is harder because historical fire is present but does not continue into the prediction window. In this case, WISPv3 predicts $2.97$ fire queries on average, with a mean rendered fire probability of $1.79\%$, indicating that recent fire evidence can still induce persistence-like query activations when future fire disappears.

\textbf{Query number trades coverage.} Larger query budgets improve event coverage, especially in the continued-fire regime. WISPv6 achieves the highest continued-fire AP@14 and Hit@14, with $43.8\%$ and $66.7\%$, respectively, and also gives the best new ignition AP@14 and Hit@14. However, this improvement comes with less conservative behaviour in future-no-fire regimes: in \textbf{Extinguished fire}, the average number of predicted fire queries increases to $5.59$, compared with $2.97$ for WISPv3; in \textbf{Quiescent cases}, it increases from $0.34$ to $0.88$. Thus, $Q=50$ is preferable when recall and coverage are prioritized, while WISPv3 provides a more conservative operating point with stronger fire-query suppression.

\subsection{Limitations and Set-Prediction Diagnostics}
\label{Sec4_6:Mechanistic}

WISP’s limitations mainly arise from the sparse set-prediction formulation rather than a single architectural component. 

\textbf{Sparse not dense.} Active fire clusters are sparse localized events, while most of the prediction domain is background. This makes dense semantic segmentation a poor fit for the primary prediction object, especially because WISP uses heterogeneous covariates such as ERA5 weather at \(0.25^\circ\) resolution but predicts on the VIIRS \qty{375}{\meter} grid (see \autoref{Sec3:1_problem_formulation} and \autoref{Sec4_1:data}). A pixel-wise model must assign independent probabilities at a scale much finer than several forcing fields can resolve, and its loss is dominated by foreground--background imbalance. We think ignition is intrinsically sparse: a future fire event occurs only when an ignition source, receptive fuel, and favourable environmental conditions coincide at a localized place and time. This view is consistent with statistical wildfire modelling, where ignition or occurrence is commonly represented as a point pattern~\cite{Koh2023, OPITZ2020}.

\textbf{Surrogate loss versus event metrics.} WISP is trained with a differentiable Hungarian set objective, whereas evaluation depends on event AP, cluster coverage, duplicate suppression, and score ranking (see \autoref{Sec4_2:results}). This creates the usual detection gap between trainable classification/localization losses and non-differentiable ranking metrics~\cite{Chen2019,Tao2021}, but it is sharper here because only a small number of active queries determine both recall and false alarms. A query may be well localized after matching but poorly ranked at inference, while multiple high-scoring queries around one cluster can improve local confidence but hurt duplicate-sensitive diagnostics. We therefore select checkpoints using validation event-level metrics rather than loss.

\textbf{Assignment churn.} Hungarian matching assigns each target cluster to one query, so small changes in query scores or locations can switch which query receives localization supervision. Nearby unmatched queries are then trained as no-fire, producing non-smooth supervision for individual query slots, a known issue in DETR-style models~\cite{Liu2023_stable_matching}. Our matching ablation reflects this tension: the location-only matcher in WISPv1 gives more recall-oriented behaviour, but WISPv3 produces a more compact query set with lower cardinality error, fewer duplicates, and fewer active predictions. This suggests that assignment geometry and final confidence ranking should eventually be decoupled more explicitly~\cite{Pu2023_rank_detr}.

\section{Conclusion}
\label{Sec5:conclusion}

To the best of our knowledge, WISP is the first model that formulates wildfire prediction as a set-prediction problem. WISP predicts a compact ranked set of future fire-cluster centres on the VIIRS \qty{375}{\meter} grid. Experiments show that WISP can localize future fire clusters across different ignition regimes. The ablations further show that query budget and matching design control different aspects of performance: larger query sets improve coverage and recall, while conservative $Q=10$ variants provide more compact predictions with fewer duplicate detections and fewer excessive fire-query activations. Overall, WISP establishes sparse set prediction as a viable path for high-resolution wildfire forecasting, with future work extending it toward more stable assignment, improved score calibration, driver attribution, and coupled ignition--spread modelling.

\section*{Acknowledgement and Disclosure of Funding}
This work has received funding from the European Union’s Horizon Europe WIDERA Coordination and Support Actions under Grant Agreement no.101159723 (MeDiTwin), and from the European Union’s Horizon Europe research and innovation programme under Grant Agreement No: 101120237 (ELIAS).

\newpage

\bibliography{ref}
\bibliographystyle{ieeetr}
\medskip

\newpage
\appendix

\renewcommand{\thesection}{\Alph{section}}
\renewcommand{\thesubsection}{\thesection.\arabic{subsection}}

\makeatletter
\renewcommand{\theHsection}{appendix.\Alph{section}}
\renewcommand{\theHsubsection}{appendix.\Alph{section}.\arabic{subsection}}
\makeatother

\section{Feature List}
\label{App_feature_list}
\begin{table}[!h]
\centering
\tiny
\setlength{\tabcolsep}{2.2pt}
\renewcommand{\arraystretch}{1.08}
\caption{Input features used by WISP, grouped by model feature interface and native data resolution. All variables are harmonized to the VIIRS reference grid before entity extraction.}
\label{tab:feature_list}
\begin{tabularx}{\linewidth}{@{}p{0.13\linewidth}p{0.14\linewidth}p{0.11\linewidth}p{0.07\linewidth}p{0.045\linewidth}X@{}}
\toprule
Group & Product & Temporal res. & Spatial res. & Count & Variables \\
\midrule

\multirow{7}{=}{Weather}
& ERA5 surface
& Hourly
& \(0.25^\circ\)
& 4
& \texttt{t2m}, \texttt{skt}, \texttt{d2m}, \texttt{vpd} \\

& ERA5 surface
& Hourly
& \(0.25^\circ\)
& 3
& \texttt{wind\_speed}, \texttt{blh}, \texttt{cape} \\

& ERA5 surface
& Hourly
& \(0.25^\circ\)
& 2
& \texttt{msl}, \texttt{sp} \\

& ERA5 surface
& Hourly accum.
& \(0.25^\circ\)
& 4
& \texttt{tp}, \texttt{lsp}, \texttt{cp}, \texttt{pev} \\

& ERA5 surface
& Hourly accum.
& \(0.25^\circ\)
& 5
& \texttt{ssr}, \texttt{ssrd}, \texttt{sshf}, \texttt{avg\_snlwrf}, \texttt{avg\_snswrf} \\

& ERA5 surface
& Hourly
& \(0.25^\circ\)
& 6
& \texttt{swvl1}, \texttt{swvl2}, \texttt{swvl3}, \texttt{swvl4}, \texttt{tcrw}, \texttt{tcrw\_mask} \\

& ERA5 pressure levels
& Hourly
& \(0.25^\circ\)
& 10
& \texttt{temperature\_700hpa}, \texttt{temperature\_850hpa}, \texttt{relative\_humidity\_700hpa}, \texttt{wind\_speed\_300hpa}, \texttt{wind\_speed\_850hpa}, \texttt{vertical\_velocity\_700hpa}, \texttt{vertical\_velocity\_700hpa\_mask}, \texttt{geopotential\_700hpa}, \texttt{geopotential\_850hpa}, \texttt{divergence\_300hpa} \\

\addlinespace[1pt]

\multirow{2}{=}{Vegetation}
& Copernicus vegetation
& 10-day
& 300 m
& 4
& \texttt{gdmp}, \texttt{gdmp\_mask}, \texttt{fapar}, \texttt{fcover} \\

& VIIRS vegetation
& 8-day
& 500 m
& 1
& \texttt{lai} \\

\addlinespace[1pt]

\multirow{3}{=}{Land}
& Terrain
& Static
& \(0.1^\circ\)
& 2
& \texttt{elevation}, \texttt{slope} \\

& Hydrologic terrain
& Static
& 90 m
& 1
& \texttt{hand} \\

& Landform classes
& Static
& 90 m
& 10
& \texttt{geomorphon\_1}, \texttt{geomorphon\_2}, \texttt{geomorphon\_3}, \texttt{geomorphon\_4}, \texttt{geomorphon\_5}, \texttt{geomorphon\_6}, \texttt{geomorphon\_7}, \texttt{geomorphon\_8}, \texttt{geomorphon\_9}, \texttt{geomorphon\_10} \\

\addlinespace[1pt]

Human
& Population
& Annual
& 100 m
& 1
& \texttt{population\_density} \\

\addlinespace[1pt]

Fire
& VIIRS active fire
& 12h
& 375 m
& 2
& \texttt{frp}, \texttt{active\_fire} \\

\midrule
Total
& --
& --
& --
& 55
& -- \\
\bottomrule
\end{tabularx}
\end{table}

\section{Historical Fire Feature Construction}
\label{App_hist_fire_feature}

WISP derives two additional fire-history features from the historical active-fire observations:
\begin{equation}
\label{eq:5_hist_fire_4ch}
    \mathbf{A}^{h} = \left(\mathrm{AF}^{h}, \mathrm{FRP}^{h}\right),
    \quad
    \Psi_{\mathrm{fire}}\!\left(\mathrm{AF}^{h}\right) =
    \left( \mathrm{AF}^{h}_{\mathrm{mask}}, \mathrm{AF}^{h}_{\mathrm{rec}} \right),
\end{equation}
where $\mathrm{AF}^{h}_{\mathrm{mask}}$ is the fire mask derived from historical active-fire confidence, and $\mathrm{AF}^{h}_{\mathrm{rec}}$ is the normalized fire-recency feature. The raw fire-observation channels and the derived fire-history channels are then concatenated as
\begin{equation}
\label{eq:6_fire_summary_maps}
    \mathbf{A}^{h}_{\mathrm{hist}}
    =
    \left[
    \mathbf{A}^{h},
    \Psi_{\mathrm{fire}}\!\left(\mathrm{AF}^{h}\right)
    \right],
    \quad
    \mathbf{B}^{h}_{\mathrm{sum}}
    =
    \left(
    \mathbf{B}^{h}_{\mathrm{any}},
    \mathbf{B}^{h}_{\mathrm{rec}}
    \right)
    \in \mathbb{R}^{2\times H\times W}.
\end{equation}
Here, $\mathbf{B}^{h}_{\mathrm{any}}$ is a binary map indicating whether each pixel has any historical fire observation, and $\mathbf{B}^{h}_{\mathrm{rec}}$ records the normalized elapsed time since the most recent historical fire observation; pixels without historical fire are assigned a value of 1.

\section{Auxiliary Land and Fire Encoders}
\label{App_auxiliary_encoders}

The auxiliary encoders produce four-level pyramids aligned with the historical context pyramid at resolutions \(s\in\{128,64,32,16\}\). The three encoders do not share parameters: \(\Phi_{\mathrm{land}}\) is an HRNet-style encoder~\cite{Wang2021_HRNet} that keeps a \(128\times128\) high-resolution branch active while constructing
\(64\times64\), \(32\times32\), and \(16\times16\) branches, followed by cross-scale exchange blocks; \(\Phi_{\mathrm{af}}\) encodes the full historical fire sequence \(\mathbf{A}^{h}_{\mathrm{hist}}\). It combines temporal attention pooling over hourly fire features with a flattened-sequence convolutional path, thereby preserving both recent fire dynamics and aggregated historical fire evidence; \(\Phi_{\mathrm{fire}}\) is a lightweight convolutional pyramid for the two-channel fire-summary map \(\mathbf{B}^{h}_{\mathrm{sum}}\). Although \(\mathbf{B}^{h}_{\mathrm{sum}}\) has only two channels, it is lifted to the same multi-scale interface used by the other sources, which simplifies scale-wise condition injection. The subsequent gated fusion module regulates its contribution relative to other sources.

\section{Hungarian cost matrix}
\label{App_Hungarian_cost_matrix}
\begin{table}[!h]
\centering
\scriptsize
\setlength{\tabcolsep}{5.5pt}
\renewcommand{\arraystretch}{1.15}
\caption{Illustration of the Hungarian query-to-cluster cost matrix.}
\label{tab:hungarian_cost_matrix}
\begin{tabular}{lccc}
\toprule
 & Target \(g_1\) & Target \(g_2\) & Target \(g_3\) \\
\midrule
Query \(q_1\) & \(\mathbf{-0.41}\) & \(0.78\) & \(0.92\) \\
Query \(q_2\) & \(-0.20\) & \(0.10\) & \(0.65\) \\
Query \(q_3\) & \(0.55\) & \(\mathbf{-0.35}\) & \(0.50\) \\
Query \(q_4\) & \(0.48\) & \(0.44\) & \(\mathbf{-0.05}\) \\
\bottomrule
\end{tabular}
\end{table}

In \autoref{tab:hungarian_cost_matrix}, the rows are predicted queries and columns are ground-truth fire clusters. Each entry is the matching cost
\(\mathcal{D}_{qk} = \lambda_{\mathrm{m,cls}}(-\hat{p}_q) + \lambda_{\mathrm{m,loc}} \|\hat{\mathbf{y}}_q-\mathbf{g}_k\|_1\). Bold entries indicate the selected one-to-one assignment.

\section{Data preprocessing and model hyperparameter}
\label{App_data_preprocessing_model_hyperparameter}

\textbf{Preprocessing.} We restrict candidate region cubes to the latitude band from \(60^\circ\mathrm{S}\) to \(80^\circ\mathrm{N}\), excluding polar regions. Then, VIIRS records are filtered to presumed vegetation fires, rounded to hourly time slots, and rasterized into two channels: an active fire confidence code and FRP. The active-fire convention is $-1 \rightarrow \text{unobserved},\; 1/2/3 \rightarrow \text{low/nominal/high confidence fire}.$ When multiple records fall into the same pixel-hour, we keep only the highest confidence code. Then all variables are interpolated to a VIIRS \qty{375}{\meter} resolution. Continuous meteorological and land variables are converted into physically comparable units, with transformations such as Kelvin-to-Celsius conversion, pressure conversion to hPa, \(\log(1+x)\) for skewed variables, temporal differencing for accumulated fluxes, clipping for bounded land variables, and one-hot encoding for geomorphon classes. For variables whose valid physical support is sparse or sign-dependent, we additionally construct binary mask channels. Missing source values and provider-specific fill values are first standardized, and non-fire covariate gaps are filled with 0. The z-score normalization statistics used by the learning pipeline, mean and std, are estimated from the training dataset only. 

\begin{figure}[!h]
    \centering
    \includegraphics[width=0.9\linewidth]{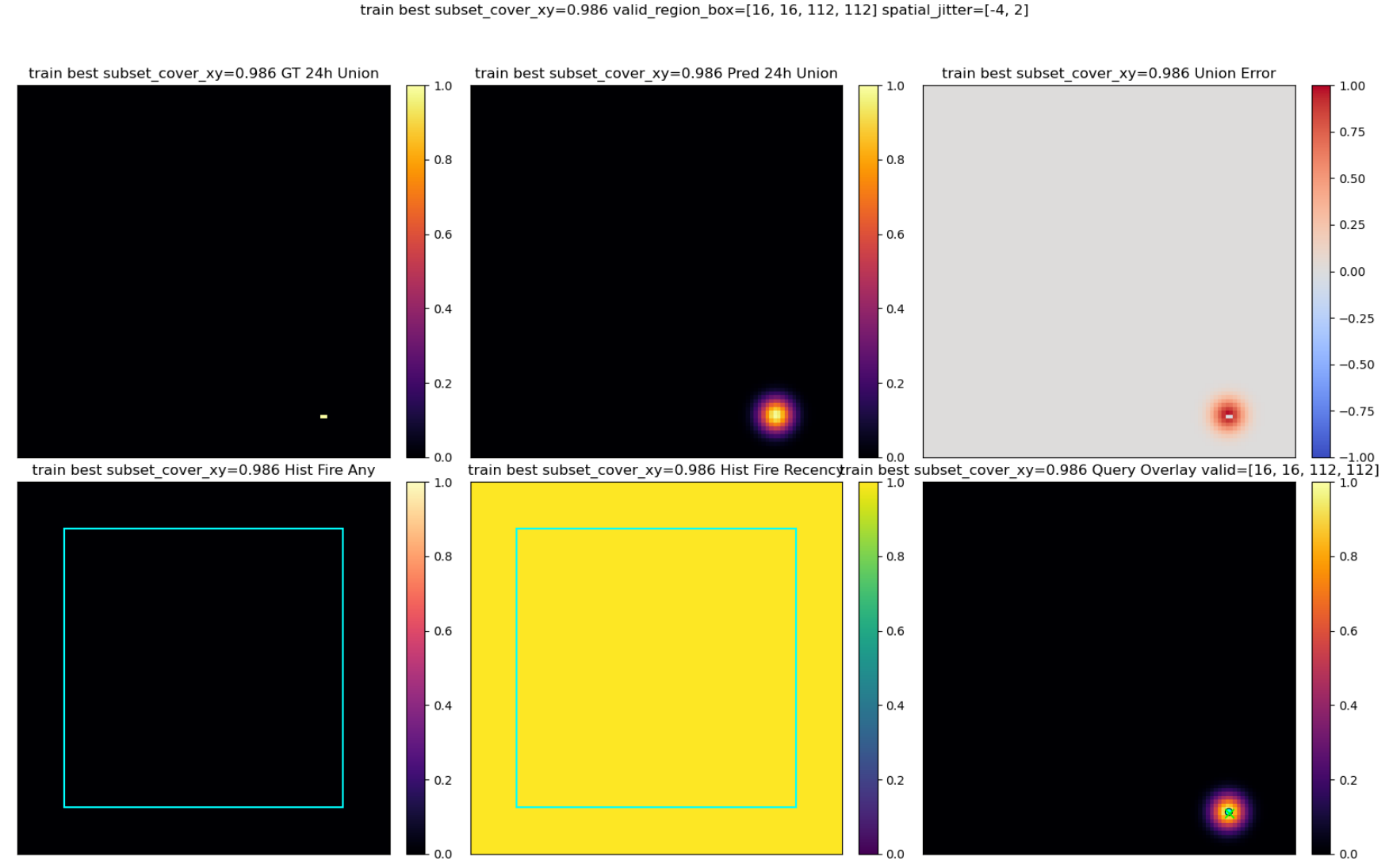}
    \caption{The valid-region and spatial-jitter diagnostic during training process.
    \label{App:fig_jitter_valid_region}}
\end{figure}

\begin{figure}[t]
    \centering
    \includegraphics[width=1\linewidth]{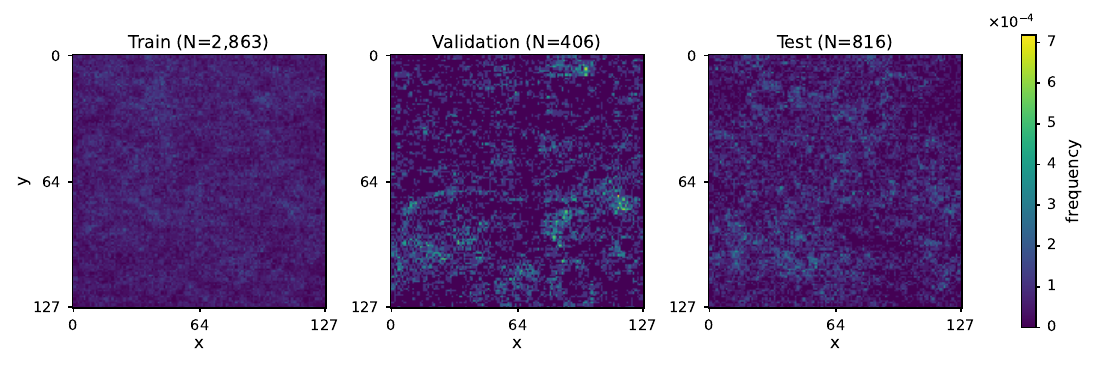}
    \caption{
    Burning-pixel frequency maps for the train, validation, and test splits.
    For each split, the frequency at each \(128\times128\) pixel is computed
    over the future 24-hour target windows. All panels use the same color scale,
    so spatial concentration and split-level sampling differences can be compared
    directly.
    \label{App:fig_burning_pixel_frequency}}
\end{figure}

\textbf{Data augmentation.}
During training, WISP applies a random spatial jitter to the full \(128\times128\) input entity before target construction. The model still receives the full context window, but supervision and evaluation are restricted
to the central valid region \(\texttt{valid\_region\_box}=[16,16,112,112]\), corresponding to a \(96\times96\) crop (see cyan box in \autoref{App:fig_jitter_valid_region}). This design keeps contextual information around the target area while excluding boundary regions where shifted inputs may contain padding or incomplete neighbourhood context. It also prevents the model from exploiting a fixed sampling prior in which future fires always appear near the centre of the input window or some fixed locations (see~\autoref{App:fig_burning_pixel_frequency}). The sampled offset is recorded as \(\texttt{spatial\_jitter}=[dy,dx]\); for example, \([-4,2]\) shifts the entity by 4 pixels upward and 2 pixels to the right in \autoref{App:fig_jitter_valid_region}. The displayed \(\texttt{subset\_cover\_xy}=0.986\) is a structure-aware diagnostic used to select the best training snapshot: it measures how well a one-to-one subset of positive queries covers the ground-truth fire-cluster centres. Higher is better.

\textbf{Training configuration.}
All model variants are trained with AdamW, learning rate \(10^{-4}\), and weight decay \(10^{-4}\). Training is distributed with Accelerate on 4 nodes with 4 A100 GPUs per node, for 16 GPUs in total. We use bfloat16 mixed precision and activation checkpointing. The per-GPU micro-batch size is 1; since \(\texttt{grad\_accum\_steps}=4\) by default, the effective batch size is \(1\times16\times4=64\). The model uses dropout with probability 0.1 in the spatio-temporal attention backbone and query decoder. Each branch is trained for up to 5000 epochs, and the checkpoint with the best validation event mAP over AP@7, AP@14, and AP@21 is selected for test evaluation. The operating threshold for threshold-dependent query metrics is 0.5 unless otherwise specified.

\textbf{Set-prediction hyperparameters.}
The default query budget is $Q=10$, with ablations using $Q=20$ and $Q=50$. The query decoder uses two reference-refinement layers. Hungarian matching uses classification and localization weights $(\lambda_{\mathrm{m,cls}},\lambda_{\mathrm{m,loc}})$ as reported in the main tables (see \autoref{tab:1_main_detection_coverage_results} and \autoref{tab:2_main_query_diagnostic_results}). The default configuration uses $(1,2)$. The localization loss weight is $\lambda_{\mathrm{loc}}=5$, and the no-fire class weight in cross entropy is $w_{\mathrm{eos}}=0.1$. Query predictions are rendered into an auxiliary 24-hour union probability map with Gaussian kernels of $\sigma=3$ pixels; this rendered map is used for diagnostics and raster-level metrics, but it is not directly supervised by the training loss.

\textbf{Future union and cluster targets.}
For supervision, future active-fire observations are restricted to the central valid region. A 24-hour union fire mask is constructed by marking a pixel positive if it is a burning pixel at any future hour. Set targets are obtained by grouping the union mask into connected components using a $7\times7$ footprint, corresponding to a Chebyshev radius of 3 pixels. For each component, we compute an FRP-weighted centre; if the component has zero total FRP, we use the unweighted centroid. Components are ranked by total FRP mass, then by component size, and then by spatial position for deterministic tie-breaking. During training, only the first $Q$ components are used as set targets; the full target set is retained for validation diagnostics and for reporting ground-truth truncation.

\section{Evaluation Metrics}
\label{App_eval_metrics}

WISP predicts a fixed set of query outputs
\(\hat{\mathcal{Y}}=\{(\hat{p}_q,\hat{\mathbf{y}}_q)\}_{q=1}^{Q}\), where
\(\hat{p}_q\) is the fire probability and \(\hat{\mathbf{y}}_q\in[0,1]^2\) is a
normalized point location. Ground truth is a set of fire-cluster centres
\(\mathcal{G}=\{(\mathbf{g}_k,m_k)\}_{k=1}^{K}\), where \(m_k\) is the FRP mass
of cluster \(k\). Distances are computed in pixel units on the valid VIIRS grid.

\textbf{Event AP.}
For a spatial tolerance \(r\), predictions are sorted by decreasing
\(\hat{p}_q\). A prediction is counted as a true positive if it lies within
\(r\) pixels of an unmatched ground-truth centre; otherwise it is a false
positive. This gives a precision--recall curve over all query predictions in the
dataset, and AP@\(r\) is the area under this curve. We report AP@7, AP@14, and
AP@21, and define
\[
\mathrm{mAP}
=
\frac{1}{3}
\left(
\mathrm{AP@7}+\mathrm{AP@14}+\mathrm{AP@21}
\right).
\]

\textbf{Mass coverage.}
At the operating threshold \(\hat{p}_q\geq0.5\), positive queries are matched
one-to-one to ground-truth clusters by spatial similarity. Let \(\pi\) denote the
matched query--cluster pairs, and let \(d_{qk}\) be the pixel distance between
query \(q\) and cluster \(k\). The FRP-mass coverage at radius \(r\) is
\[
\mathrm{MassCov@}r
=
\frac{
\sum_{(q,k)\in\pi}
m_k \, \mathbb{I}[d_{qk}\leq r]
}{
\sum_{j=1}^{|\pi|} m_{(j)}
},
\]
where \(m_{(j)}\) are the ground-truth cluster masses sorted in decreasing order.
This denominator caps the metric by the number of active predictions, so the
score measures whether the predicted query set covers the most important
available fire-cluster mass.

\textbf{Hit rate.}
Hit@\(r\) is the fraction of matched query--cluster pairs that fall within
radius \(r\):
\[
\mathrm{Hit@}r
=
\frac{1}{|\pi|}
\sum_{(q,k)\in\pi}
\mathbb{I}[d_{qk}\leq r].
\]
We use Hit@14 as the primary matched-query localization diagnostic.

\textbf{Union AUROC.}
For raster diagnostics, query points are rendered into a 24-hour union
probability map by placing Gaussian kernels at predicted query locations. Union
AUROC is the standard pixel-level ROC AUC between this rendered probability map
and the binary 24-hour future fire union mask over the valid region. This metric
is reported only as an auxiliary diagnostic, since WISP is trained and evaluated
primarily as a point-set predictor.

\section{Query Evolution During Training}
\label{App_validation_query_evolution}

\begin{figure}[!h]
    \centering
    \includegraphics[
        width=\textwidth,
        height=0.82\textheight,
        keepaspectratio]{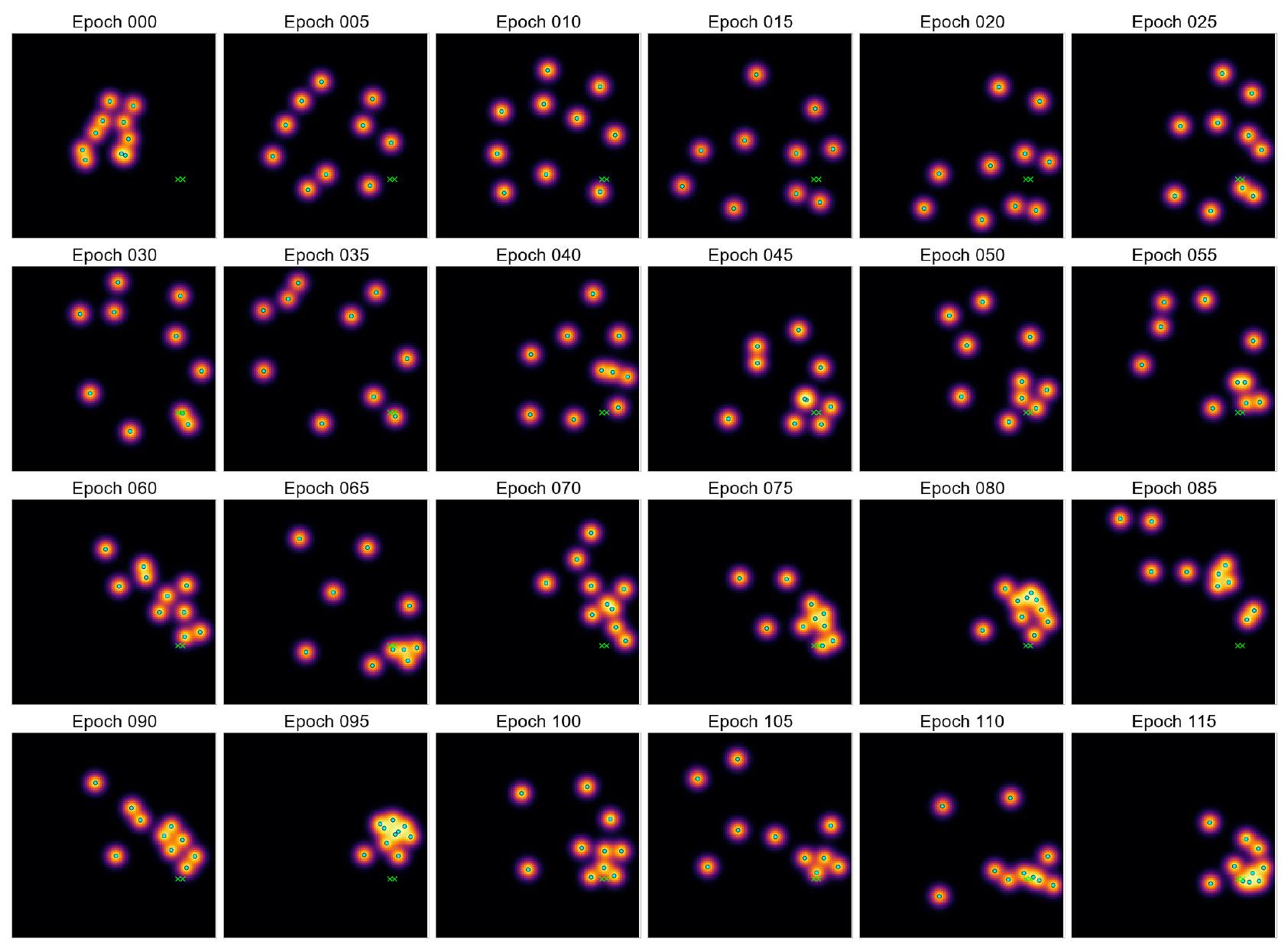}

    \caption{Evolution of WISP query predictions on the first validation entity during the first 120 training epochs. The panels show the query overlay every five epochs, from epoch 0 to epoch 115. Cyan circles denote predicted fire queries and lime crosses denote ground-truth fire-cluster centres. Over training, the active query locations increasingly concentrate around the ground-truth future fire location. This visualization is used only as a qualitative diagnostic of training dynamics.
    \label{fig:app_validation_query_evolution}
    }
\end{figure}

\section{Radius-Wise Test Metrics}
\label{App_radius_metrics}

\begin{table*}[!h]
\centering
\tiny
\setlength{\tabcolsep}{2.4pt}
\renewcommand{\arraystretch}{0.92}
\caption{
Radius-wise test-set event and query diagnostics. All values are reported in \%.
The radius \(r\) is measured in VIIRS pixels. Higher is better except LRP.
Best value at each radius is in bold.
}
\label{tab:app_radius_wise_metrics}
\resizebox{\textwidth}{!}{
\begin{tabular}{lccc crrrrrrrrr}
\toprule
WISP & $Q$ & $\lambda_{\mathrm{m,cls}}$ & $\lambda_{\mathrm{m,loc}}$
& $r$ & AP & Hit & ClusPrec & ClusRec & ClusF1 & Top10Rec & WCAP & wWCAP & LRP $\downarrow$ \\
\midrule
v1 & 10 & 0 & 1 & 7  & 6.4 & 24.9 & 9.3 & 20.1 & 13.8 & 25.8 & 20.4 & 21.8 & \textbf{46.0} \\
   &    &   &   & 14 & 23.0 & 44.6 & 18.0 & 36.2 & 26.1 & 48.8 & 40.6 & 41.2 & \textbf{42.3} \\
   &    &   &   & 21 & 33.4 & 54.1 & 24.0 & 45.9 & 34.4 & 64.6 & 50.4 & 51.2 & \textbf{40.2} \\
\addlinespace[1pt]
v2 & 10 & 1 & 1 & 7  & 4.4 & 16.6 & 6.4 & 11.5 & 9.0 & 20.8 & 14.2 & 15.2 & 58.0 \\
   &    &   &   & 14 & 21.3 & 41.3 & 18.1 & 32.4 & 25.6 & 55.4 & 41.2 & 42.1 & 49.9 \\
   &    &   &   & 21 & 36.6 & 59.9 & 29.3 & 52.1 & 41.7 & \textbf{81.4} & 61.3 & 62.1 & 43.3 \\
\addlinespace[1pt]
v3 & 10 & 1 & 2 & 7  & 5.5 & 22.8 & \textbf{12.7} & 15.8 & 15.7 & \textbf{27.2} & 19.0 & 19.6 & 55.8 \\
   &    &   &   & 14 & 22.4 & 41.5 & \textbf{24.8} & 32.1 & 31.1 & 57.4 & 40.3 & 41.5 & 48.2 \\
   &    &   &   & 21 & 36.1 & 54.6 & \textbf{34.5} & 45.9 & \textbf{43.9} & 78.5 & 54.3 & 55.2 & 42.7 \\
\addlinespace[1pt]
v4 & 10 & 1 & 5 & 7  & 5.7 & 23.1 & 10.9 & 17.5 & 14.6 & 25.6 & 19.3 & 20.0 & 55.2 \\
   &    &   &   & 14 & 22.7 & 44.0 & 21.4 & 35.9 & 29.5 & \textbf{58.2} & 41.8 & 43.3 & 48.8 \\
   &    &   &   & 21 & 34.8 & 55.3 & 29.6 & 47.5 & 40.2 & 78.6 & 54.1 & 55.0 & 44.7 \\
\addlinespace[1pt]
v5 & 20 & 1 & 2 & 7  & 8.7 & 22.6 & 10.0 & 18.0 & 14.6 & 26.1 & 21.4 & 22.0 & 59.0 \\
   &    &   &   & 14 & 30.0 & 45.4 & 22.4 & 40.7 & 32.5 & 53.1 & 46.0 & 45.5 & 49.8 \\
   &    &   &   & 21 & 42.0 & 55.4 & 28.6 & 50.6 & 41.0 & 68.4 & 56.8 & 56.4 & 45.2 \\
\addlinespace[1pt]
v6 & 50 & 1 & 2 & 7  & \textbf{12.8} & \textbf{28.6} & 10.8 & \textbf{24.1} & \textbf{16.1} & 26.3 & \textbf{28.2} & \textbf{28.5} & 54.6 \\
   &    &   &   & 14 & \textbf{38.2} & \textbf{54.1} & 21.7 & \textbf{49.6} & \textbf{32.7} & 57.0 & \textbf{57.0} & \textbf{57.4} & 45.7 \\
   &    &   &   & 21 & \textbf{50.0} & \textbf{64.0} & 26.9 & \textbf{60.5} & 40.3 & 72.5 & \textbf{65.8} & \textbf{66.0} & 41.4 \\
\bottomrule
\end{tabular}
}
\end{table*}

\section{Regime-wise Analysis for train dataset}
\label{App_train_regime}

\begin{table*}[!h]
\centering
\caption{Regime-wise operational behaviour on the training dataset. Hit@14, AP@14, Rec@14, and Mean prob. are reported in \%.}
\label{tab:regime_wise_operational_train}
\resizebox{\textwidth}{!}{
\begin{tabular}{l ccc ccc cc cc}
\toprule
& \multicolumn{3}{c}{New ignition $(n=504)$}
& \multicolumn{3}{c}{Continued fire $(n=1415)$}
& \multicolumn{2}{c}{Extinguished $(n=138)$}
& \multicolumn{2}{c}{Quiescent $(n=806)$} \\
\cmidrule(lr){2-4}\cmidrule(lr){5-7}\cmidrule(lr){8-9}\cmidrule(lr){10-11}
WISP & Hit@14 & AP@14 & Rec@14 & Hit@14 & AP@14 & Rec@14 
& Avg. pred. $\downarrow$& Mean prob. $\downarrow$& Avg. pred. $\downarrow$& Mean prob. $\downarrow$\\
\midrule
v1 & 87.2 & 39.2 & 75.7 & 86.9 & 43.4 & 79.9 & \textbf{0.20} & 0.39 & \textbf{0.10} & 0.15 \\
v2 & 38.4 & 16.6 & 34.6 & 56.6 & 27.9 & 49.5 & 1.63 & 1.15 & 0.40 & 0.48 \\
v3 & 87.4 & 69.1 & 84.9 & 77.9 & 42.6 & 73.5 & 0.45 & 0.32 & 0.15 & \textbf{0.10} \\
v4 & 92.9 & 71.8 & 91.4 & 84.8 & 44.7 & 81.4 & 0.41 & \textbf{0.28} & 0.12 & 0.10 \\
v5 & \textbf{96.5} & \textbf{84.9} & \textbf{96.2} & \textbf{90.9} & \textbf{64.5} & \textbf{89.9} & 0.68 & 0.44 & 0.19 & 0.12 \\
v6 & 66.7 & 38.7 & 62.9 & 78.1 & 56.4 & 75.6 & 2.20 & 1.88 & 0.38 & 0.61 \\
\bottomrule
\end{tabular}
}
\end{table*}

\section{Hungarian Matching Algorithm}
\label{App_algo_Hungarian}

\begin{algorithm}[!h]
\caption{Hungarian matching WISP}
\label{alg:wisp_hungarian_loss}
\begin{algorithmic}[1]
\Require Predicted query set 
$\hat{\mathcal{Y}}=\{(\mathbf{z}_q,\hat{\mathbf{y}}_q)\}_{q=1}^{Q}$;
ground-truth fire-cluster set 
$\mathcal{G}_{K}=\{\mathbf{g}_k\}_{k=1}^{K}$;
matching weights $\lambda_{\mathrm{m,cls}},\lambda_{\mathrm{m,loc}}$;
localization weight $\lambda_{\mathrm{loc}}$;
no-fire class weight $w_{\mathrm{eos}}$.
\Ensure Set prediction loss $\mathcal{L}$.

\State Compute fire probabilities
$\hat{p}_q \gets \operatorname{softmax}(\mathbf{z}_q)_1$ for all $q=1,\ldots,Q$.

\If{$K > 0$}
    \For{$q=1,\ldots,Q$}
        \For{$k=1,\ldots,K$}
            \State Compute matching cost
            \[
            \mathcal{D}_{qk}
            \gets
            \lambda_{\mathrm{m,cls}}(-\hat{p}_q)
            +
            \lambda_{\mathrm{m,loc}}
            \left\|\hat{\mathbf{y}}_q-\mathbf{g}_k\right\|_1 .
            \]
        \EndFor
    \EndFor
    \State Solve the linear assignment problem
    \[
    \pi^\star
    \gets
    \operatorname{Hungarian}(\mathcal{D}),
    \]
    where $\pi^\star$ contains one-to-one query--cluster pairs.
\Else
    \State Set $\pi^\star \gets \emptyset$.
\EndIf

\State Initialize query labels $c_q^\star \gets 0$ for all $q=1,\ldots,Q$.
\ForAll{$(q,k)\in\pi^\star$}
    \State Set $c_q^\star \gets 1$.
\EndFor

\State Compute classification loss over all queries:
\[
\mathcal{L}_{\mathrm{cls}}
\gets
\frac{1}{Q}
\sum_{q=1}^{Q}
\operatorname{CE}
\left(
\mathbf{z}_q,
c_q^\star;
\{w_{\mathrm{eos}},1\}
\right).
\]

\If{$|\pi^\star| > 0$}
    \State Compute localization loss over matched queries:
    \[
    \mathcal{L}_{\mathrm{loc}}
    \gets
    \frac{1}{2|\pi^\star|}
    \sum_{(q,k)\in\pi^\star}
    \left\|
    \hat{\mathbf{y}}_q-\mathbf{g}_k
    \right\|_1 .
    \]
\Else
    \State Set $\mathcal{L}_{\mathrm{loc}}\gets 0$.
\EndIf

\State Combine losses:
\[
\mathcal{L} \gets \mathcal{L}_{\mathrm{cls}} + \lambda_{\mathrm{loc}}\mathcal{L}_{\mathrm{loc}}.
\]
\State \Return $\mathcal{L}$.
\end{algorithmic}
\end{algorithm}




\newcommand{\rowlabel}[1]{%
    \makebox[1.3em][c]{%
        \raisebox{1.35\height}{\rotatebox{90}{\scriptsize\bfseries #1}}%
    }%
}

\newcommand{\qualgallery}[2]{%
{\centering
    \setlength{\tabcolsep}{1pt}%
    \renewcommand{\arraystretch}{0.25}%
    \begin{tabular}{c c c c}
        \rowlabel{Good} &
        \includegraphics[width=0.305\textwidth]{#1_good_1.pdf} &
        \includegraphics[width=0.305\textwidth]{#1_good_2.pdf} &
        \includegraphics[width=0.305\textwidth]{#1_good_3.pdf} \\[-0.6em]

        \rowlabel{Median} &
        \includegraphics[width=0.305\textwidth]{#1_medium_1.pdf} &
        \includegraphics[width=0.305\textwidth]{#1_medium_2.pdf} &
        \includegraphics[width=0.305\textwidth]{#1_medium_3.pdf} \\[-0.6em]

        \rowlabel{Failure} &
        \includegraphics[width=0.305\textwidth]{#1_bad_1.pdf} &
        \includegraphics[width=0.305\textwidth]{#1_bad_2.pdf} &
        \includegraphics[width=0.305\textwidth]{#1_bad_3.pdf}
    \end{tabular}
    \captionof{figure}{Additional qualitative gallery for WISP variant #2.}
    \label{fig:app_qualitative_#2}
\par}%
}
\clearpage

\clearpage

\section{Additional Qualitative Results}
\label{App_qualitative_gallery}

\noindent
In all appendix galleries, rows correspond to good, median, and failure cases from top to bottom, and columns show three examples within the same case group.

\begin{center}
\setlength{\tabcolsep}{1pt}
\renewcommand{\arraystretch}{0.25}
\begin{tabular}{c c c c}
    \makebox[1.3em][c]{\raisebox{1.35\height}{\rotatebox{90}{\scriptsize\bfseries Good}}} &
    \includegraphics[width=0.305\textwidth]{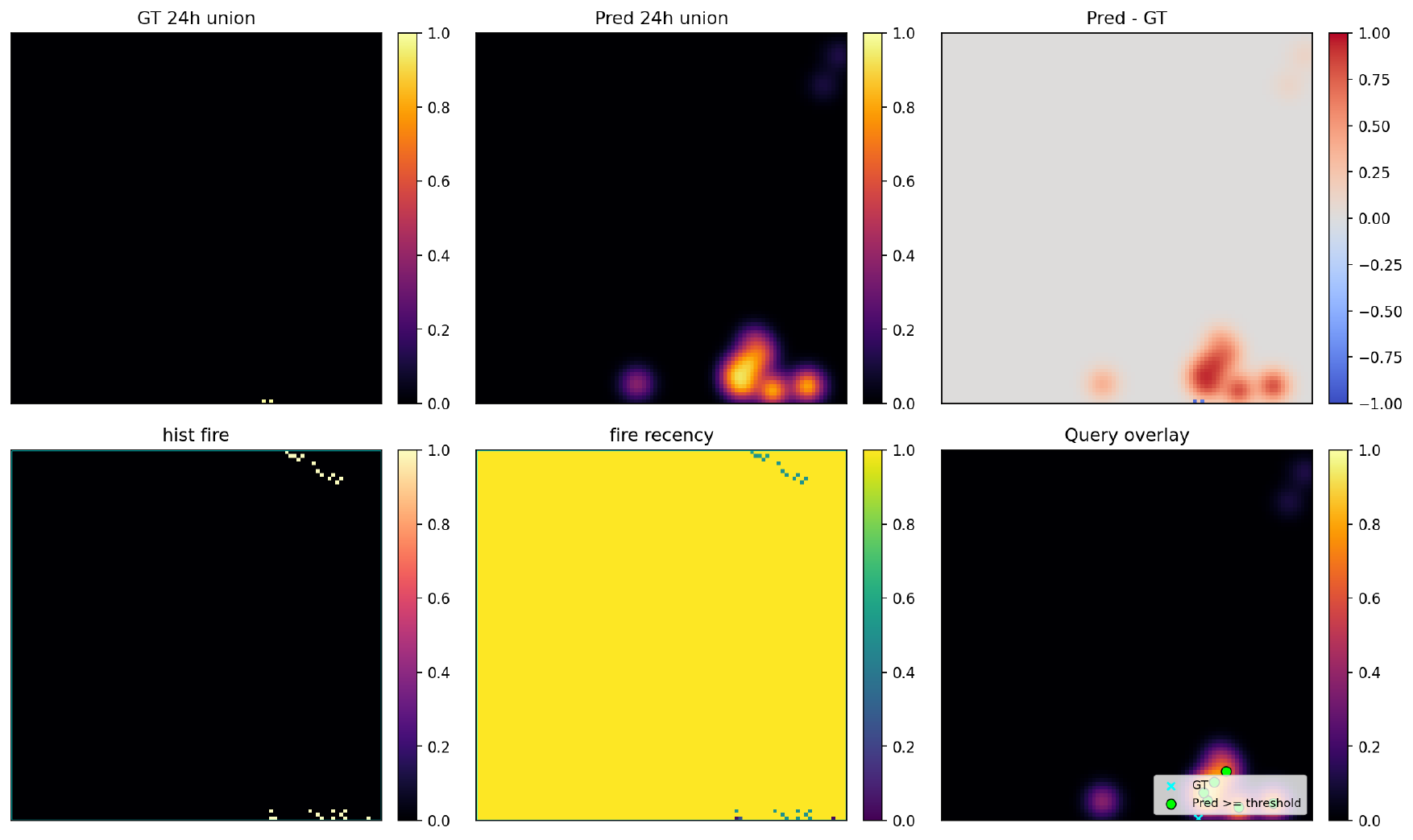} &
    \includegraphics[width=0.305\textwidth]{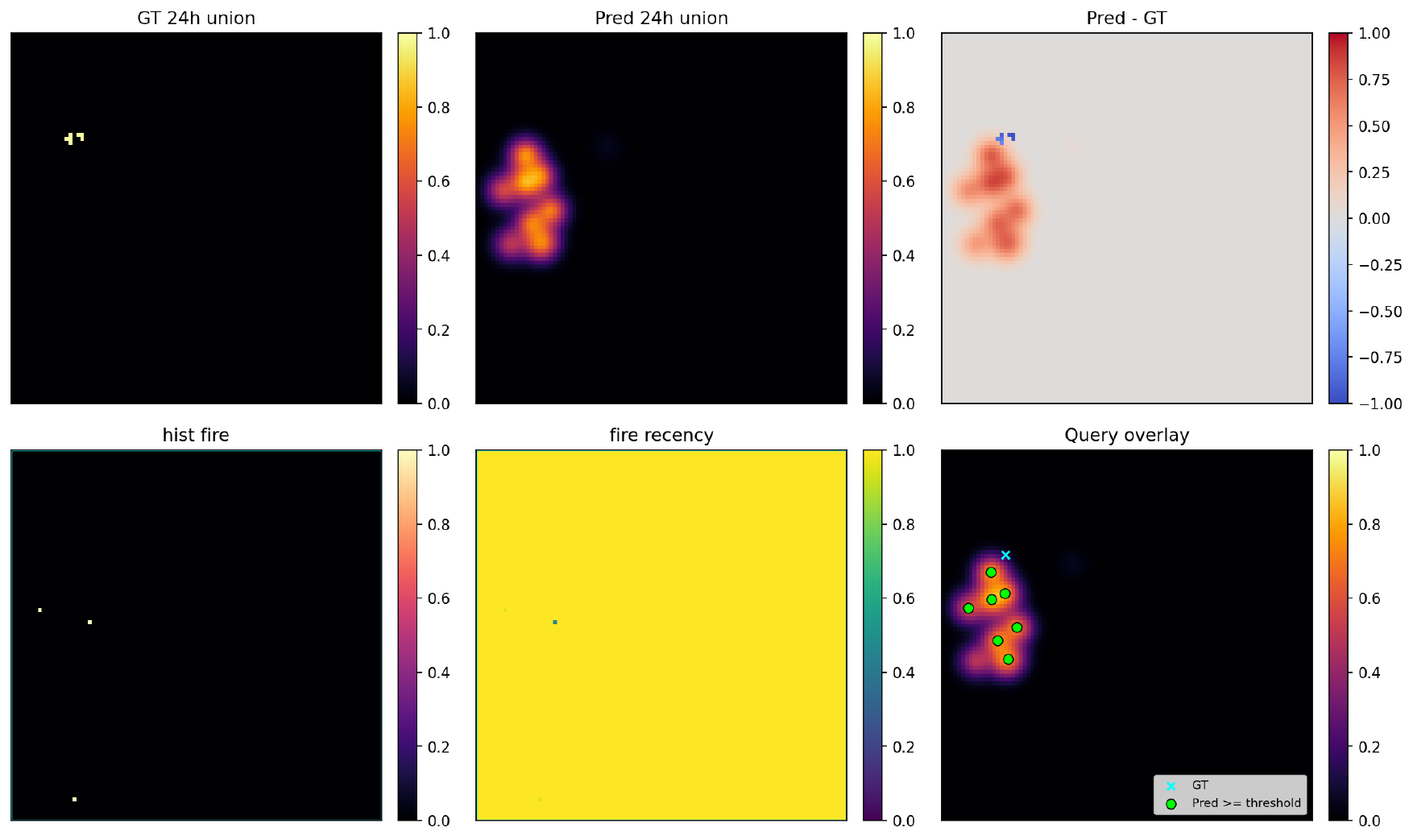} &
    \includegraphics[width=0.305\textwidth]{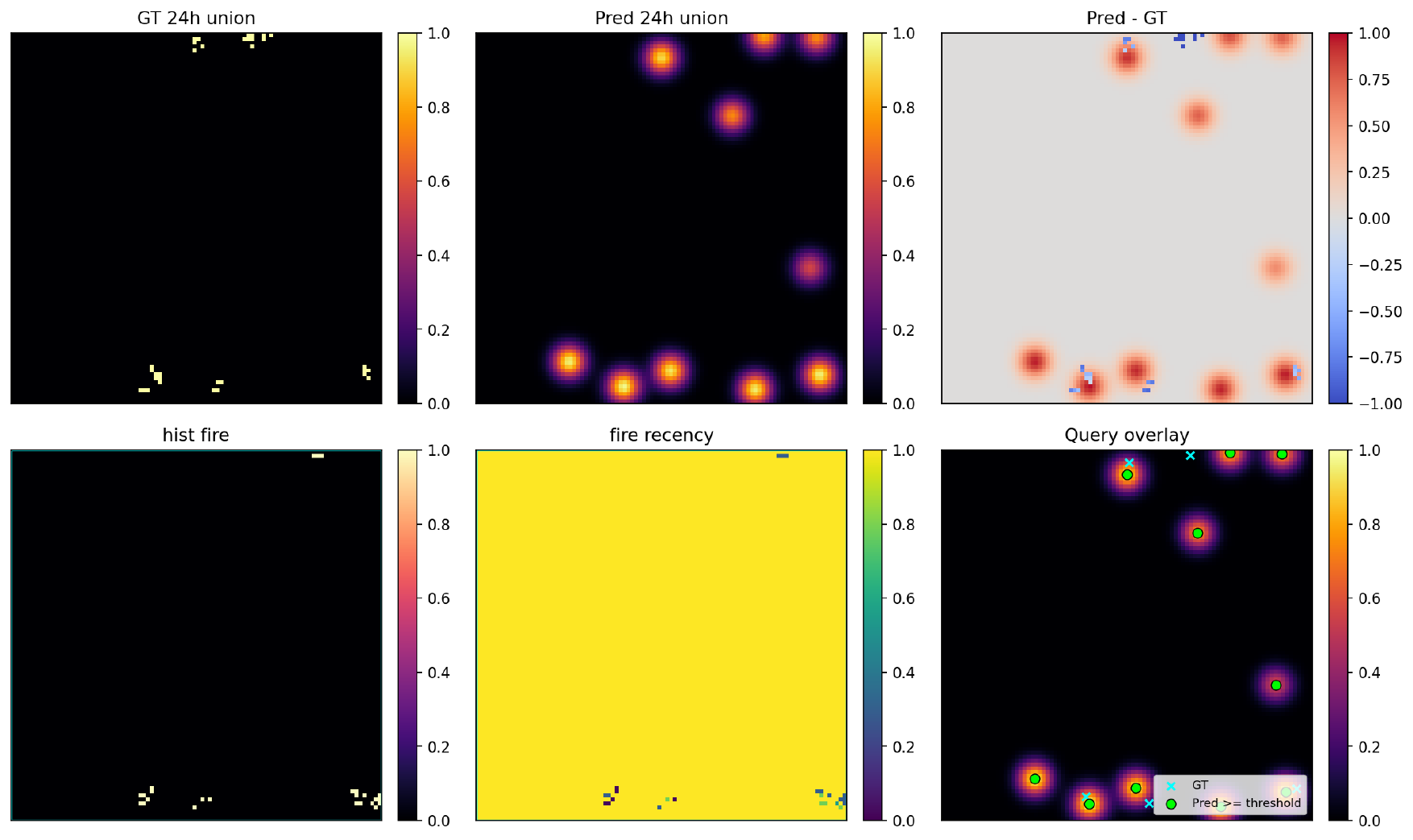} \\[-0.6em]

    \makebox[1.3em][c]{\raisebox{1.35\height}{\rotatebox{90}{\scriptsize\bfseries Median}}} &
    \includegraphics[width=0.305\textwidth]{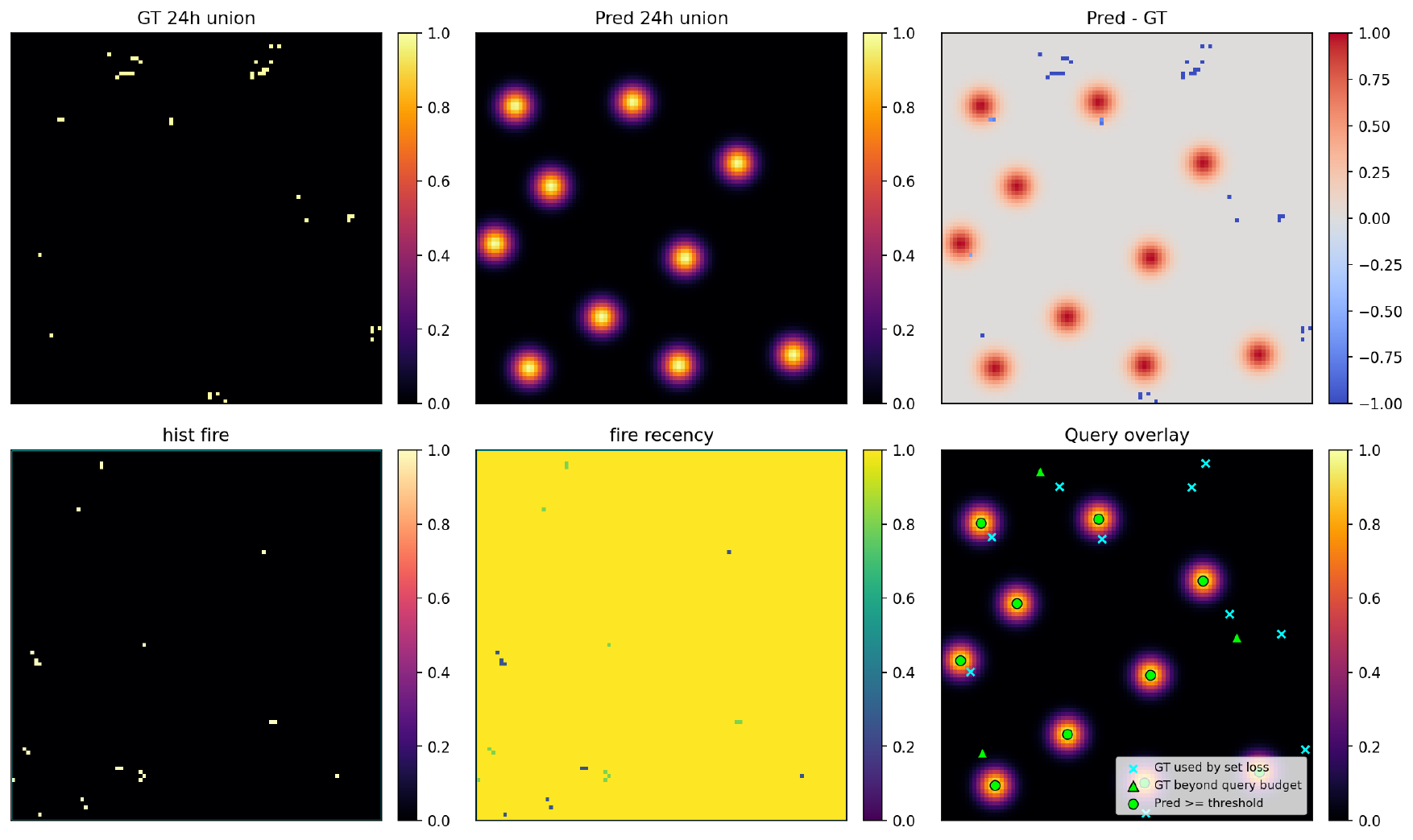} &
    \includegraphics[width=0.305\textwidth]{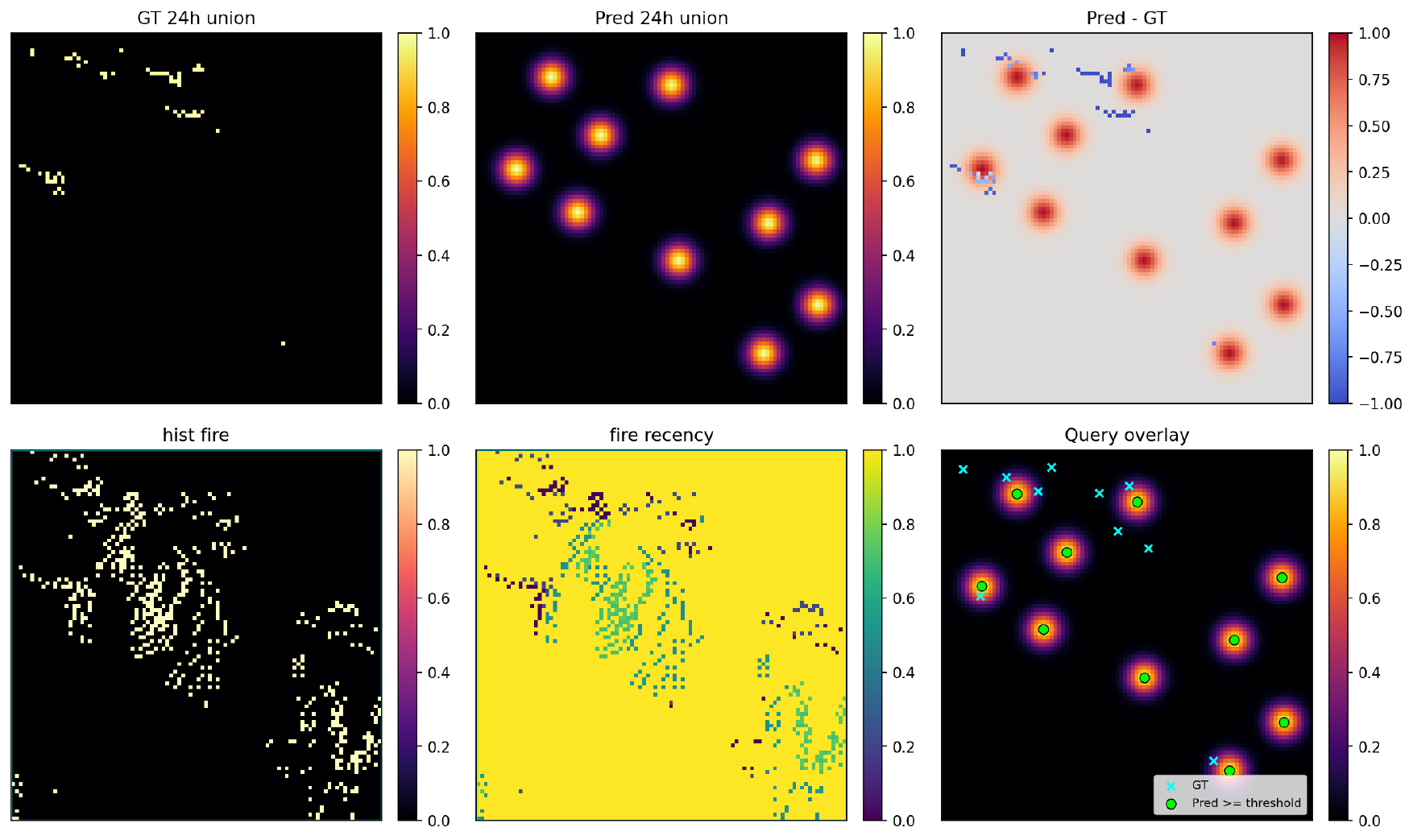} &
    \includegraphics[width=0.305\textwidth]{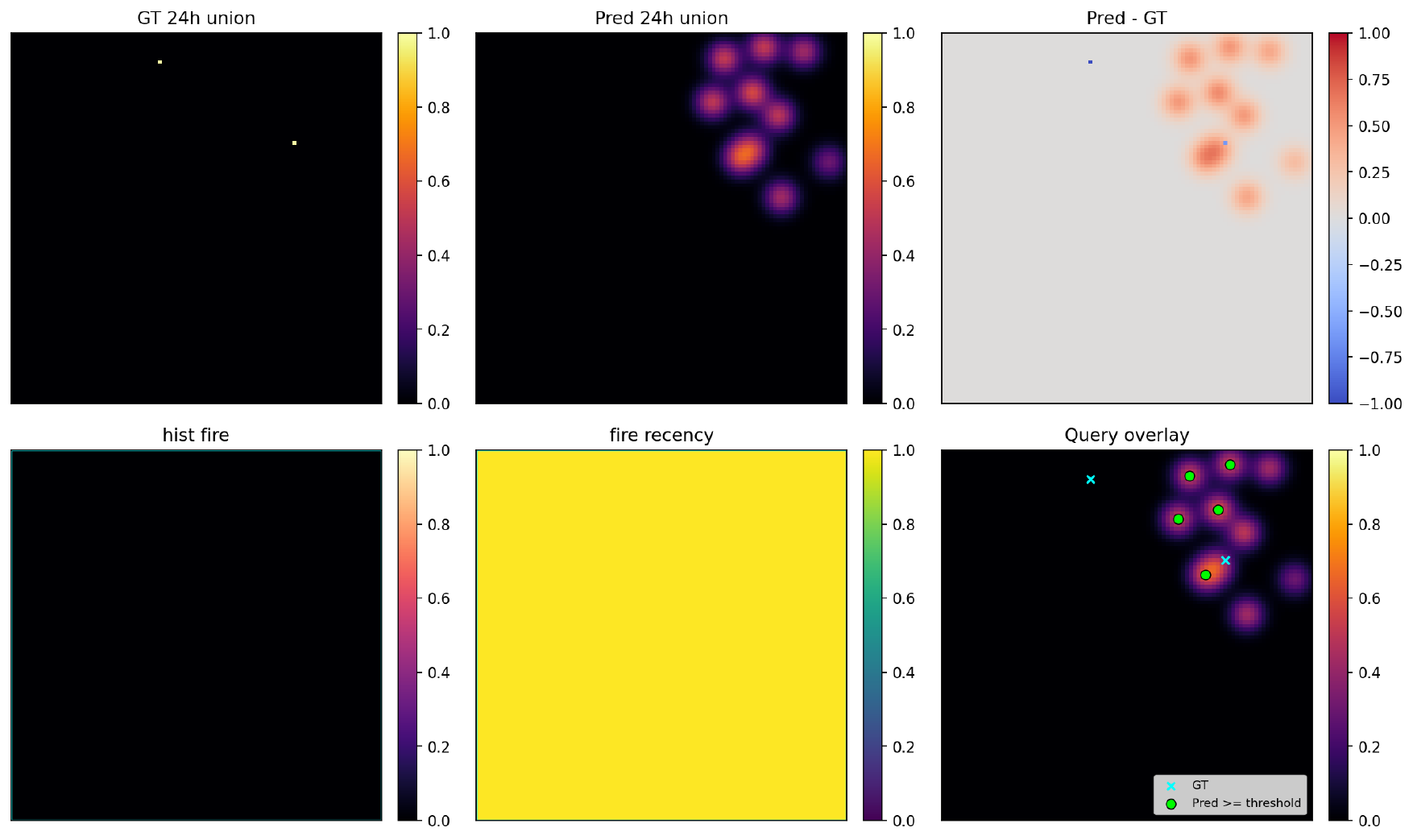} \\[-0.6em]

    \makebox[1.3em][c]{\raisebox{1.35\height}{\rotatebox{90}{\scriptsize\bfseries Failure}}} &
    \includegraphics[width=0.305\textwidth]{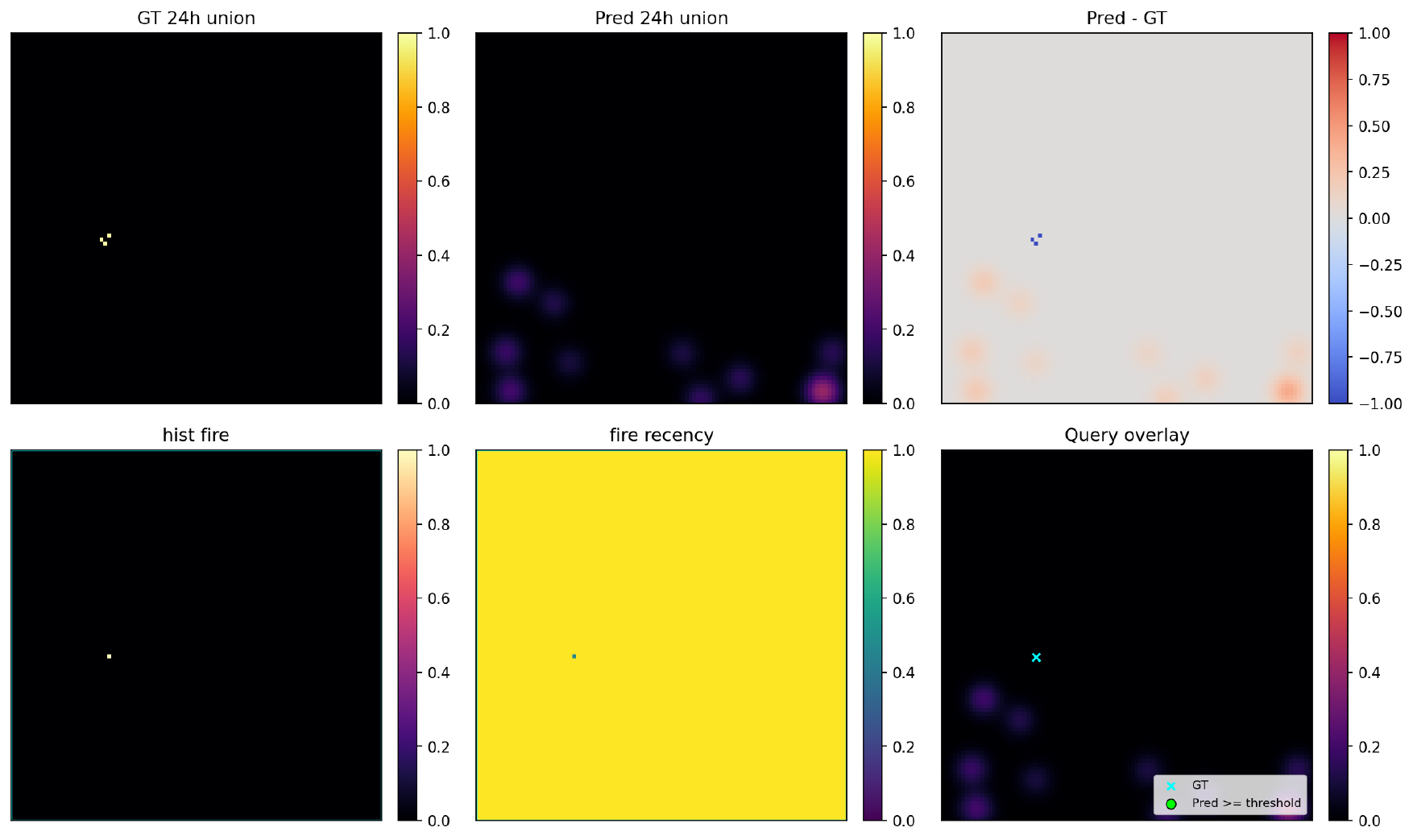} &
    \includegraphics[width=0.305\textwidth]{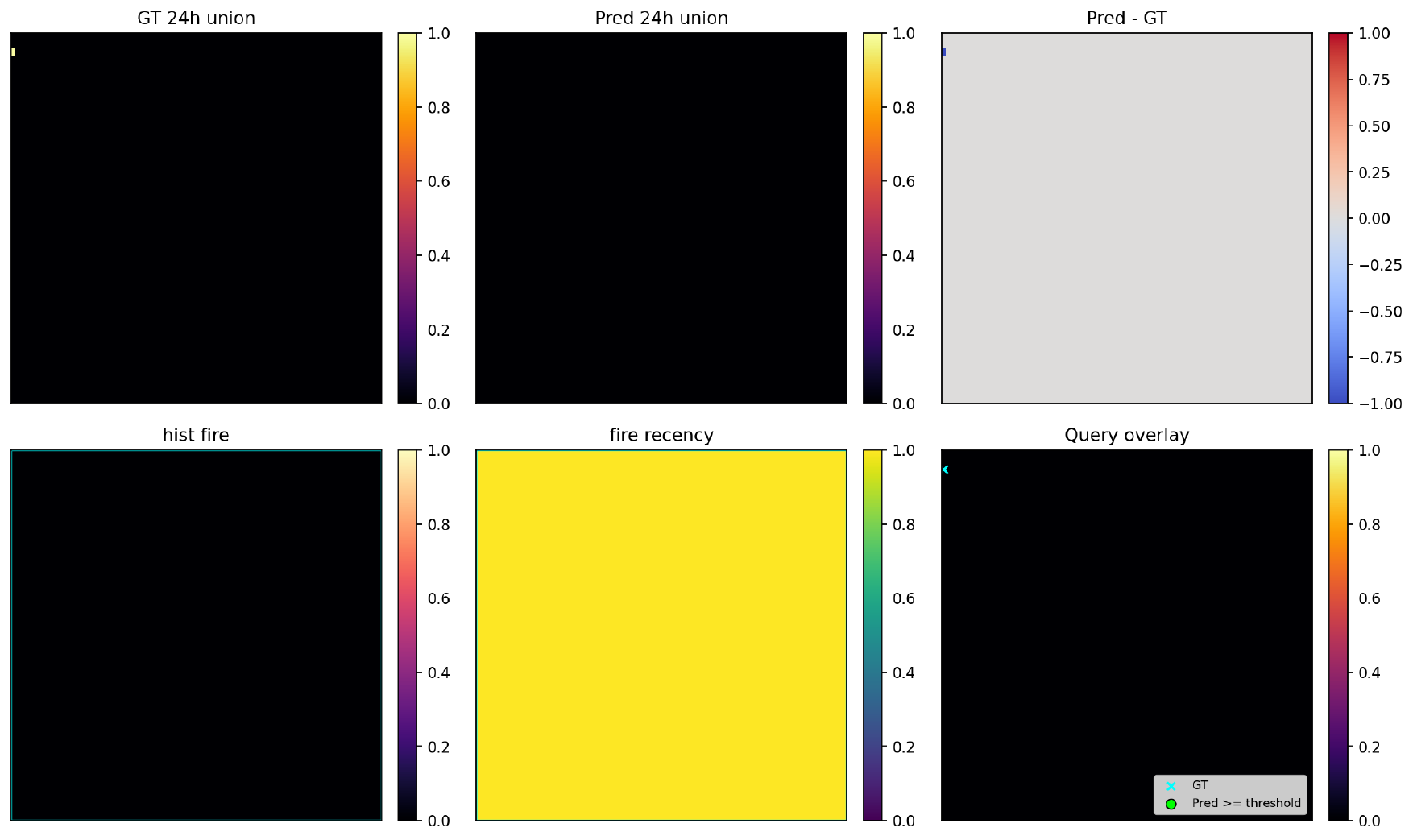} &
    \includegraphics[width=0.305\textwidth]{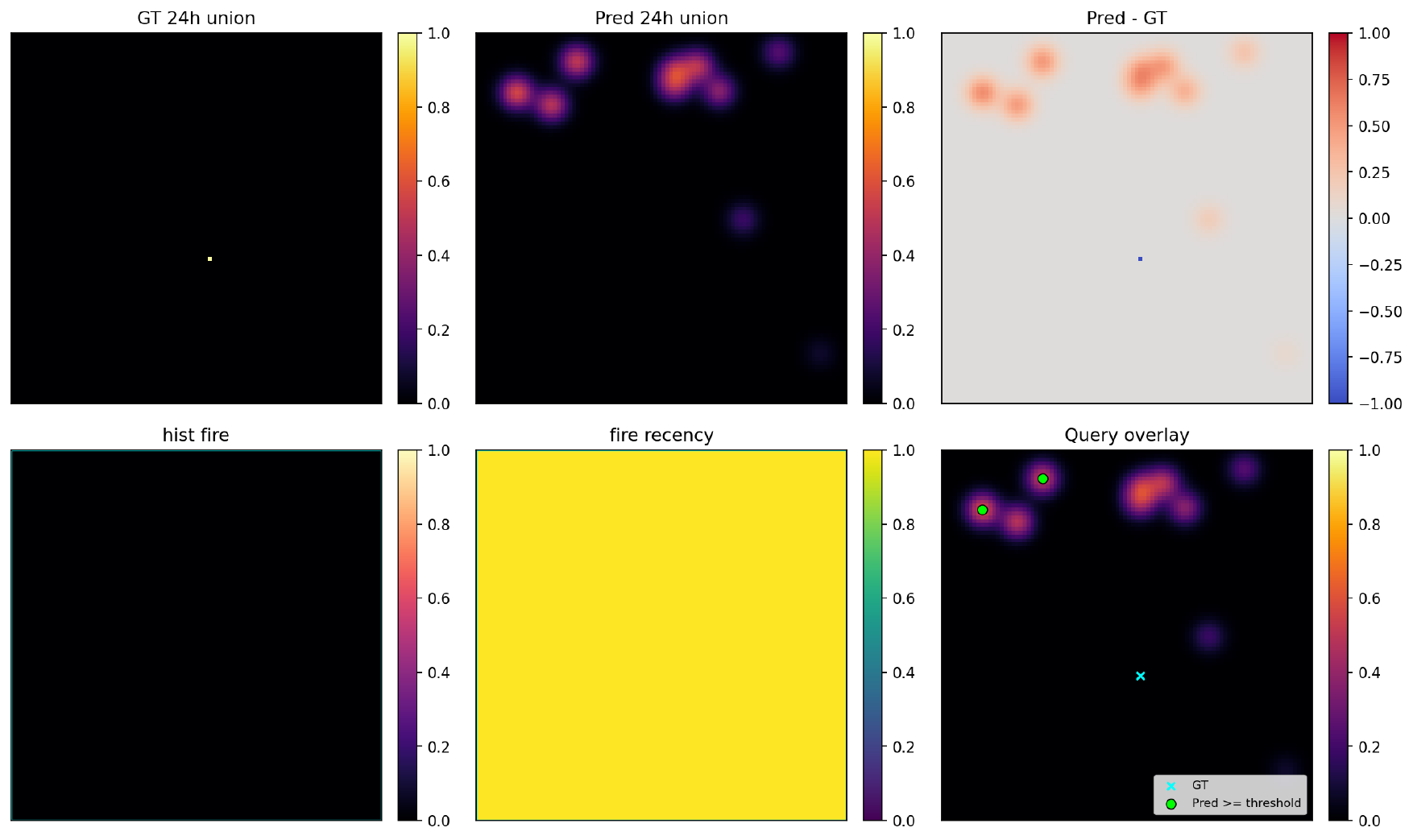}
\end{tabular}
\captionof{figure}{Additional qualitative gallery for WISP variant v1.}
\label{fig:app_qualitative_v1}
\end{center}

\begin{center}
\setlength{\tabcolsep}{1pt}
\renewcommand{\arraystretch}{0.25}
\begin{tabular}{c c c c}
    \makebox[1.3em][c]{\raisebox{1.35\height}{\rotatebox{90}{\scriptsize\bfseries Good}}} &
    \includegraphics[width=0.305\textwidth]{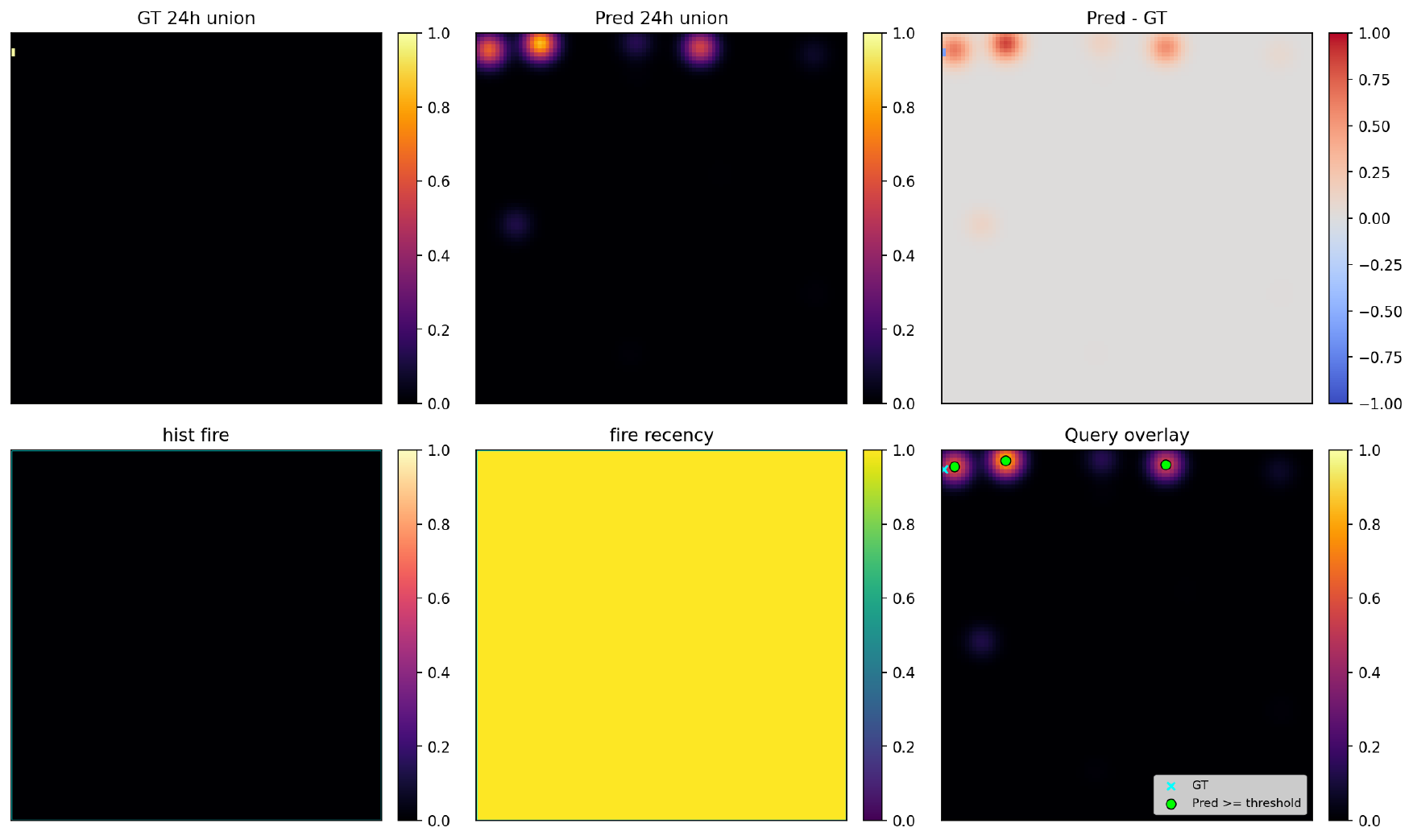} &
    \includegraphics[width=0.305\textwidth]{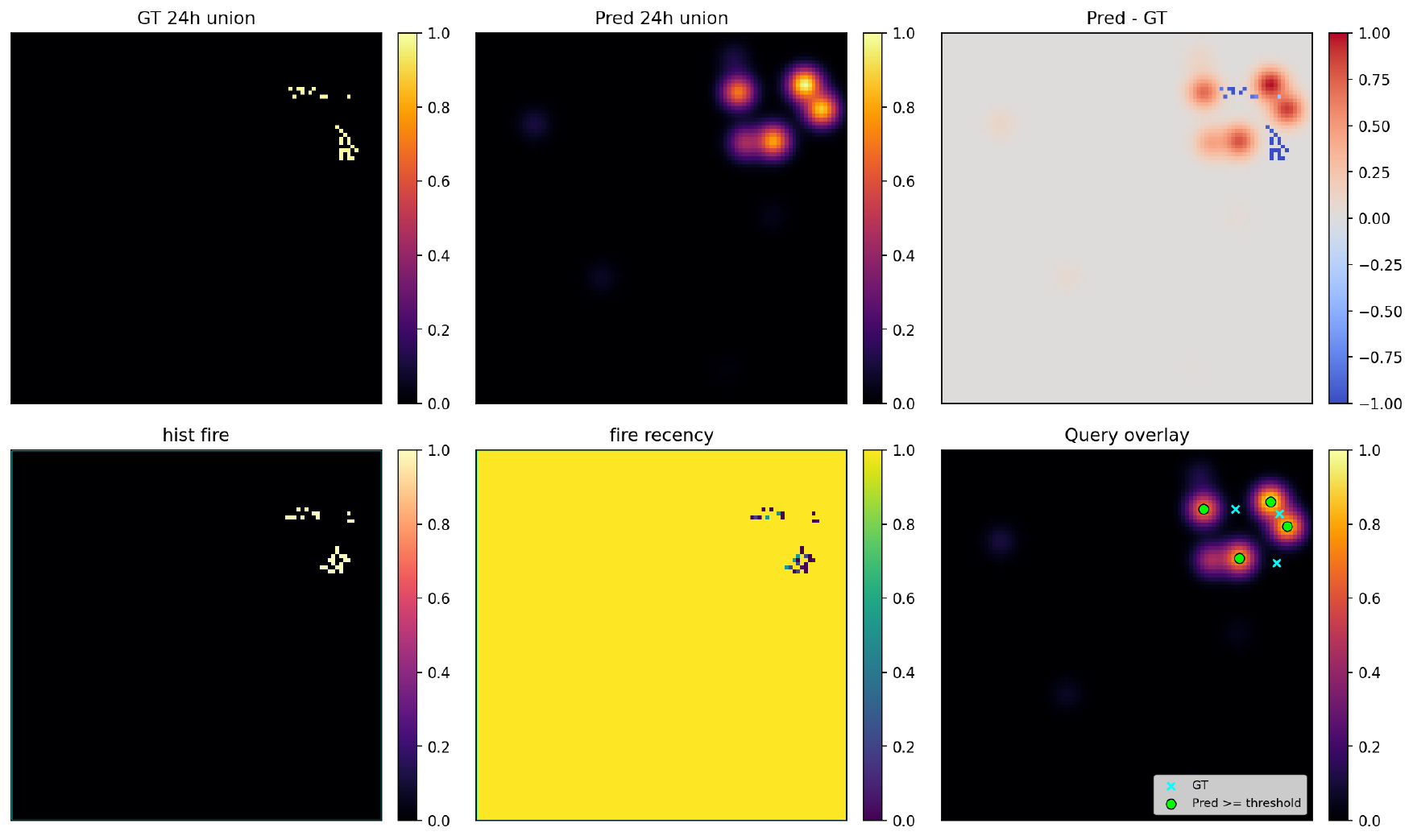} &
    \includegraphics[width=0.305\textwidth]{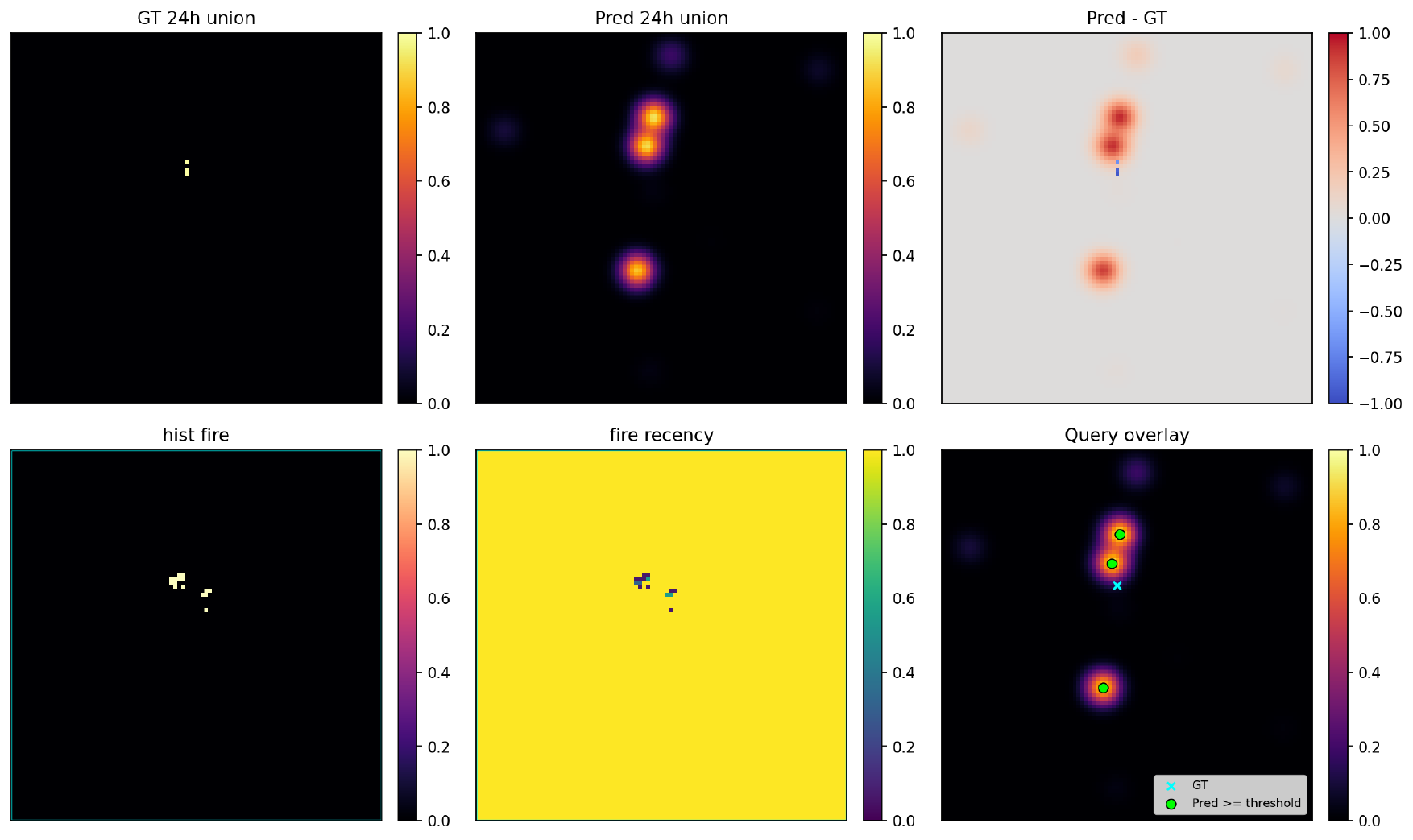} \\[-0.6em]

    \makebox[1.3em][c]{\raisebox{1.35\height}{\rotatebox{90}{\scriptsize\bfseries Median}}} &
    \includegraphics[width=0.305\textwidth]{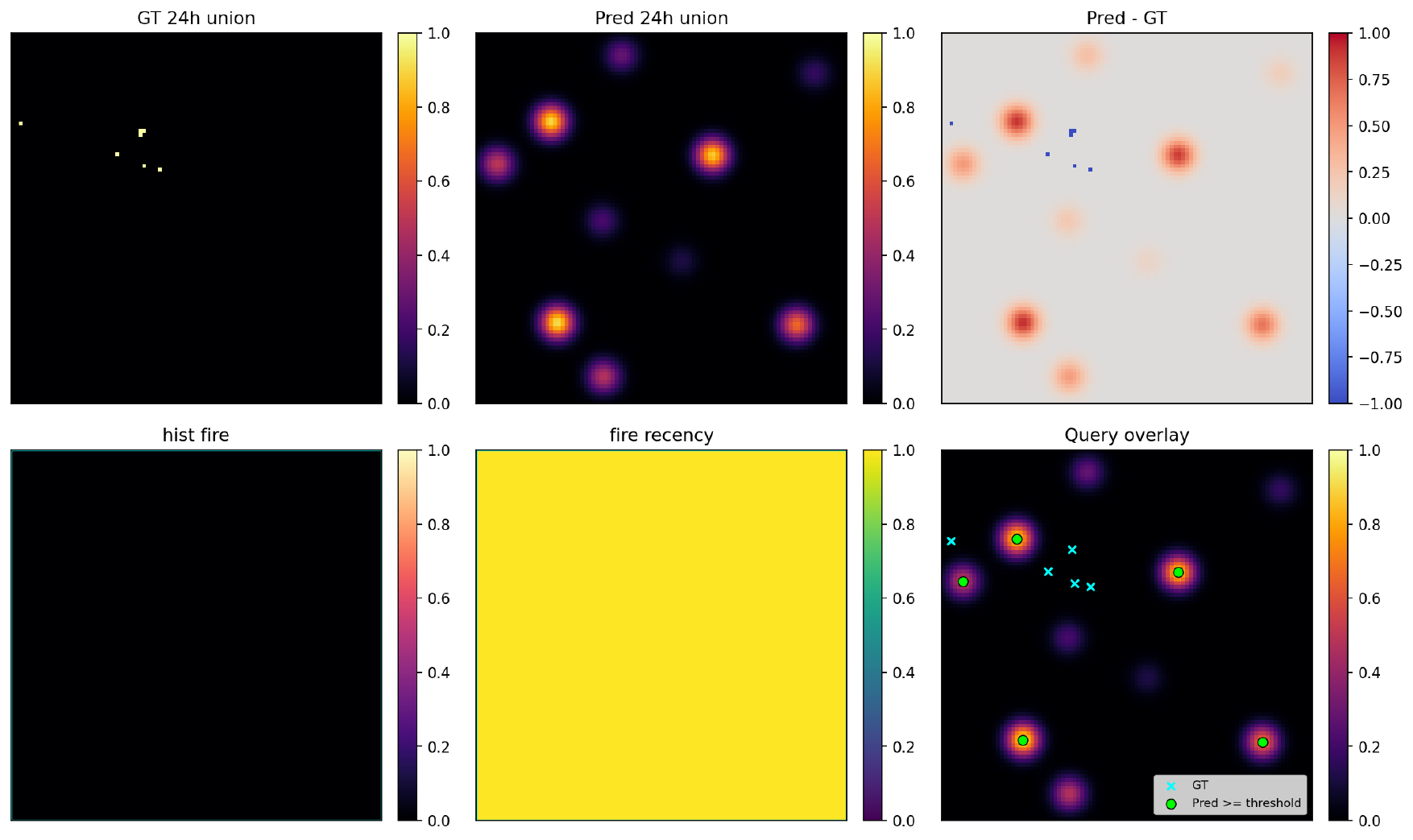} &
    \includegraphics[width=0.305\textwidth]{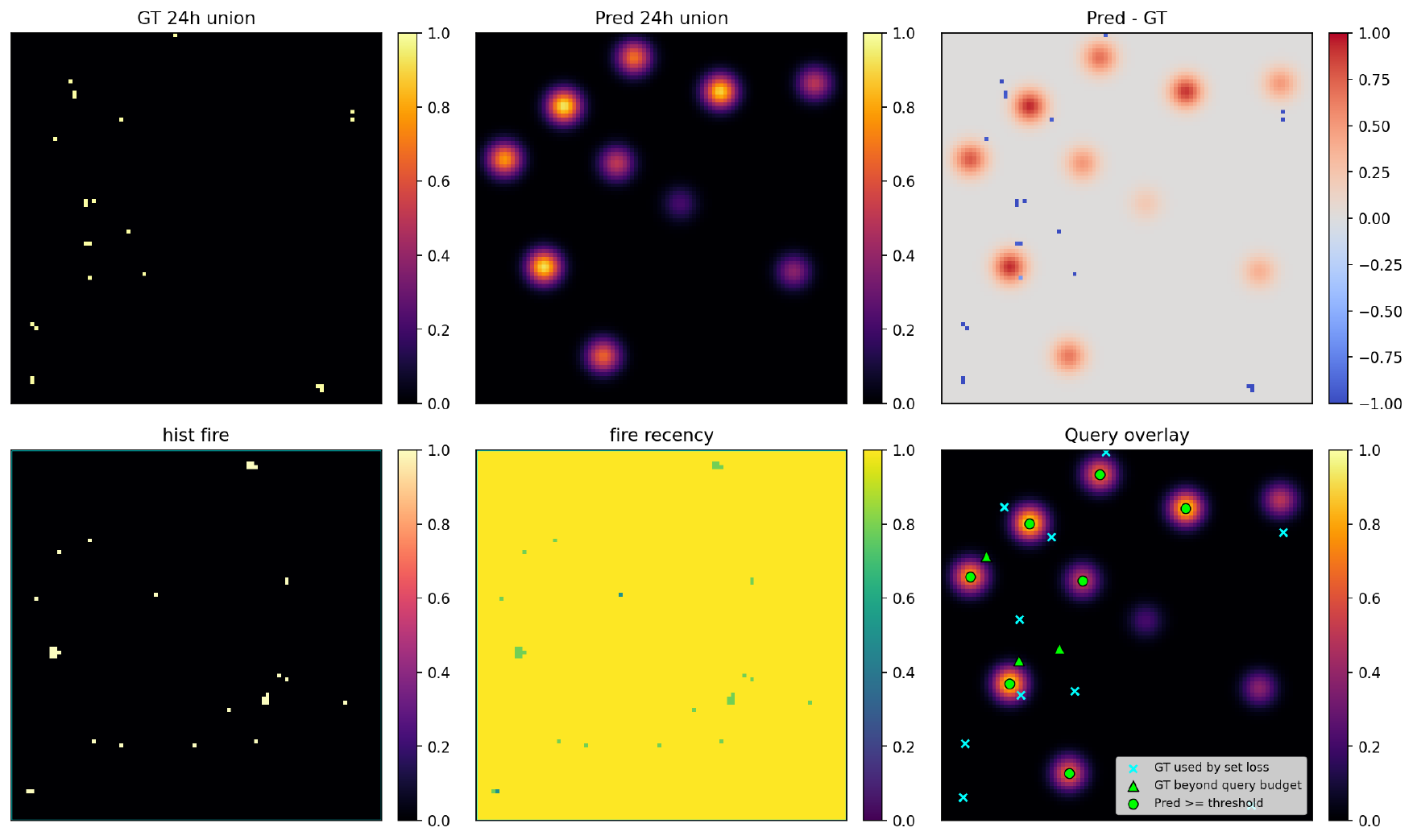} &
    \includegraphics[width=0.305\textwidth]{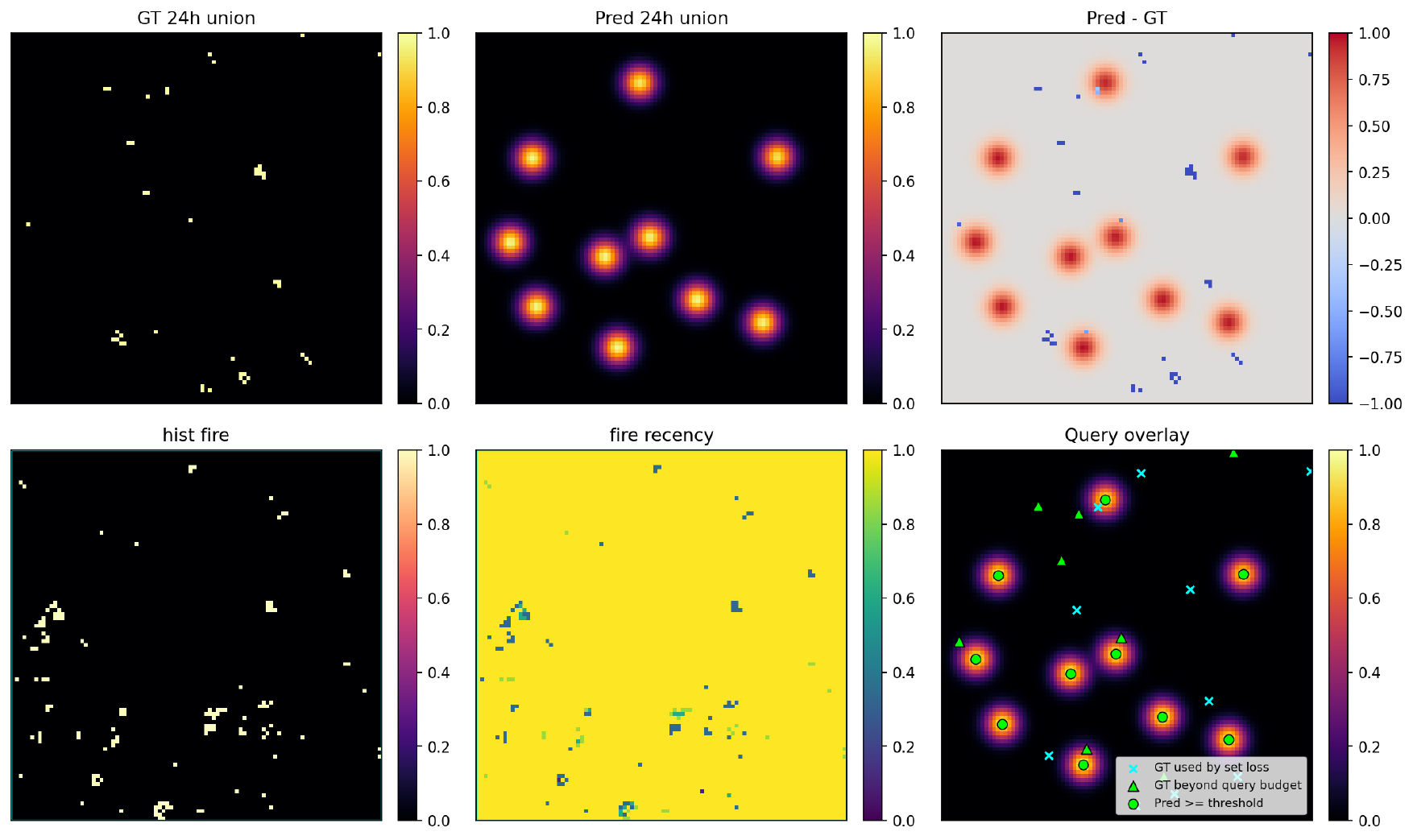} \\[-0.6em]

    \makebox[1.3em][c]{\raisebox{1.35\height}{\rotatebox{90}{\scriptsize\bfseries Failure}}} &
    \includegraphics[width=0.305\textwidth]{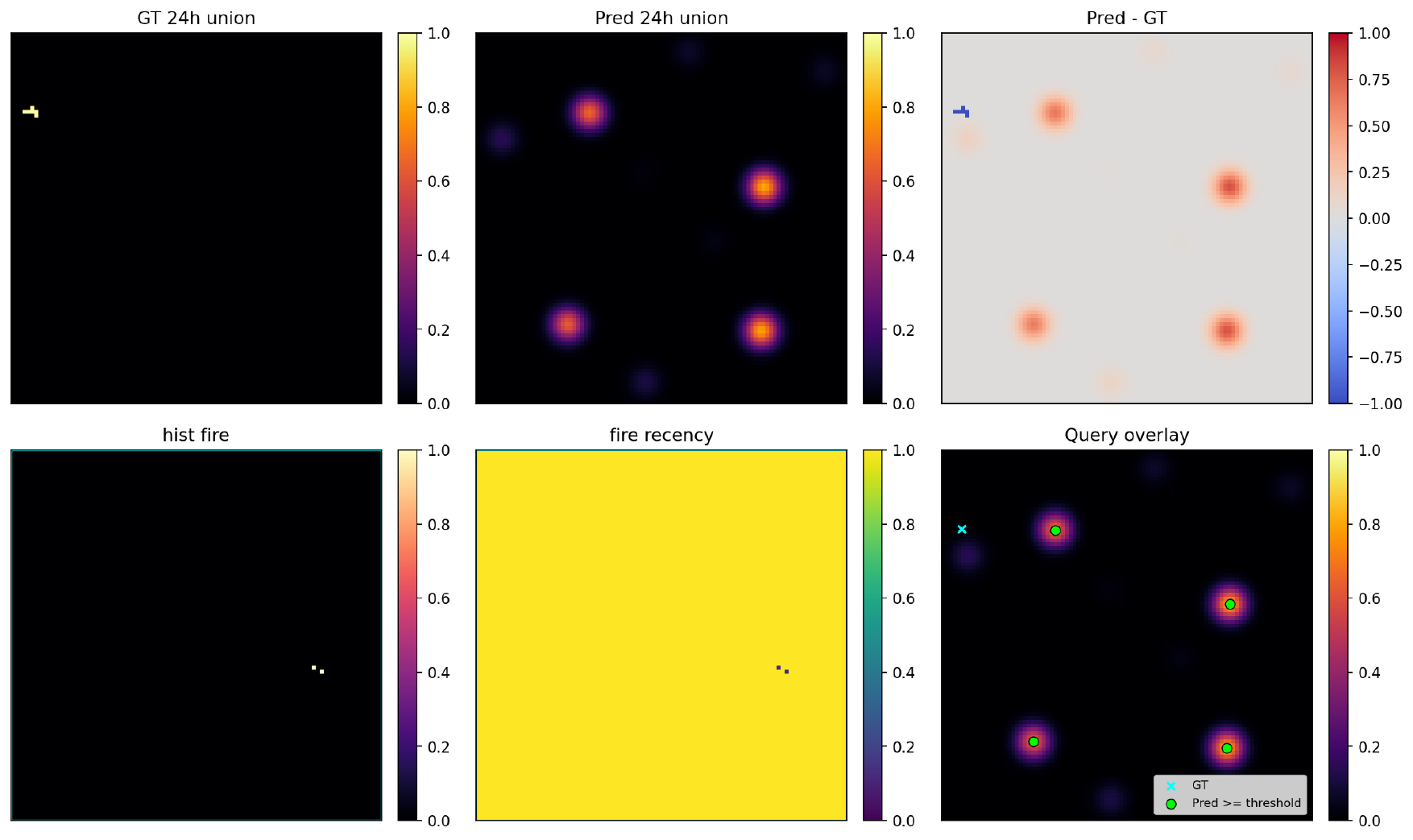} &
    \includegraphics[width=0.305\textwidth]{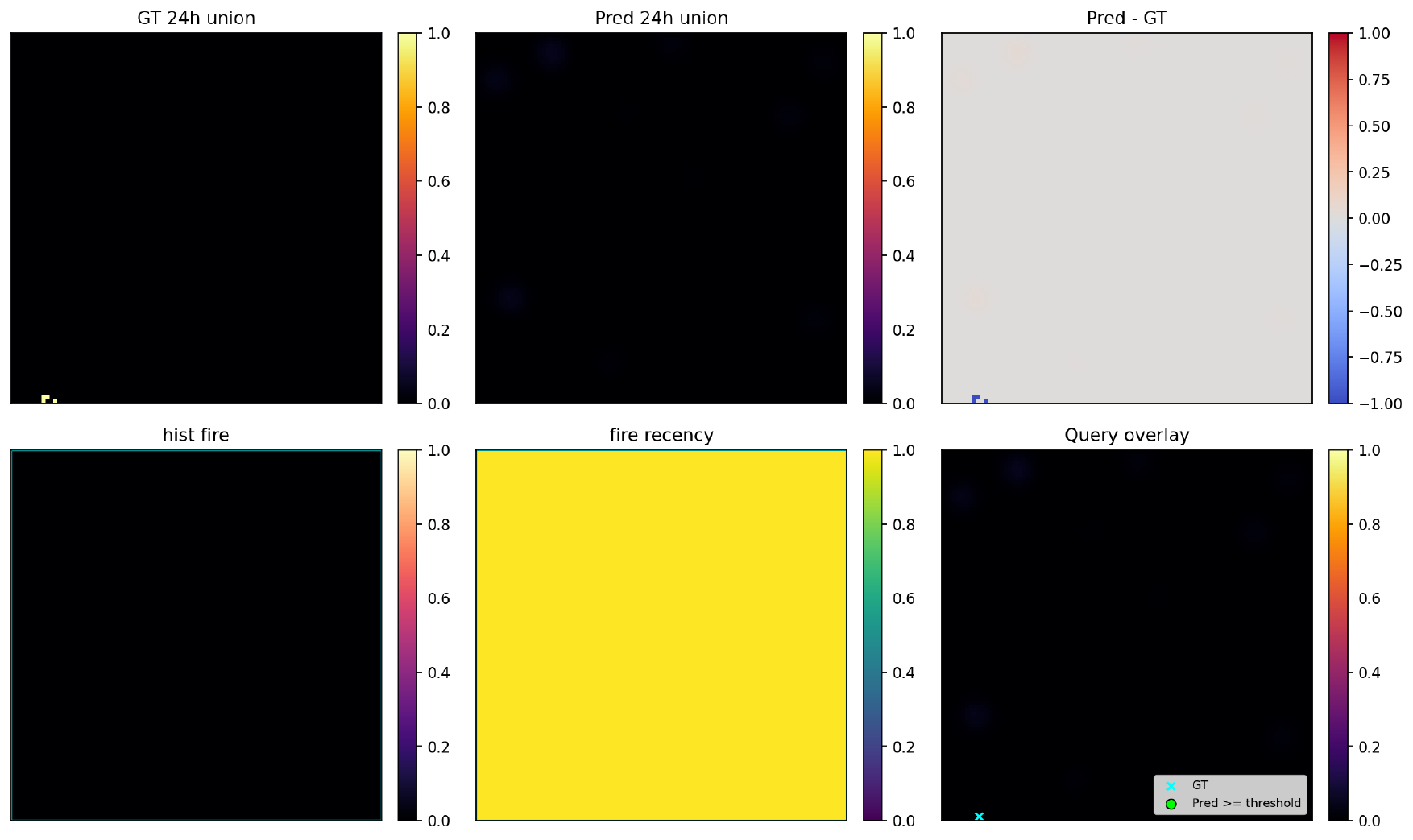} &
    \includegraphics[width=0.305\textwidth]{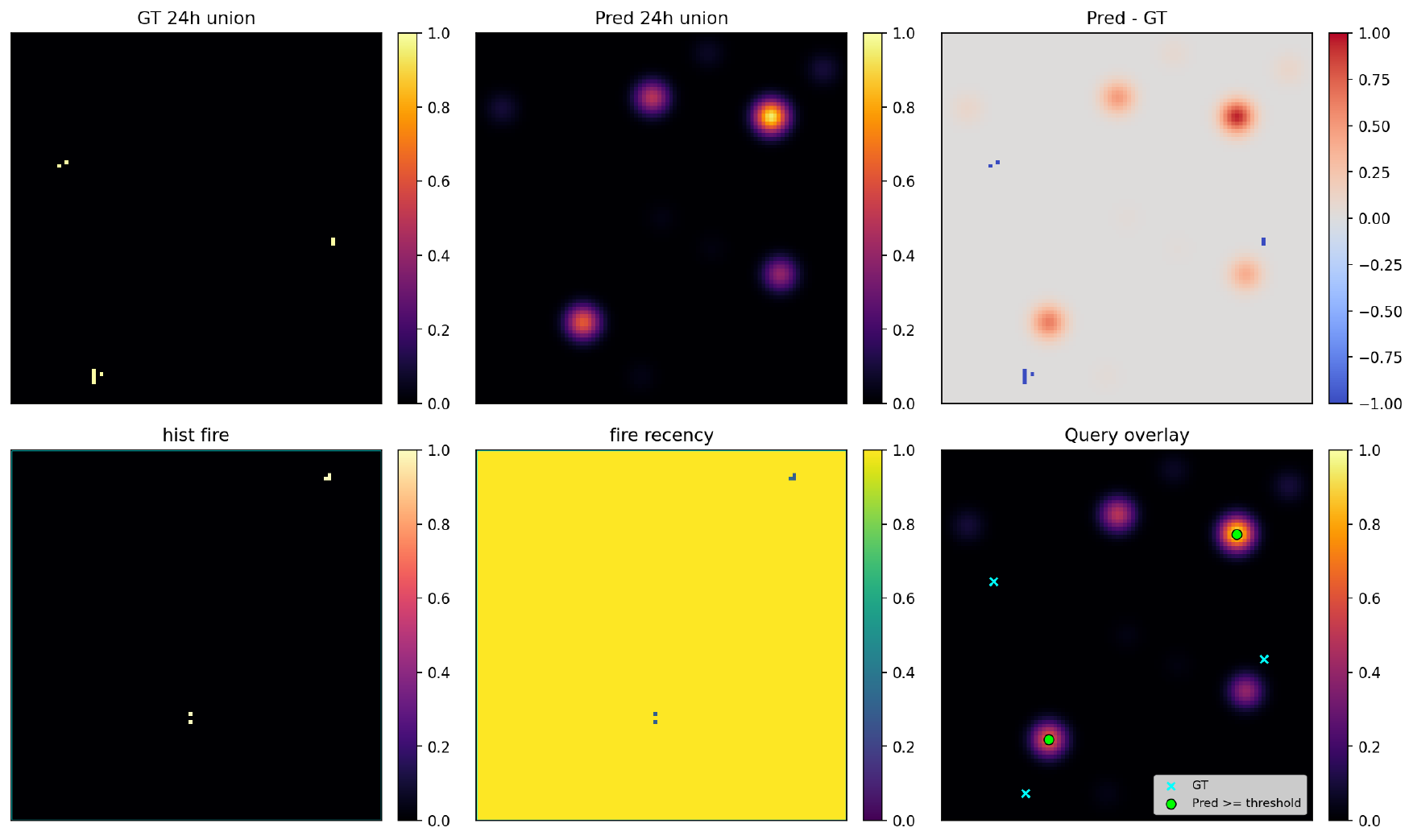}
\end{tabular}
\captionof{figure}{Additional qualitative gallery for WISP variant v2.}
\label{fig:app_qualitative_v2}
\end{center}

\clearpage

\begin{center}
\setlength{\tabcolsep}{1pt}
\renewcommand{\arraystretch}{0.25}
\begin{tabular}{c c c c}
    \makebox[1.3em][c]{\raisebox{1.35\height}{\rotatebox{90}{\scriptsize\bfseries Good}}} &
    \includegraphics[width=0.305\textwidth]{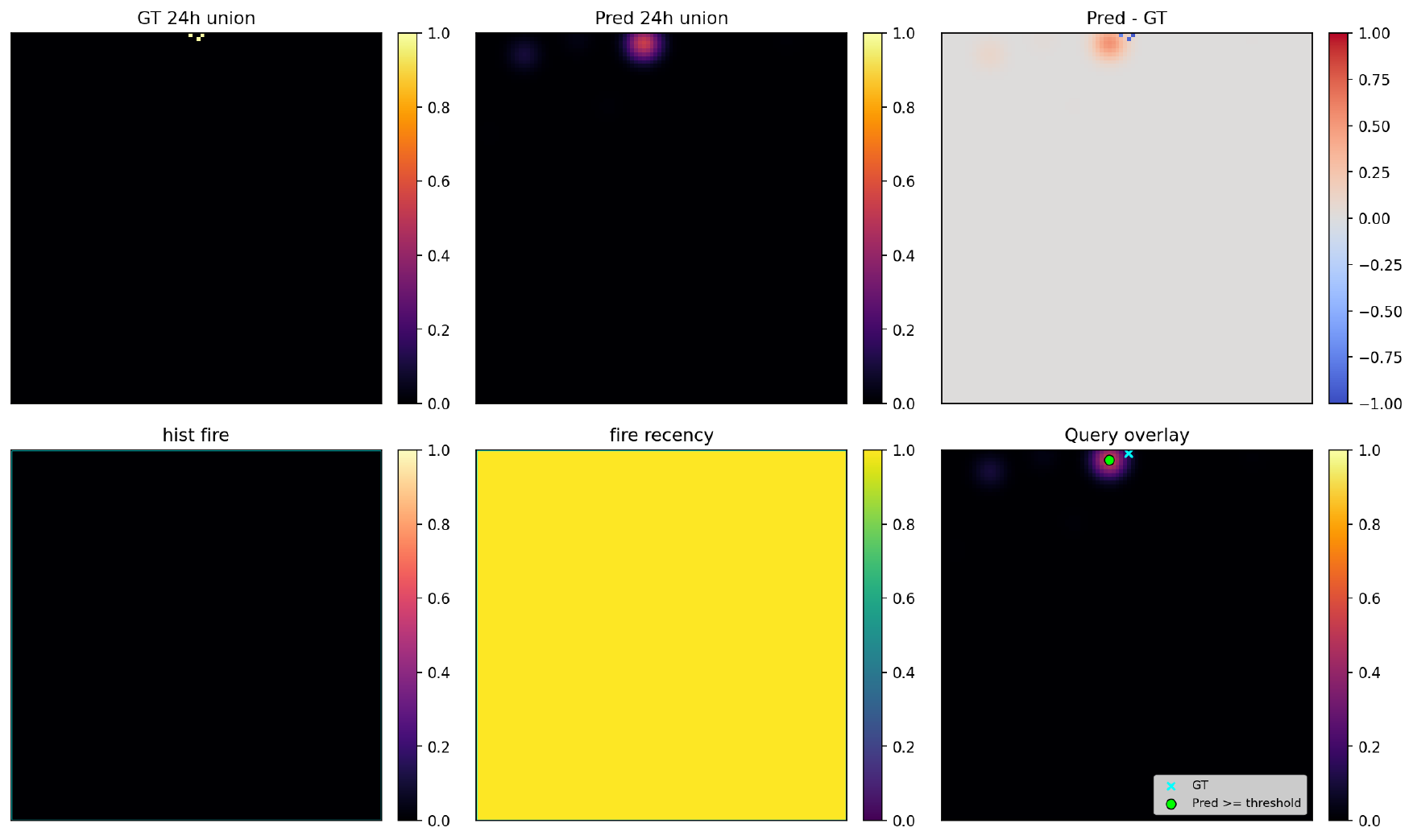} &
    \includegraphics[width=0.305\textwidth]{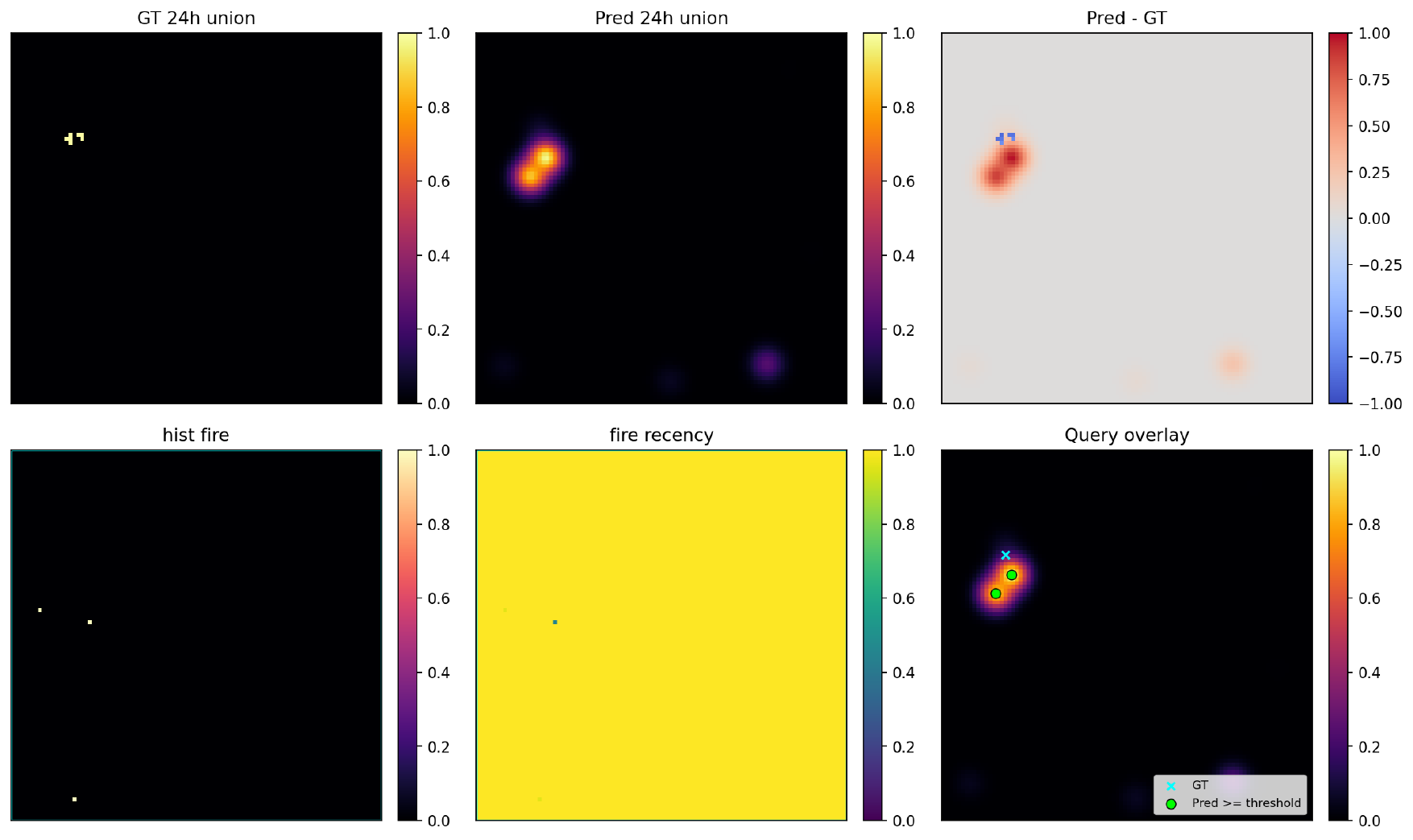} &
    \includegraphics[width=0.305\textwidth]{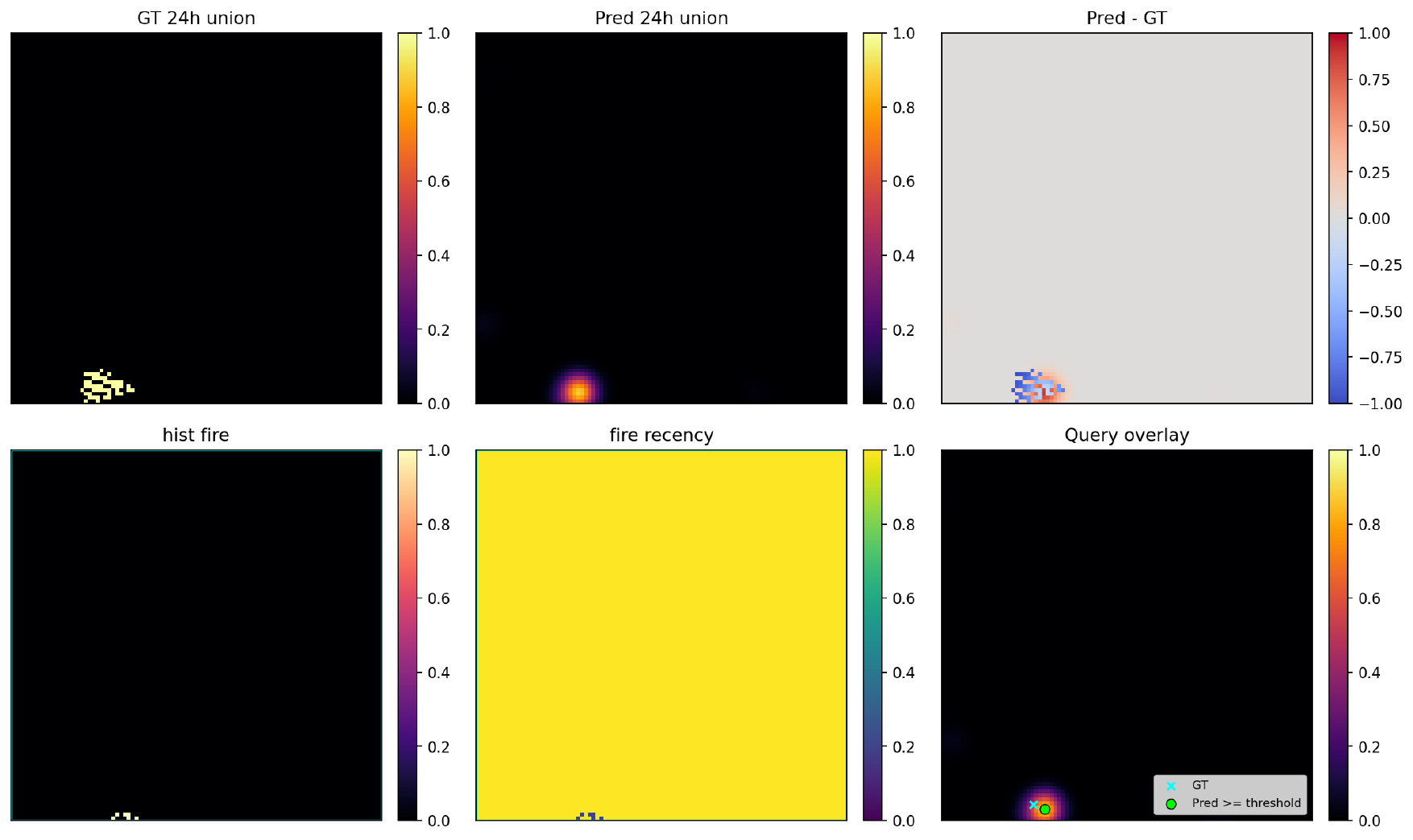} \\[-0.6em]

    \makebox[1.3em][c]{\raisebox{1.35\height}{\rotatebox{90}{\scriptsize\bfseries Median}}} &
    \includegraphics[width=0.305\textwidth]{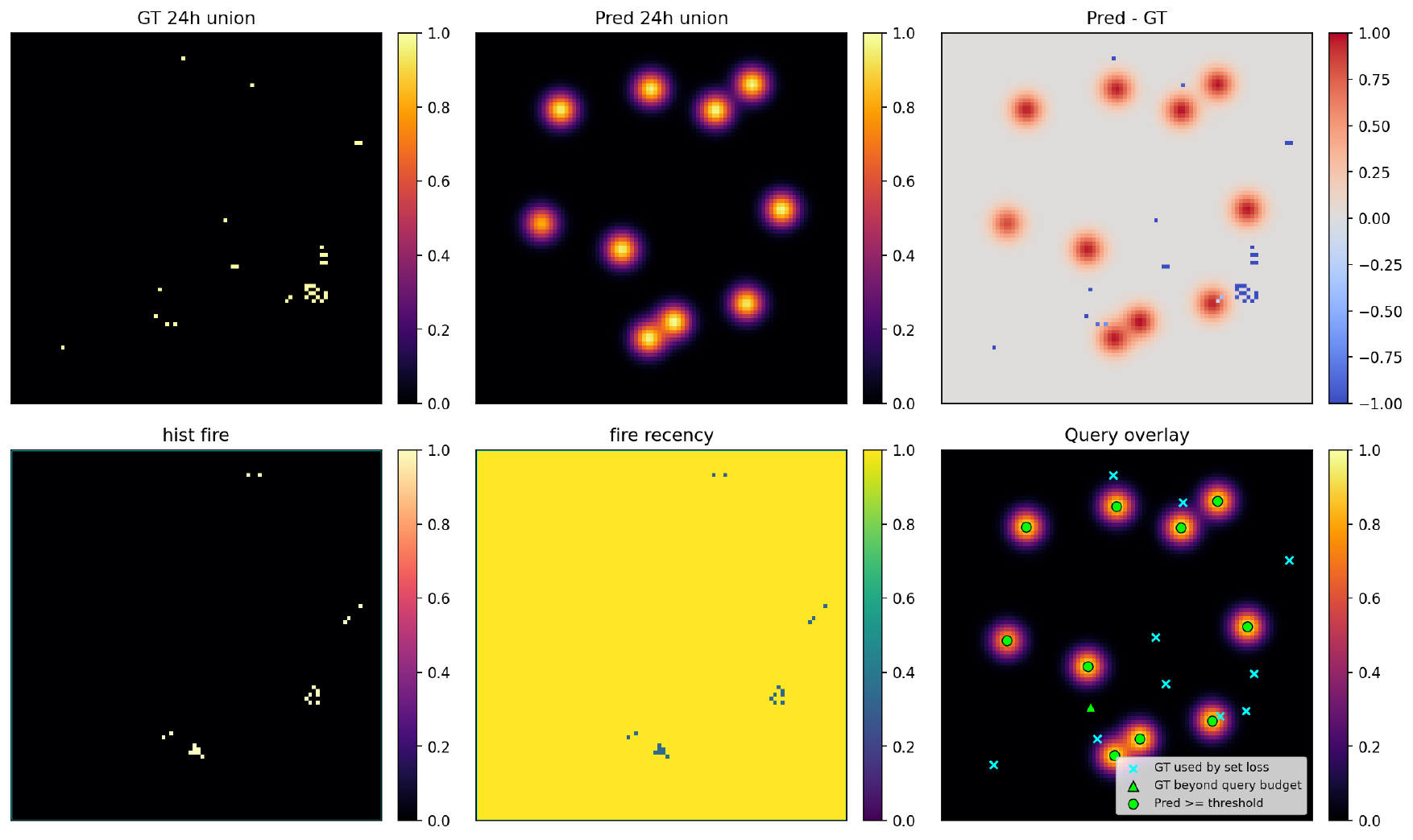} &
    \includegraphics[width=0.305\textwidth]{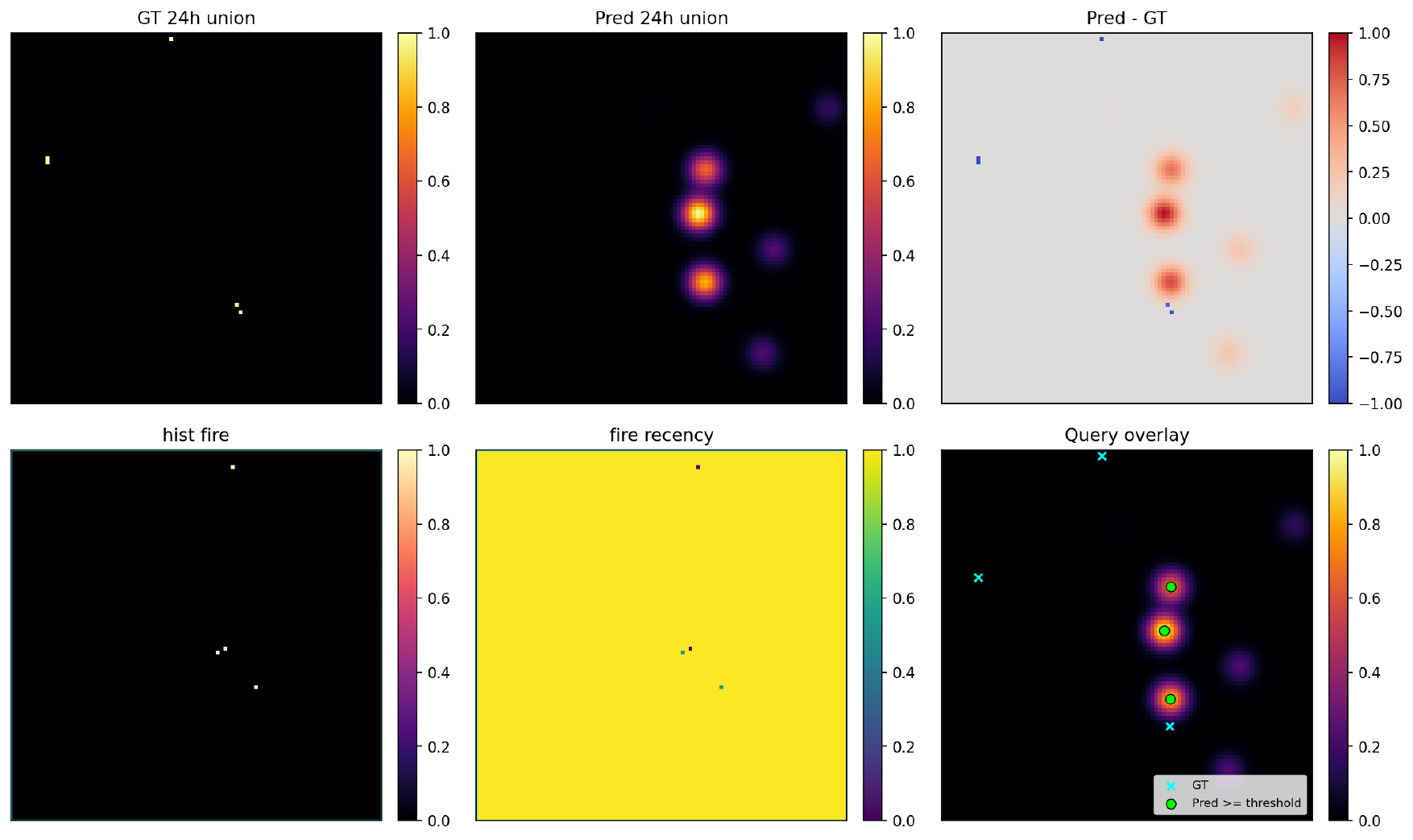} &
    \includegraphics[width=0.305\textwidth]{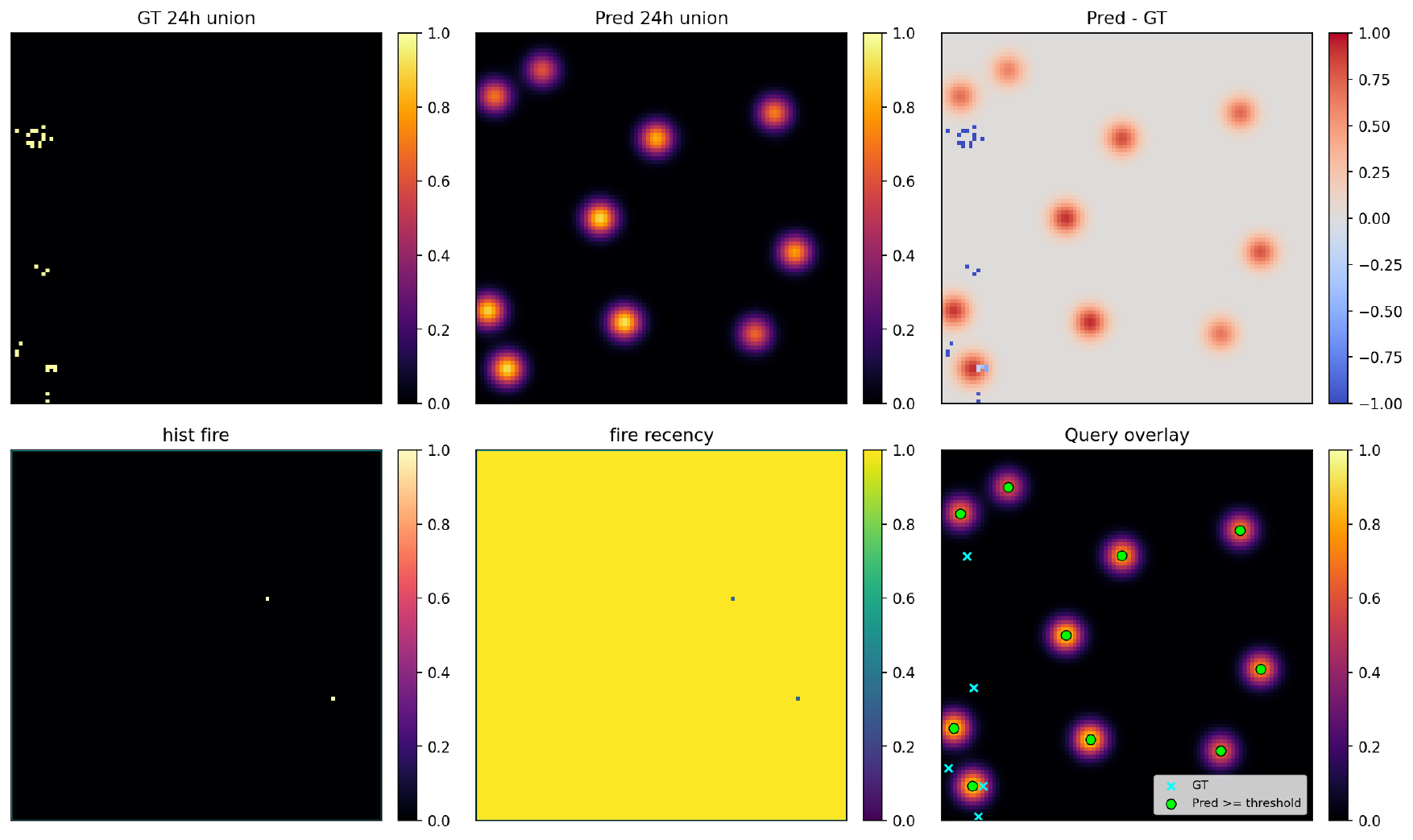} \\[-0.6em]

    \makebox[1.3em][c]{\raisebox{1.35\height}{\rotatebox{90}{\scriptsize\bfseries Failure}}} &
    \includegraphics[width=0.305\textwidth]{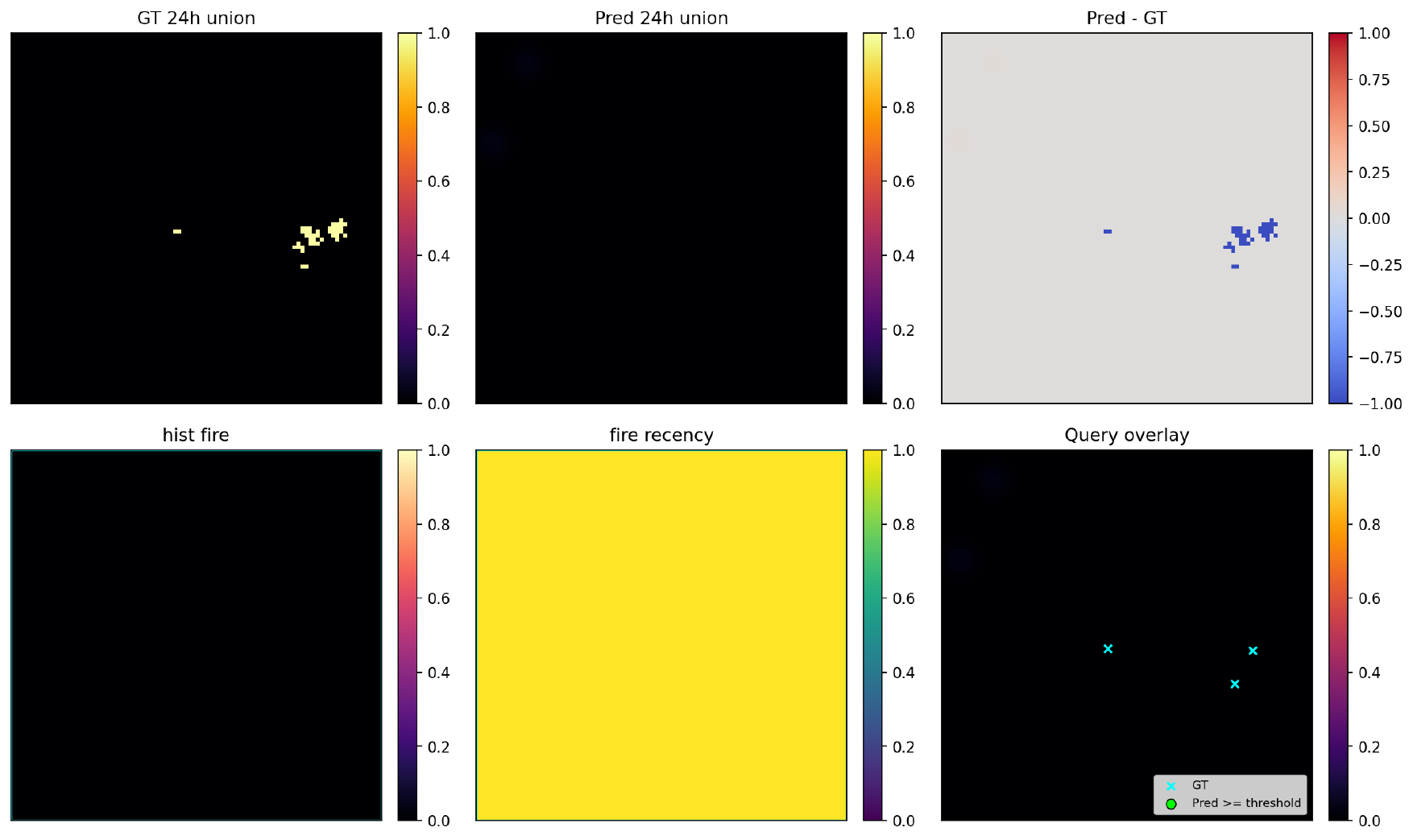} &
    \includegraphics[width=0.305\textwidth]{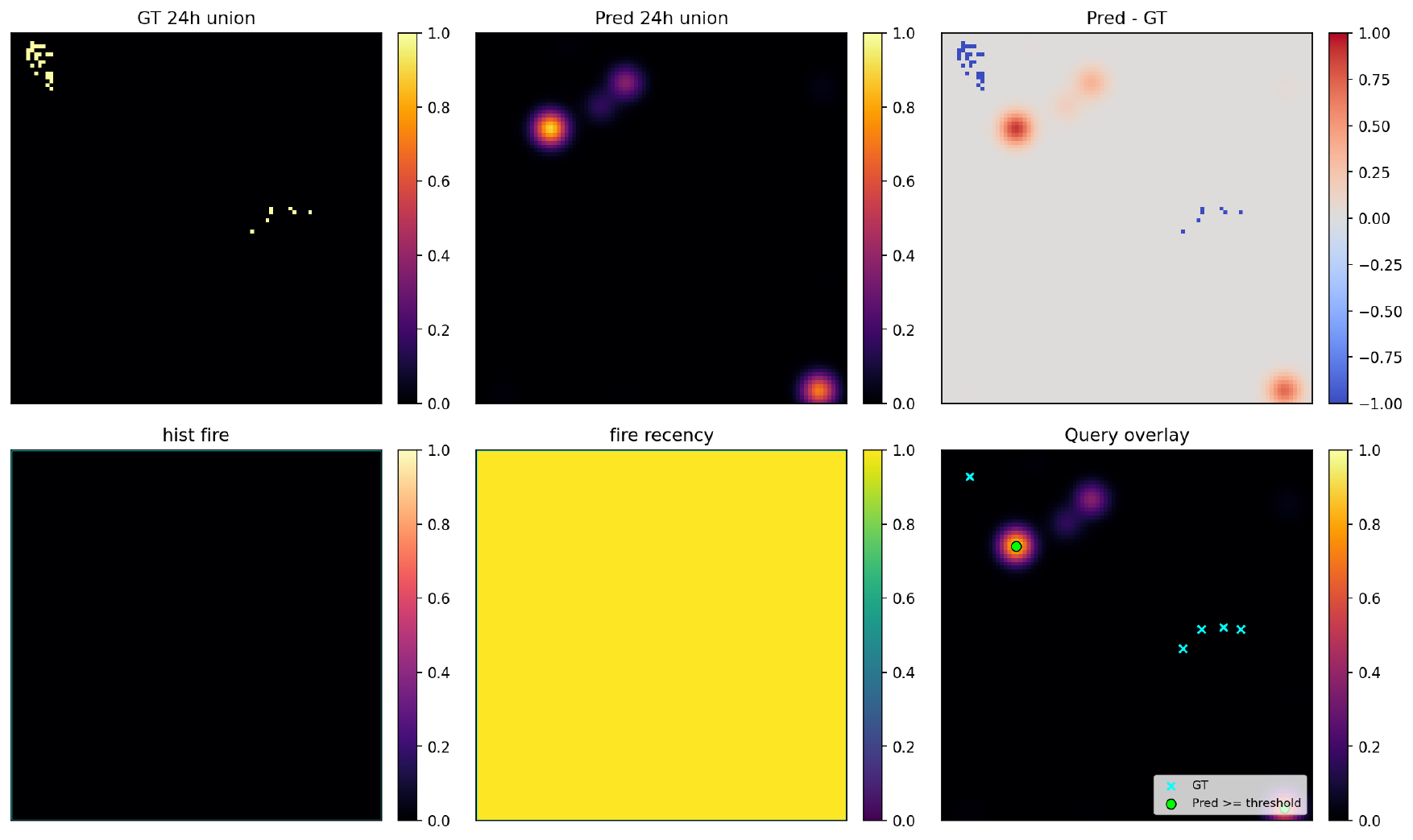} &
    \includegraphics[width=0.305\textwidth]{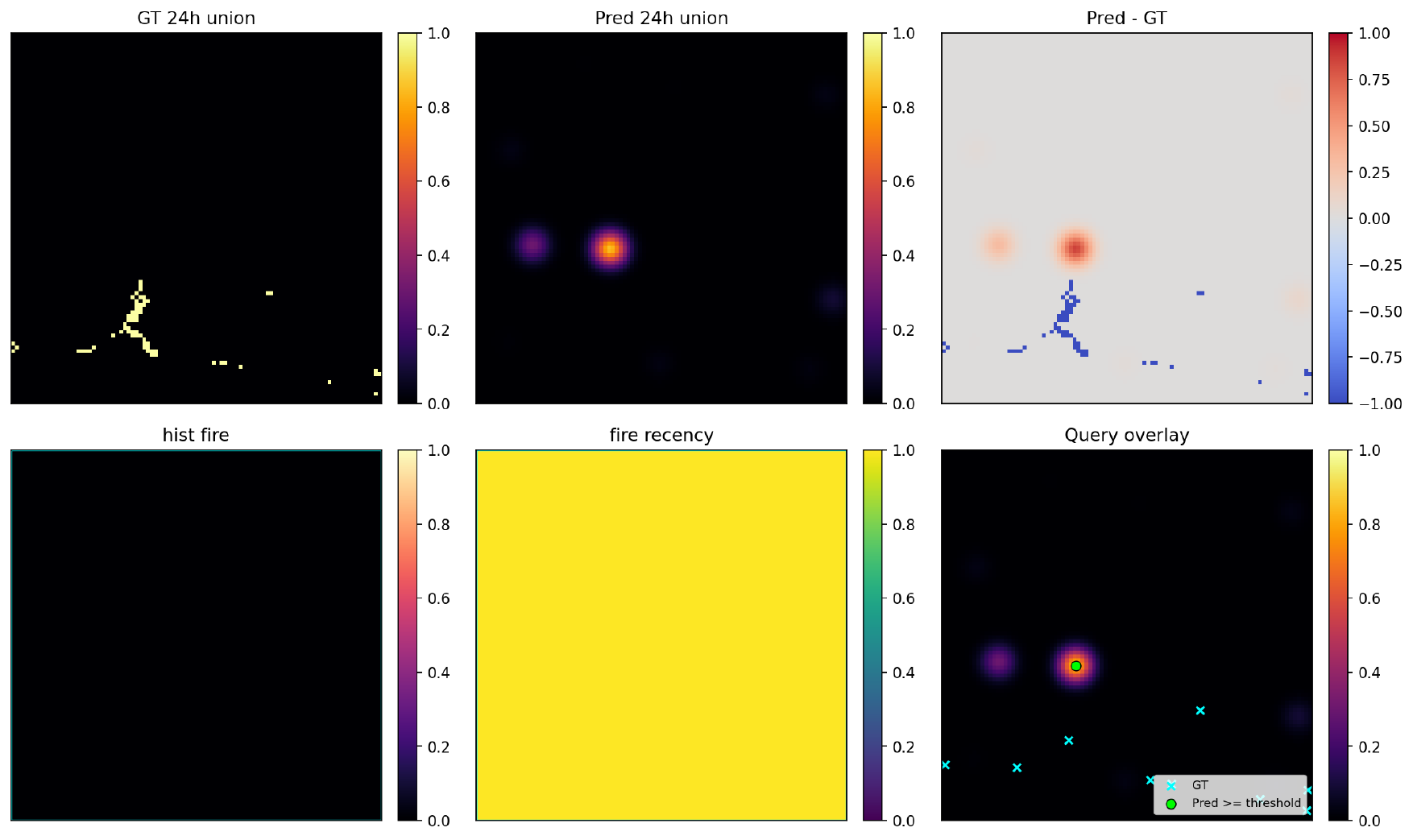}
\end{tabular}
\captionof{figure}{Additional qualitative gallery for WISP variant v3.}
\label{fig:app_qualitative_v3}
\end{center}

\begin{center}
\setlength{\tabcolsep}{1pt}
\renewcommand{\arraystretch}{0.25}
\begin{tabular}{c c c c}
    \makebox[1.3em][c]{\raisebox{1.35\height}{\rotatebox{90}{\scriptsize\bfseries Good}}} &
    \includegraphics[width=0.305\textwidth]{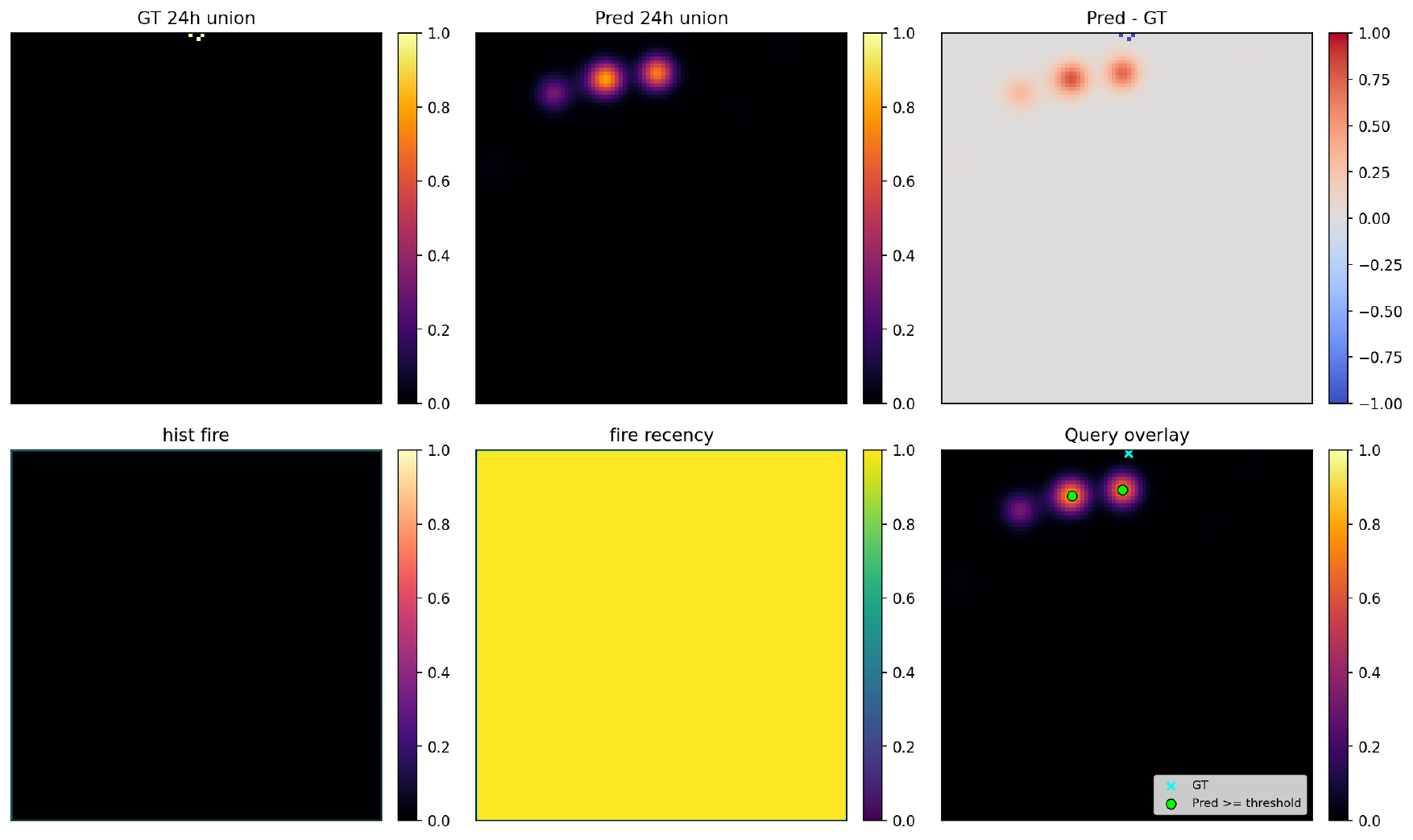} &
    \includegraphics[width=0.305\textwidth]{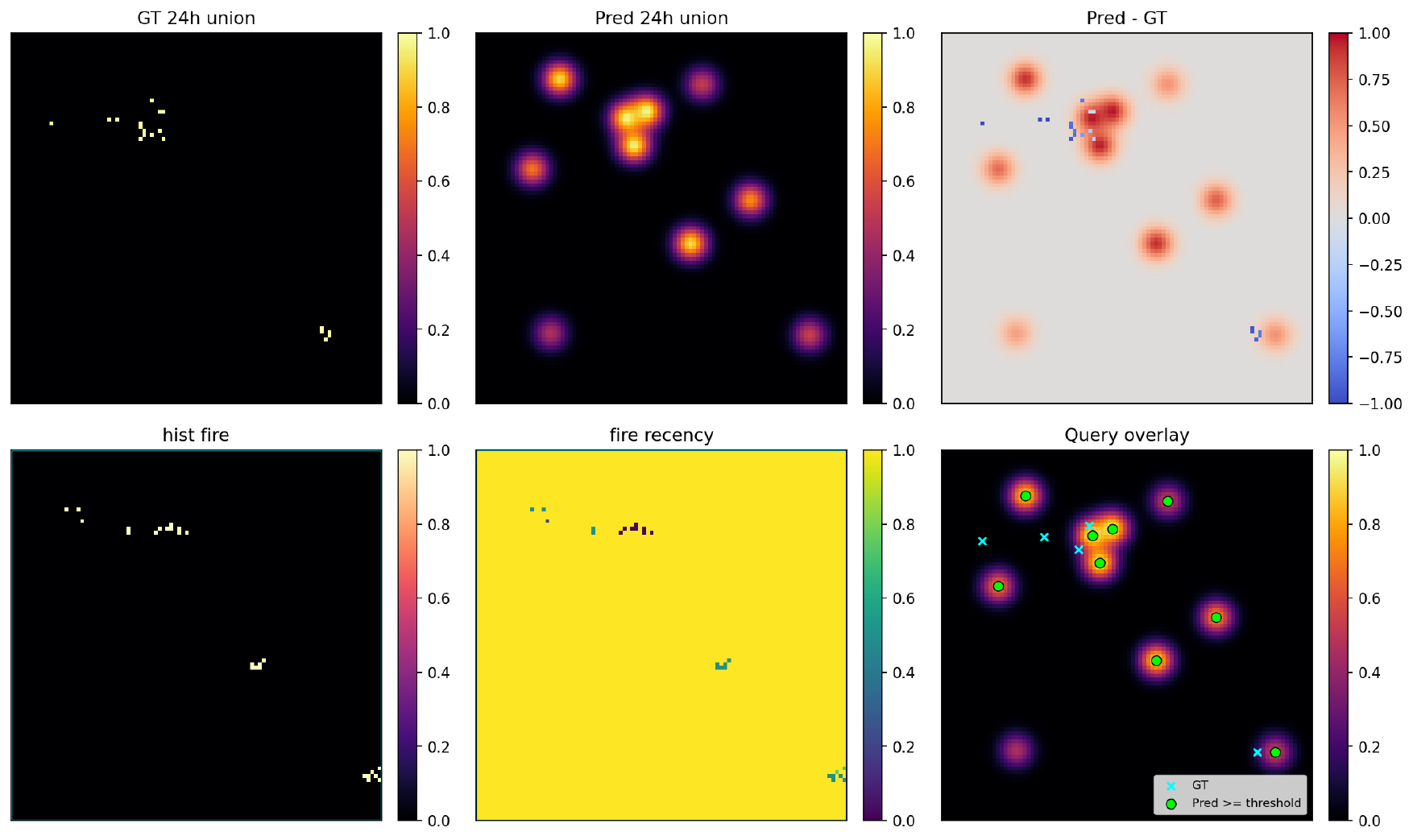} &
    \includegraphics[width=0.305\textwidth]{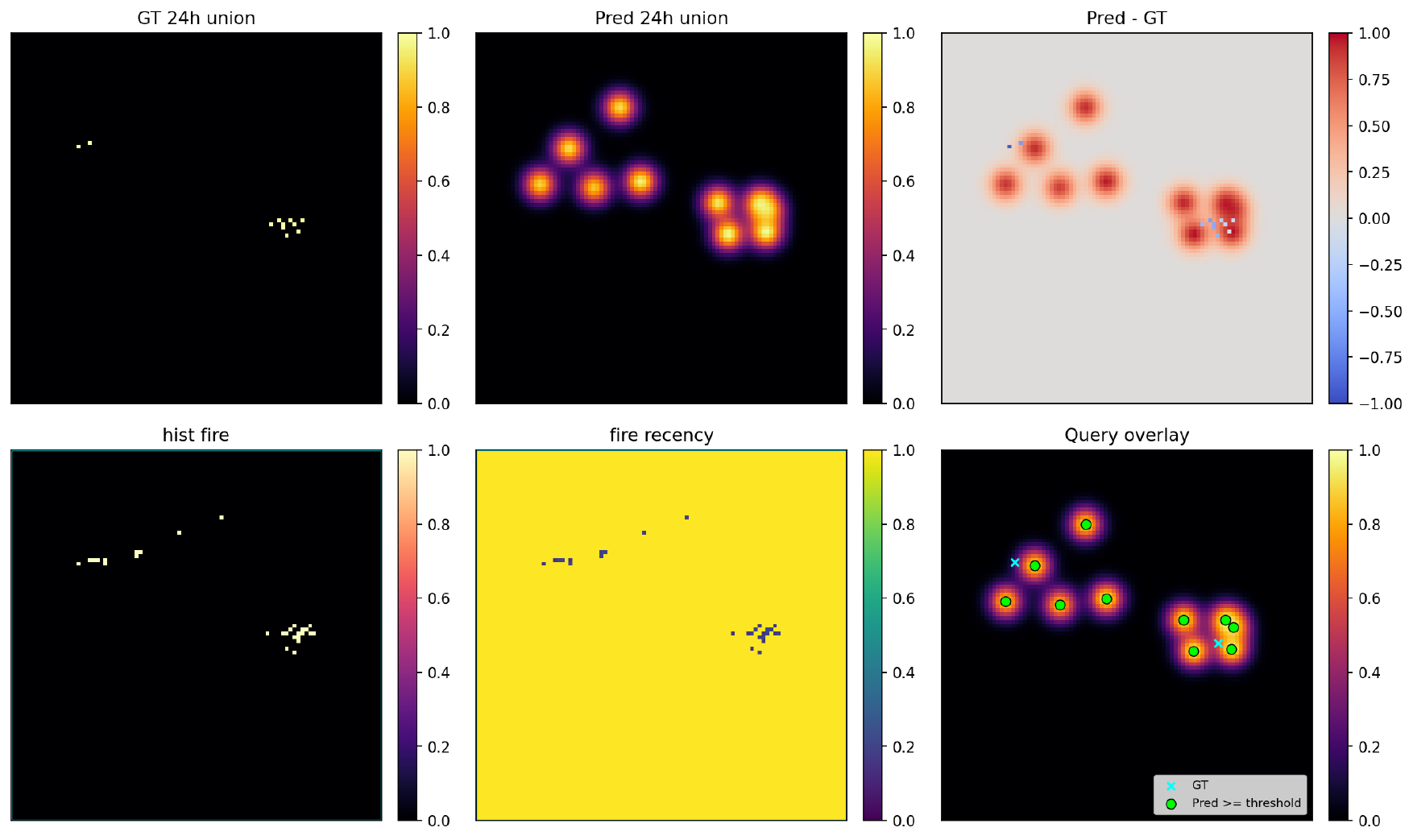} \\[-0.6em]

    \makebox[1.3em][c]{\raisebox{1.35\height}{\rotatebox{90}{\scriptsize\bfseries Median}}} &
    \includegraphics[width=0.305\textwidth]{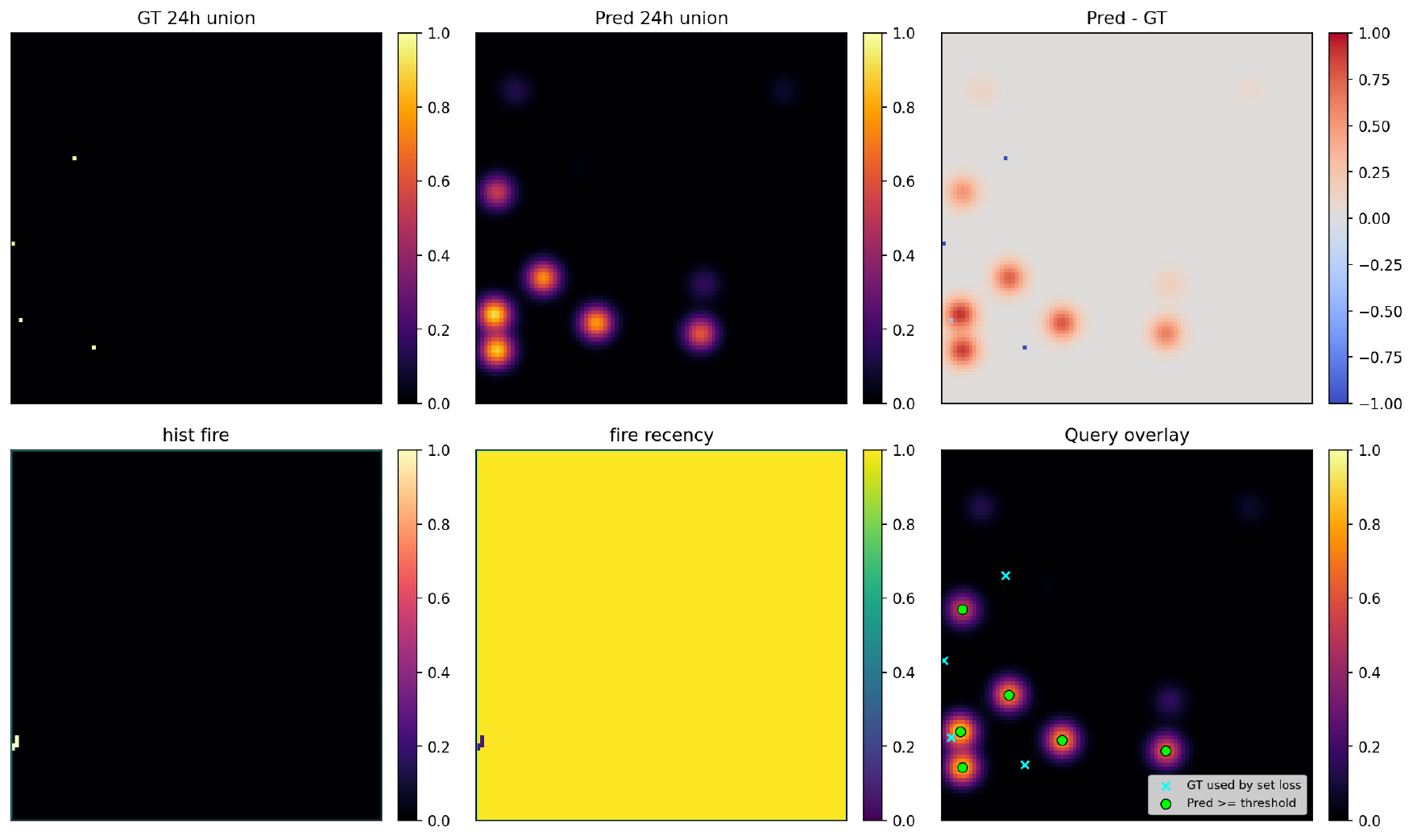} &
    \includegraphics[width=0.305\textwidth]{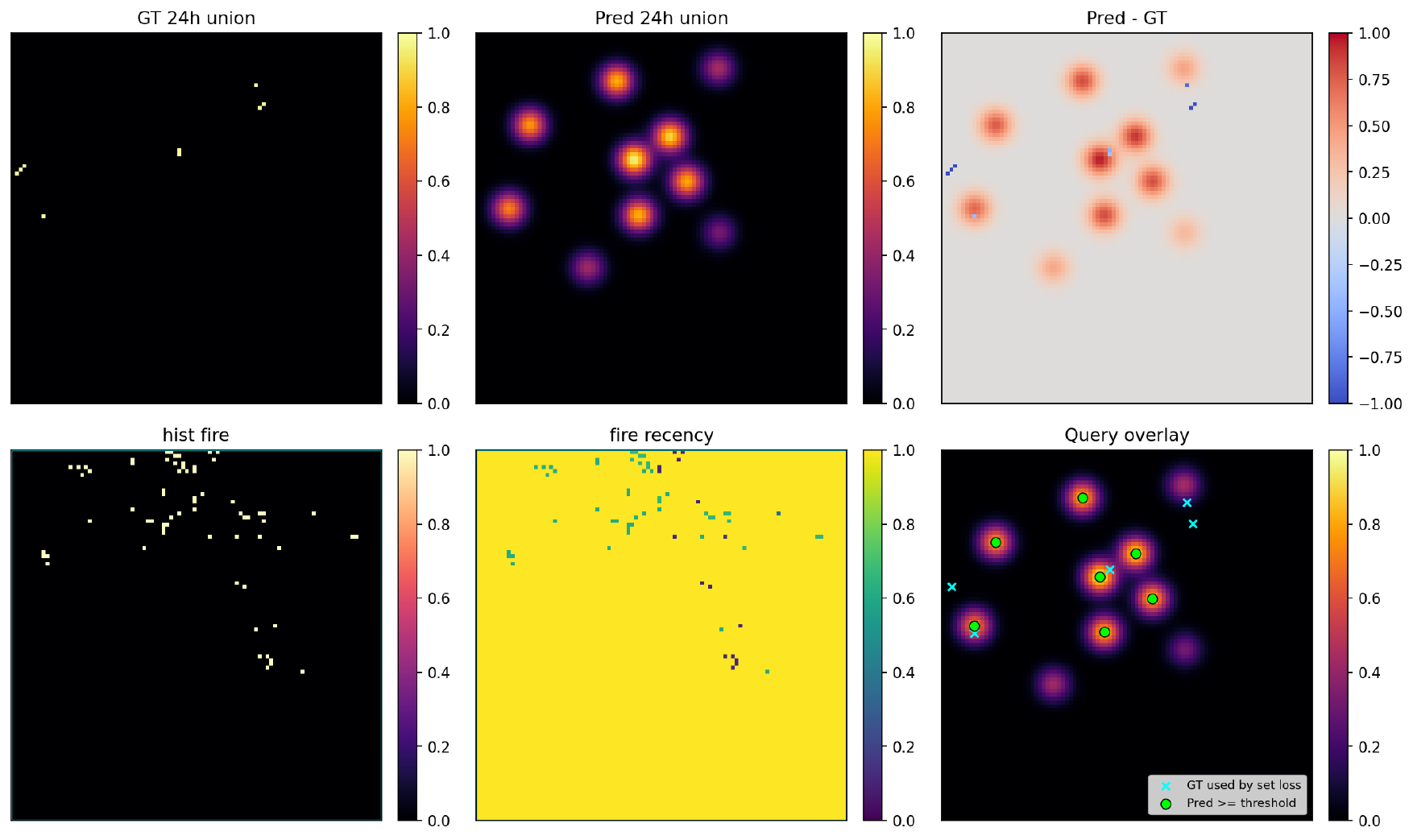} &
    \includegraphics[width=0.305\textwidth]{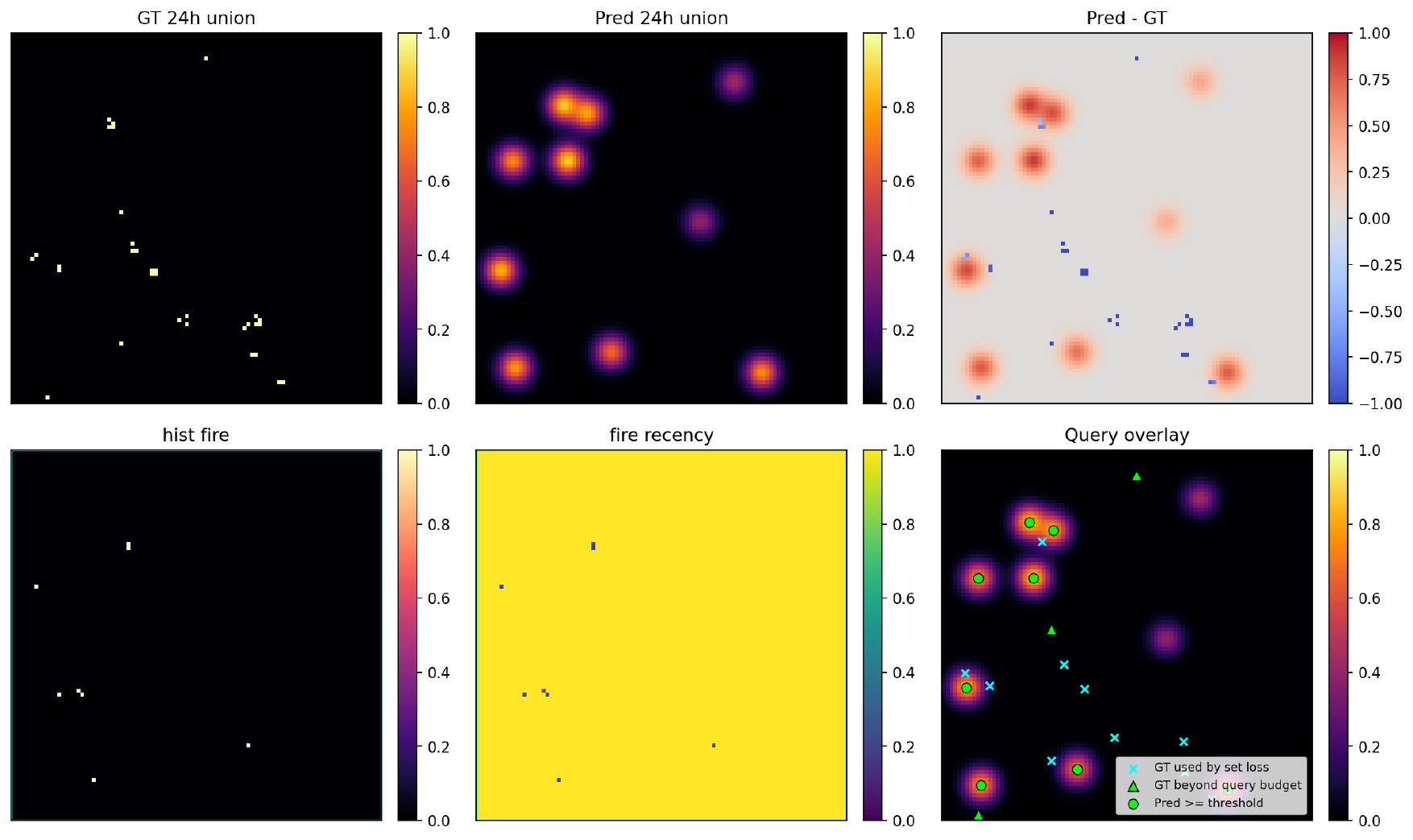} \\[-0.6em]

    \makebox[1.3em][c]{\raisebox{1.35\height}{\rotatebox{90}{\scriptsize\bfseries Failure}}} &
    \includegraphics[width=0.305\textwidth]{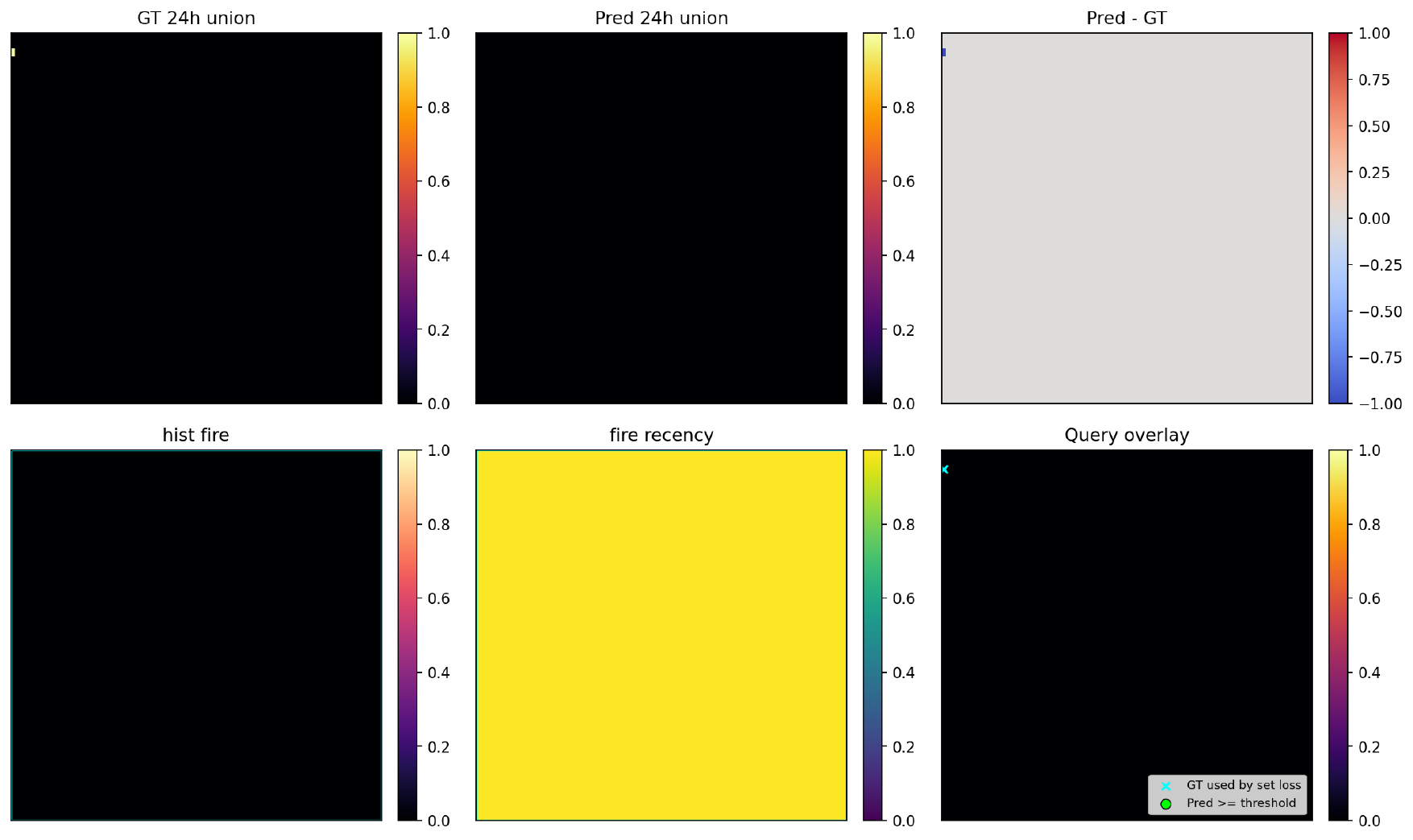} &
    \includegraphics[width=0.305\textwidth]{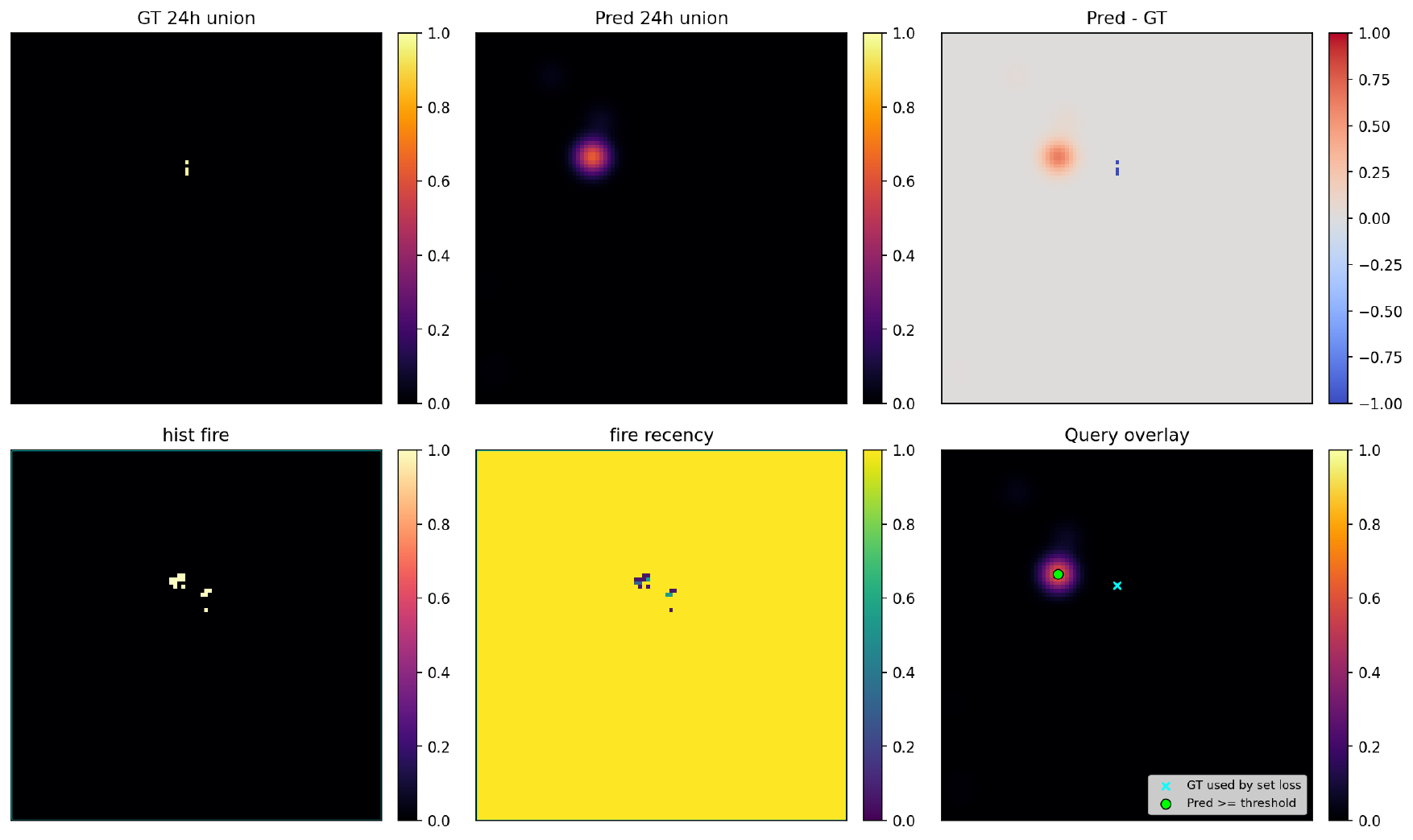} &
    \includegraphics[width=0.305\textwidth]{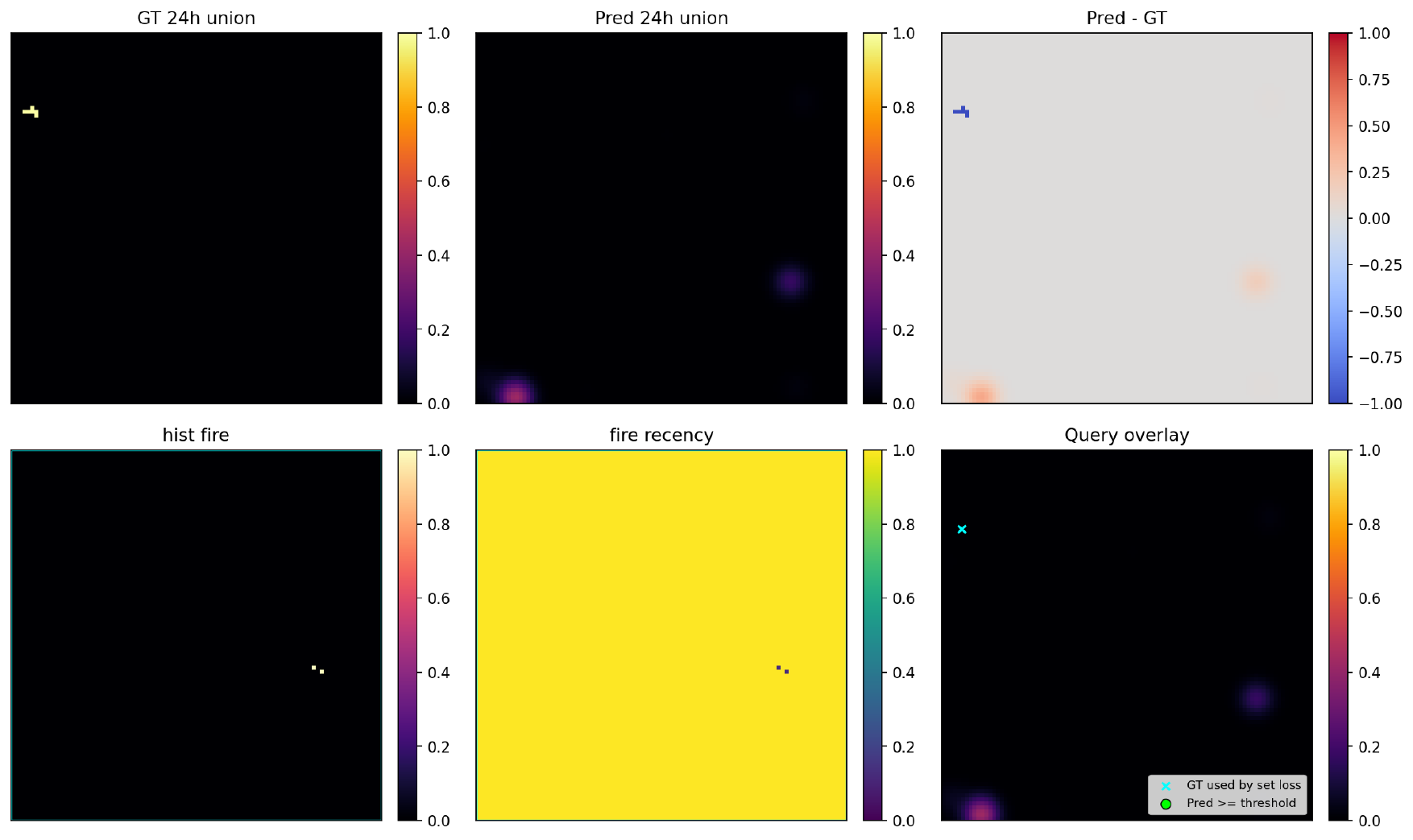}
\end{tabular}
\captionof{figure}{Additional qualitative gallery for WISP variant v4.}
\label{fig:app_qualitative_v4}
\end{center}

\clearpage

\begin{center}
\setlength{\tabcolsep}{1pt}
\renewcommand{\arraystretch}{0.25}
\begin{tabular}{c c c c}
    \makebox[1.3em][c]{\raisebox{1.35\height}{\rotatebox{90}{\scriptsize\bfseries Good}}} &
    \includegraphics[width=0.305\textwidth]{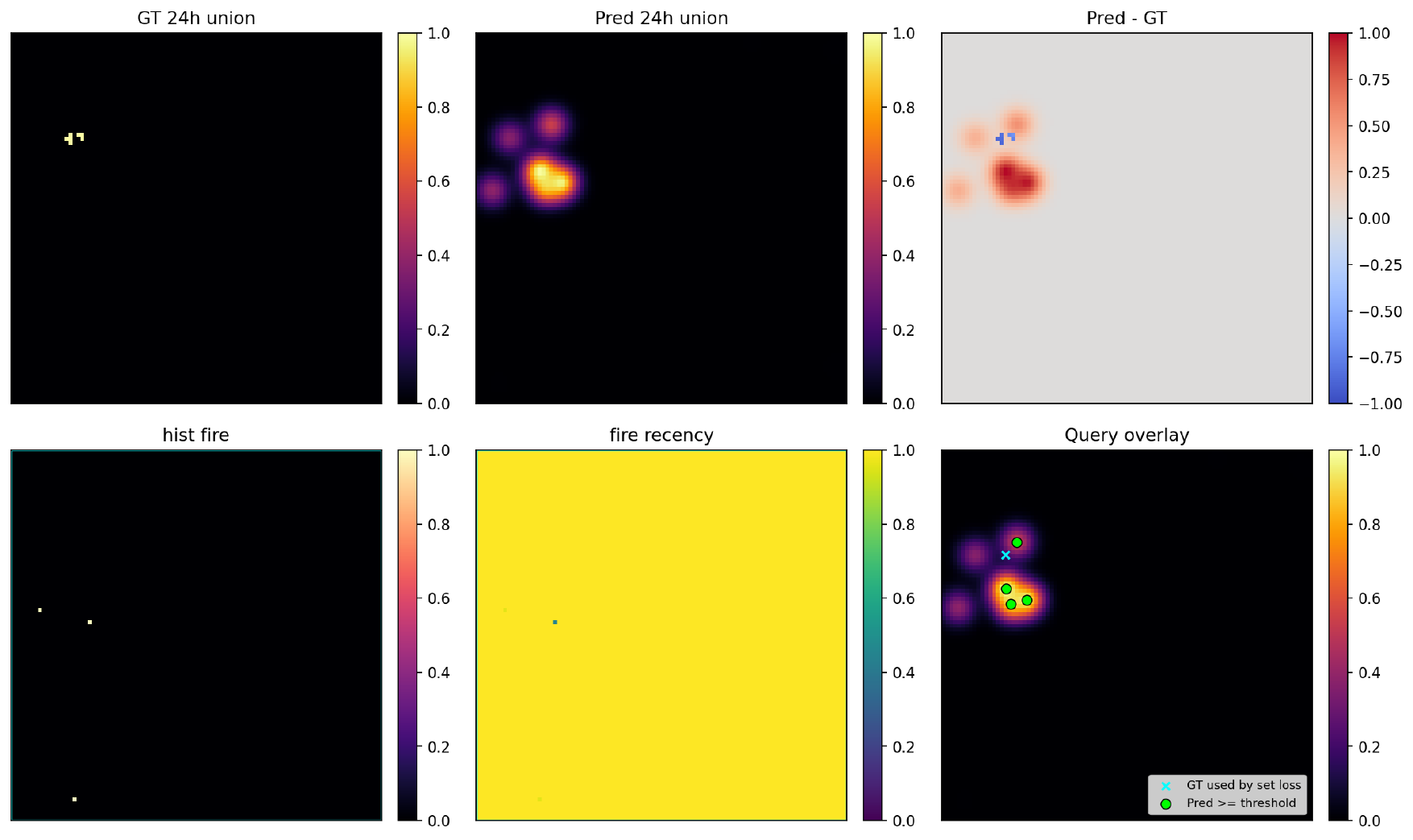} &
    \includegraphics[width=0.305\textwidth]{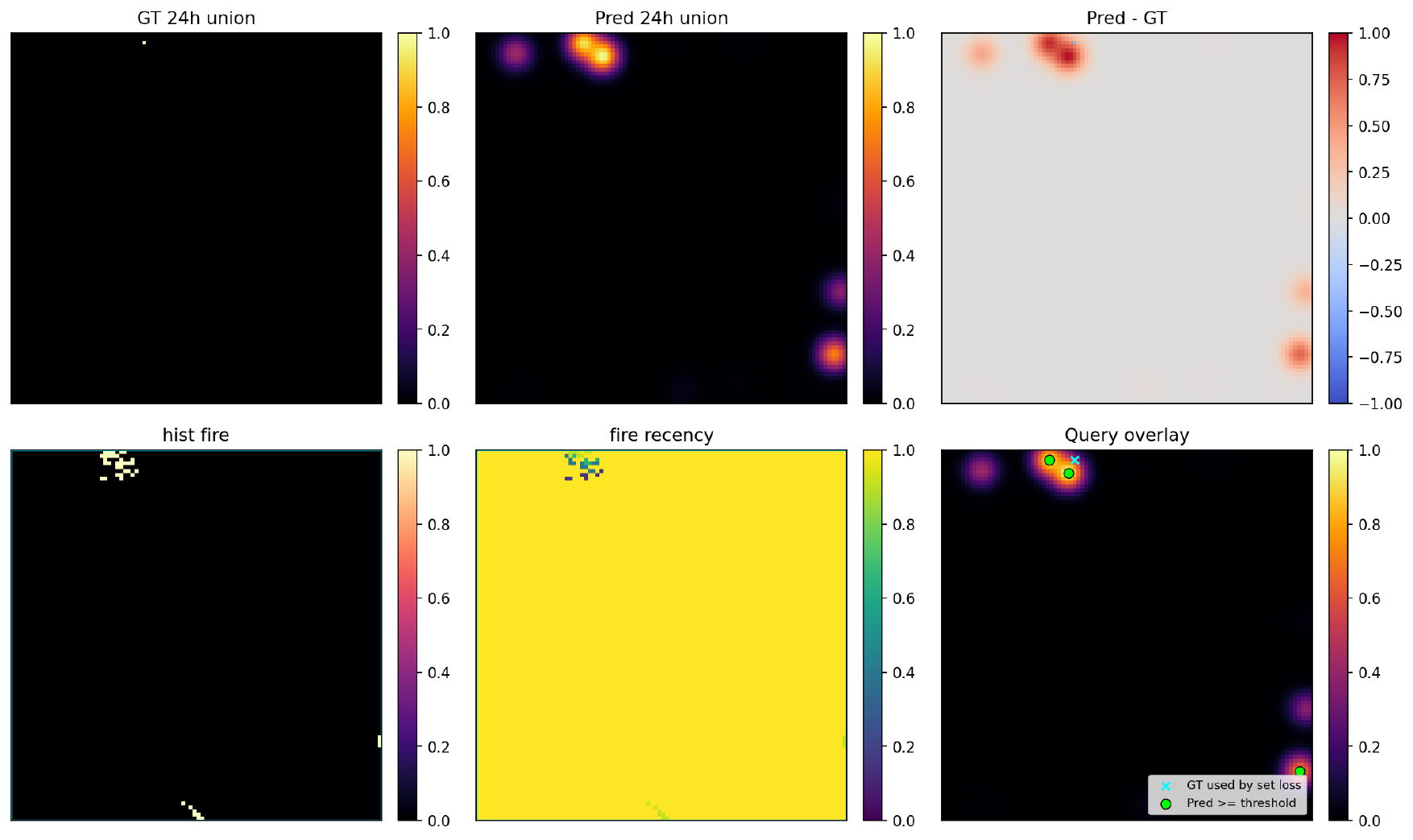} &
    \includegraphics[width=0.305\textwidth]{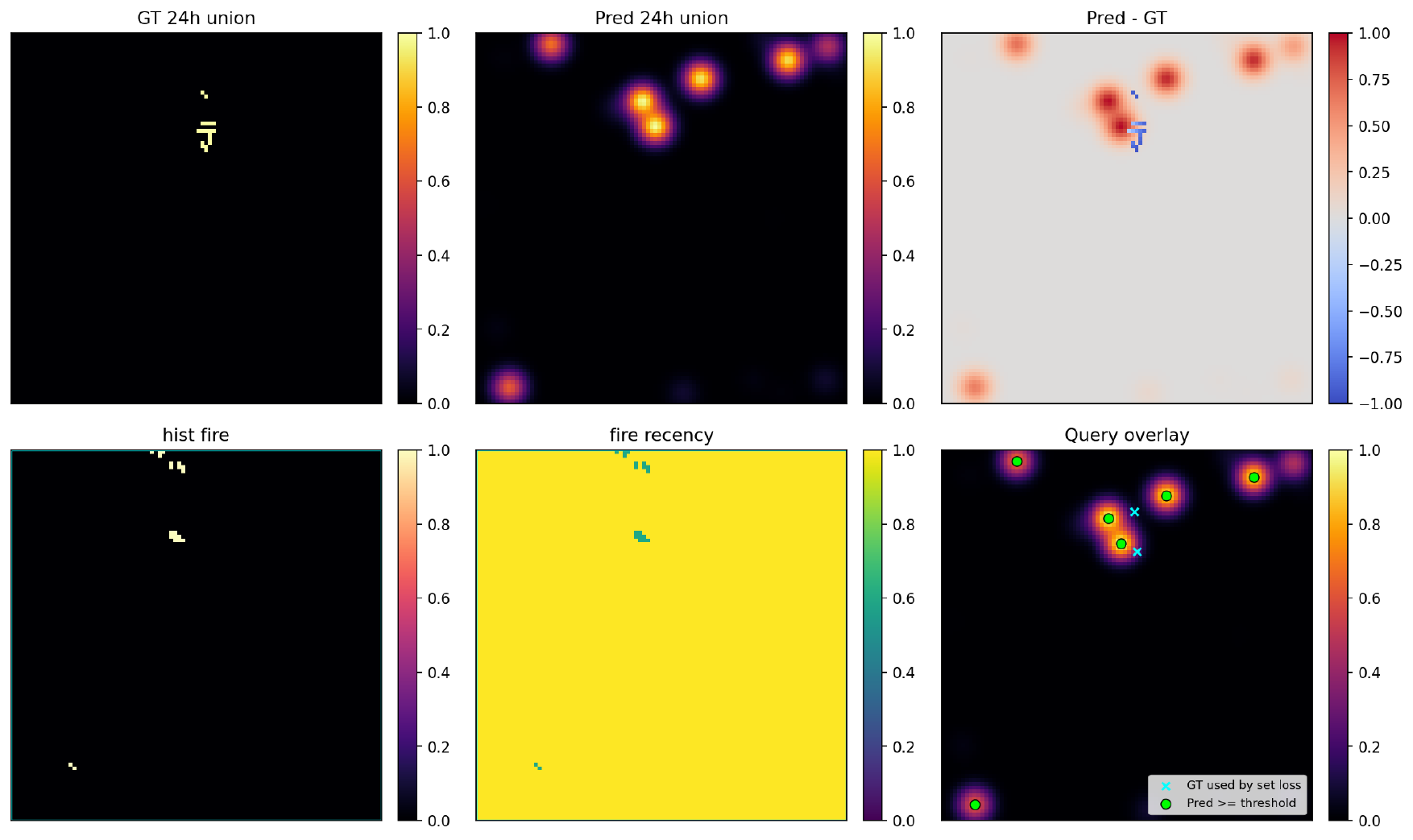} \\[-0.6em]

    \makebox[1.3em][c]{\raisebox{1.35\height}{\rotatebox{90}{\scriptsize\bfseries Median}}} &
    \includegraphics[width=0.305\textwidth]{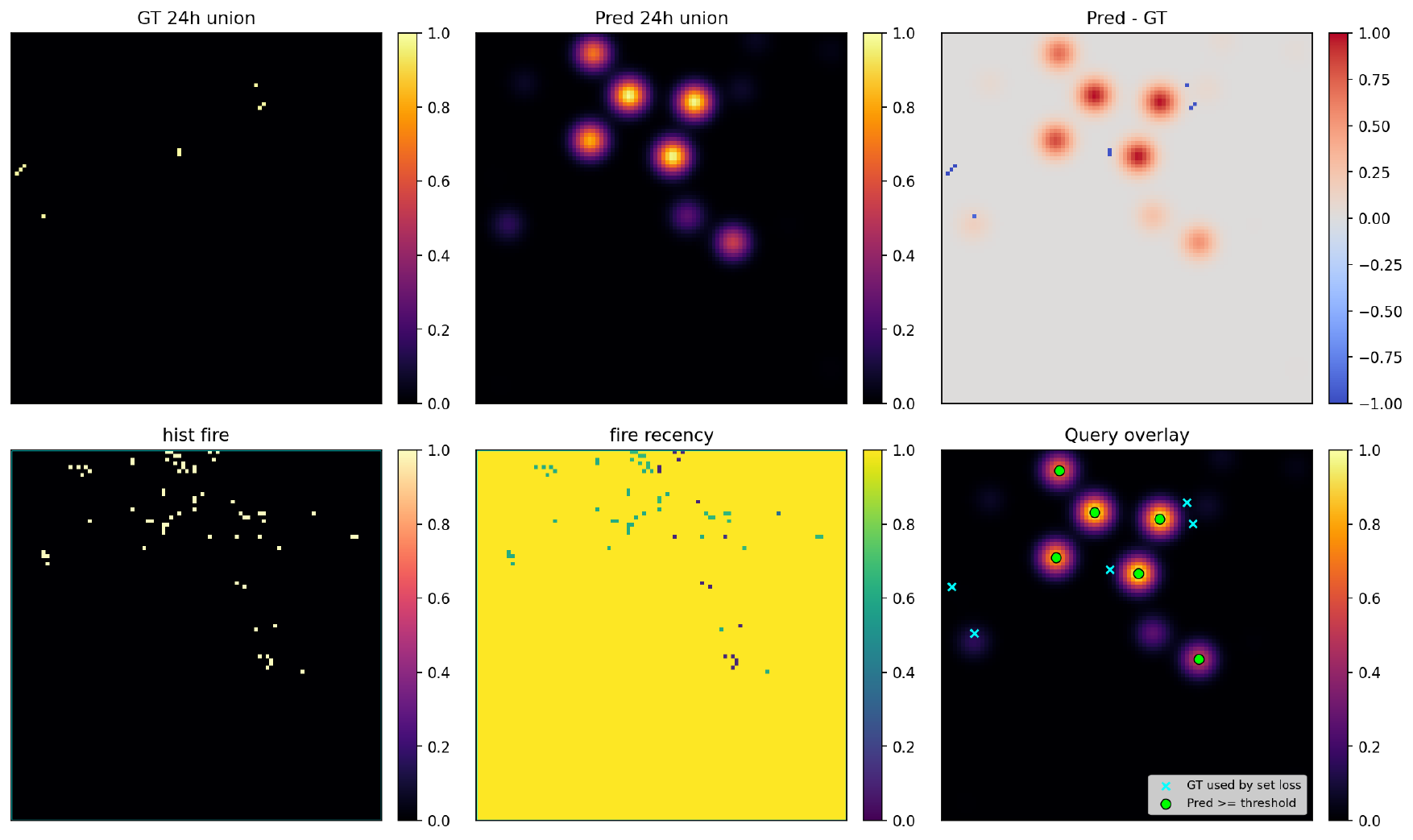} &
    \includegraphics[width=0.305\textwidth]{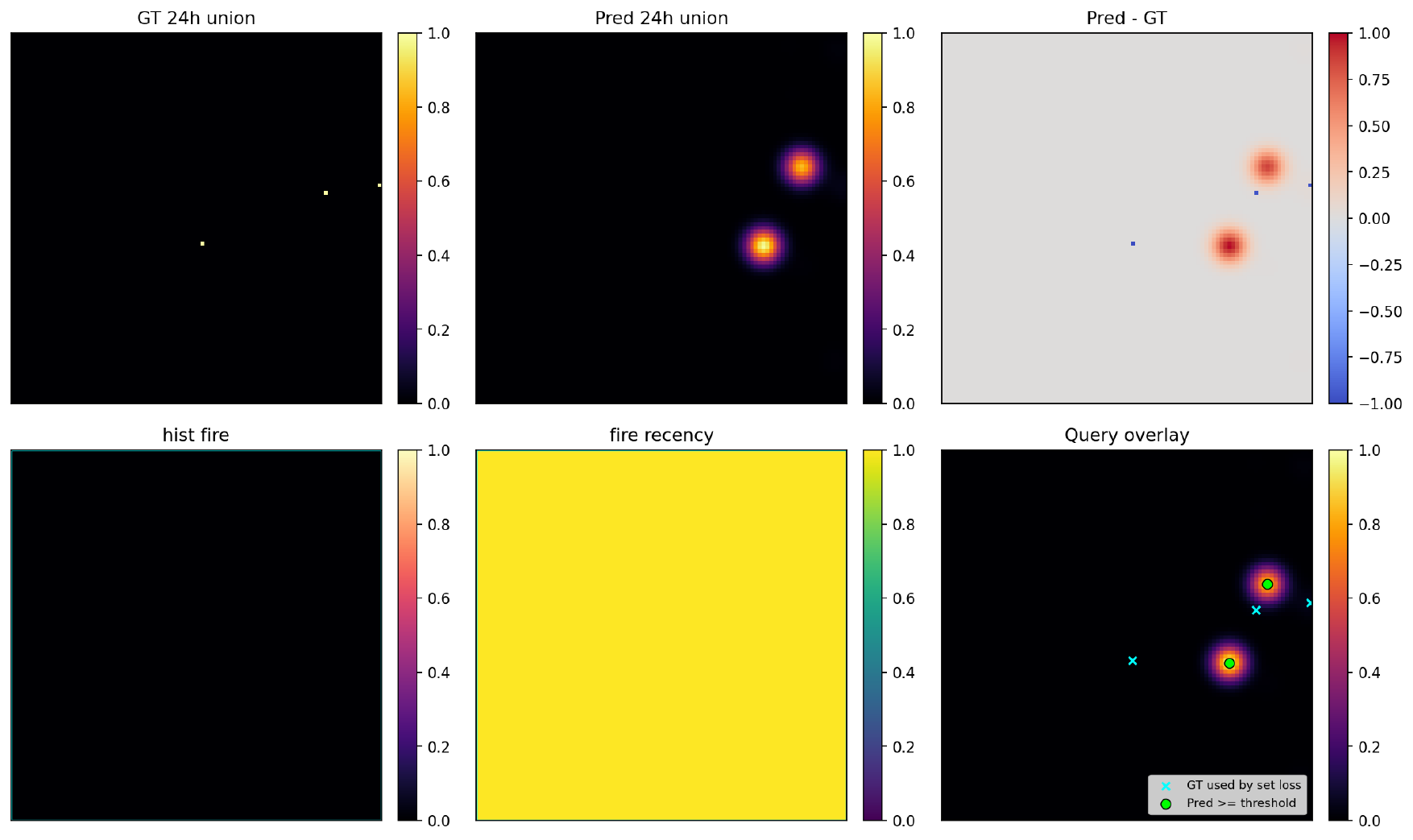} &
    \includegraphics[width=0.305\textwidth]{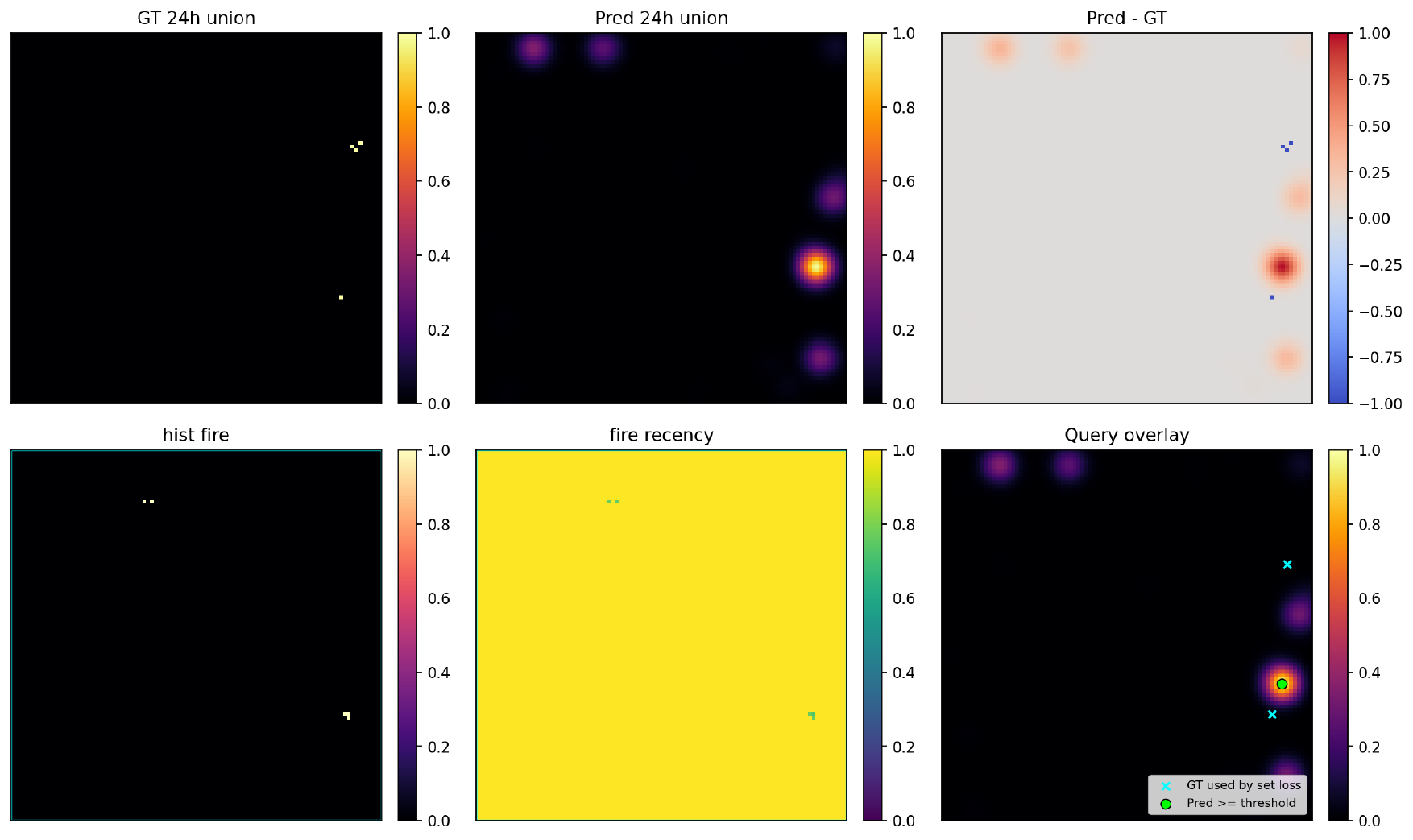} \\[-0.6em]

    \makebox[1.3em][c]{\raisebox{1.35\height}{\rotatebox{90}{\scriptsize\bfseries Failure}}} &
    \includegraphics[width=0.305\textwidth]{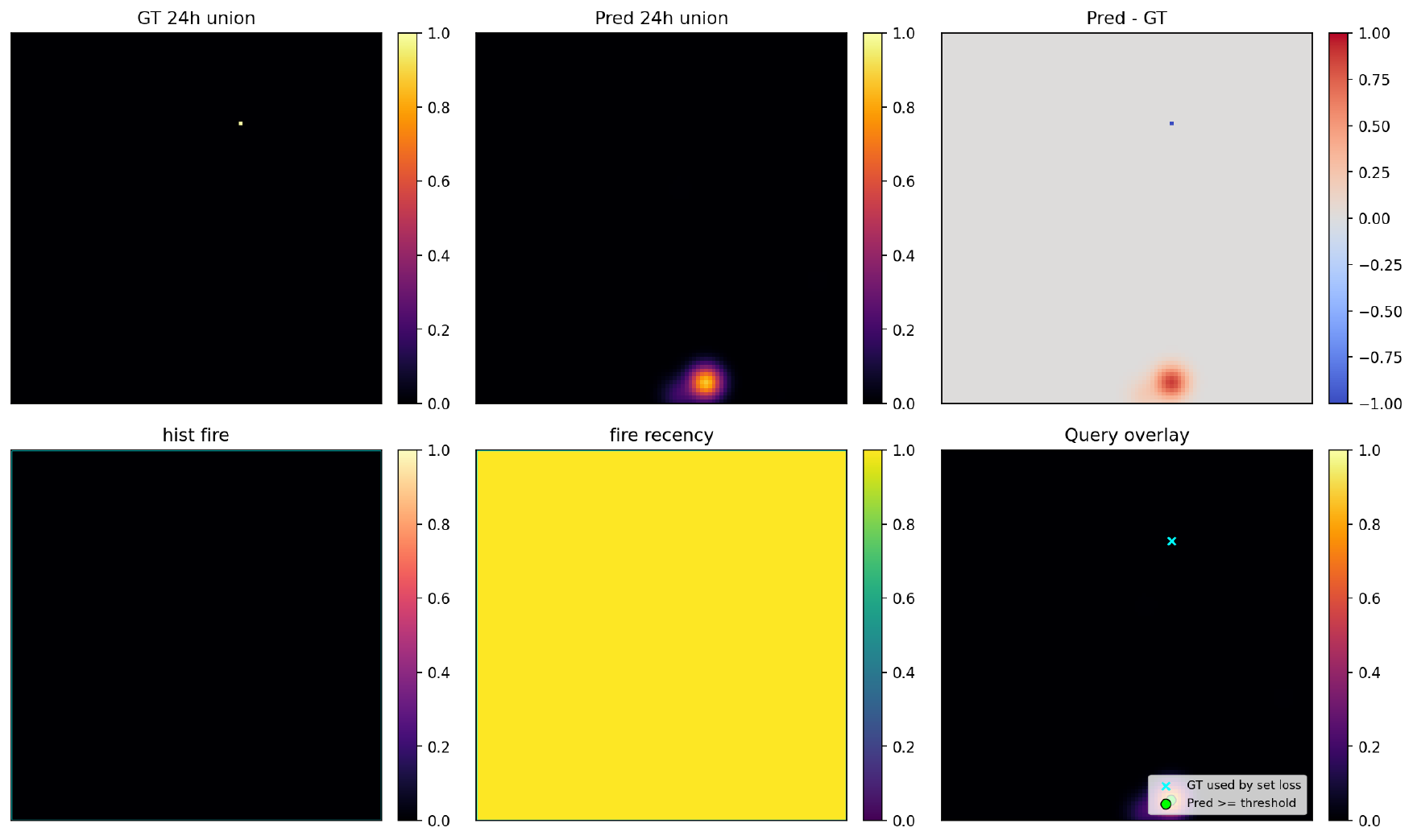} &
    \includegraphics[width=0.305\textwidth]{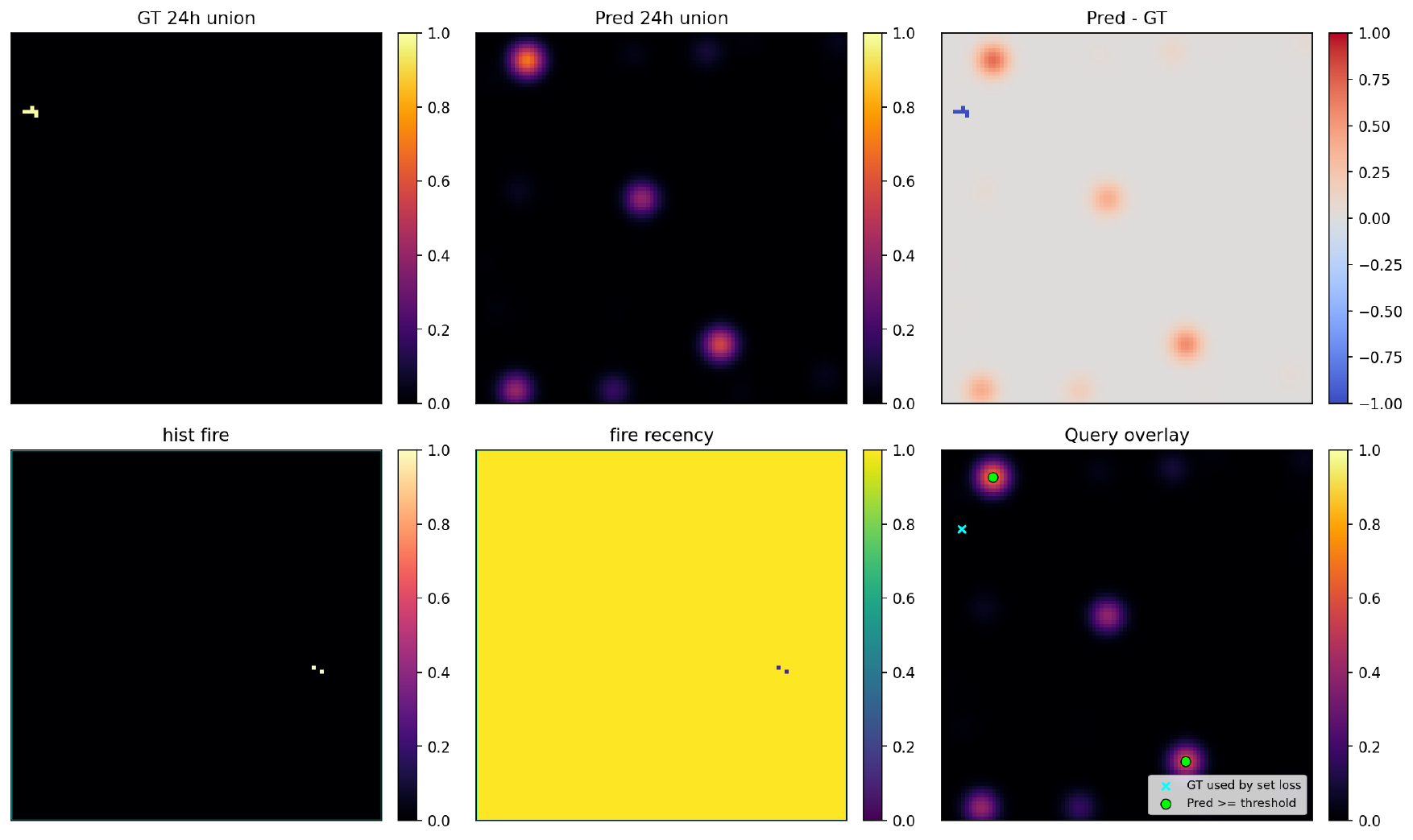} &
    \includegraphics[width=0.305\textwidth]{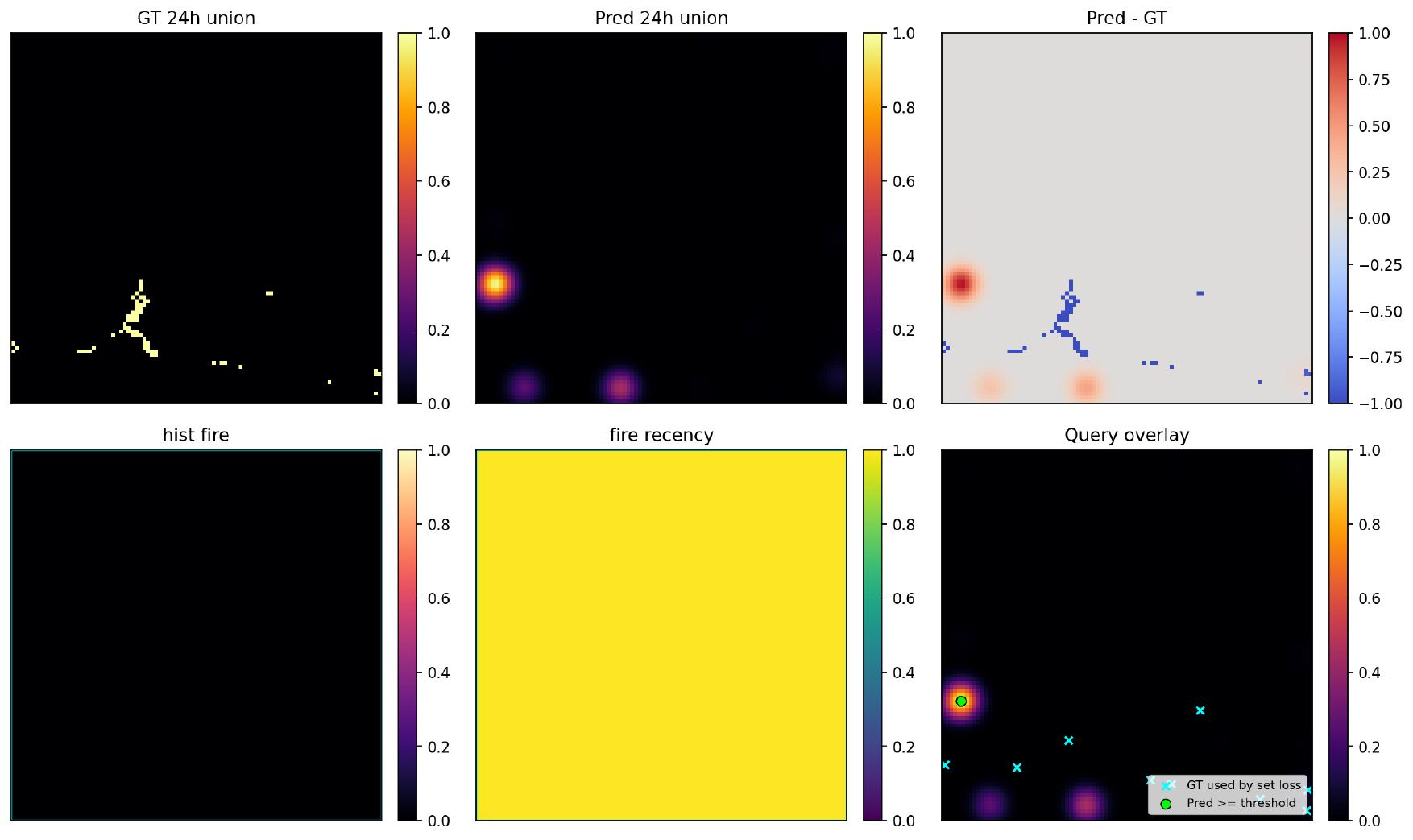}
\end{tabular}
\captionof{figure}{Additional qualitative gallery for WISP variant v5.}
\label{fig:app_qualitative_v5}
\end{center}

\begin{center}
\setlength{\tabcolsep}{1pt}
\renewcommand{\arraystretch}{0.25}
\begin{tabular}{c c c c}
    \makebox[1.3em][c]{\raisebox{1.35\height}{\rotatebox{90}{\scriptsize\bfseries Good}}} &
    \includegraphics[width=0.305\textwidth]{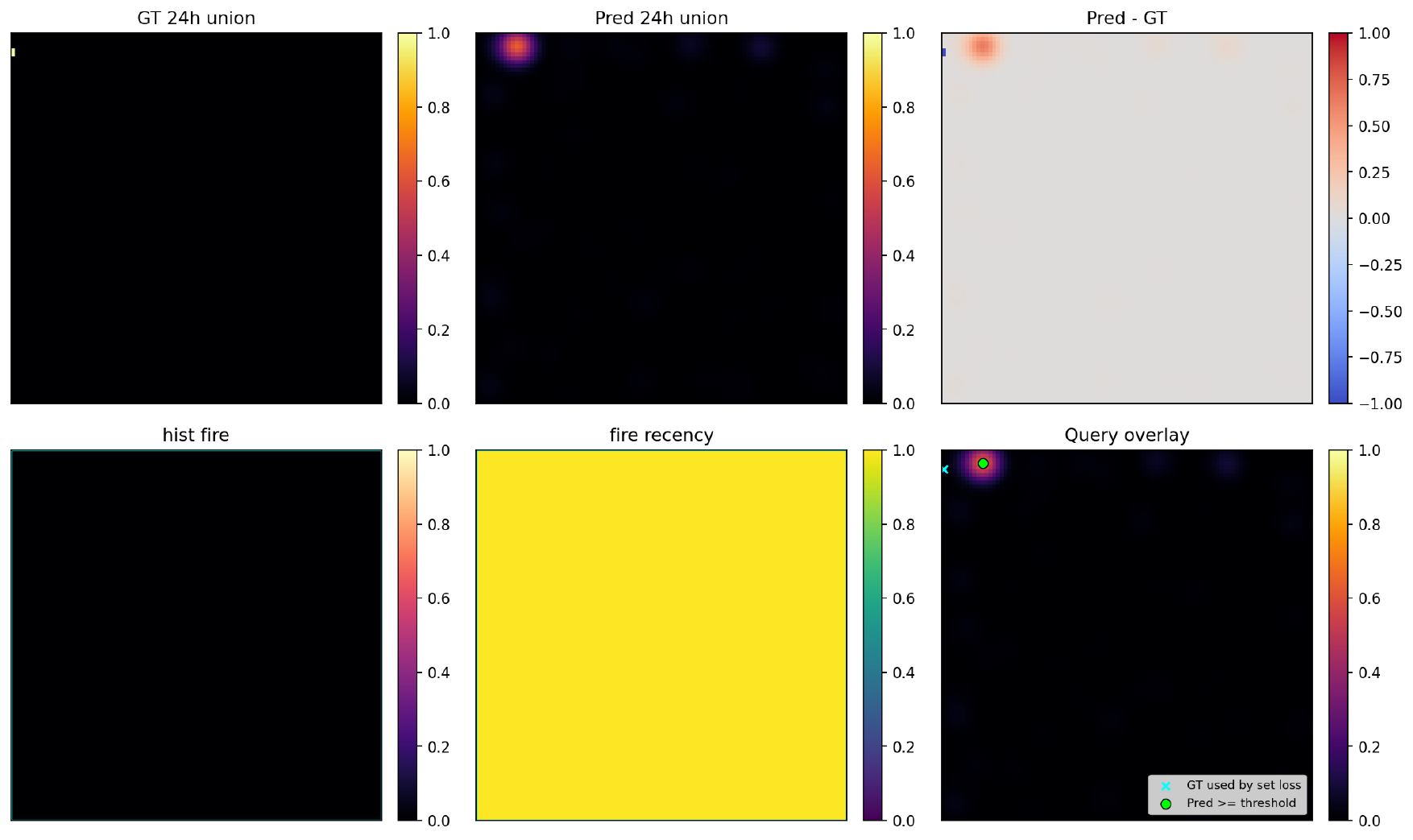} &
    \includegraphics[width=0.305\textwidth]{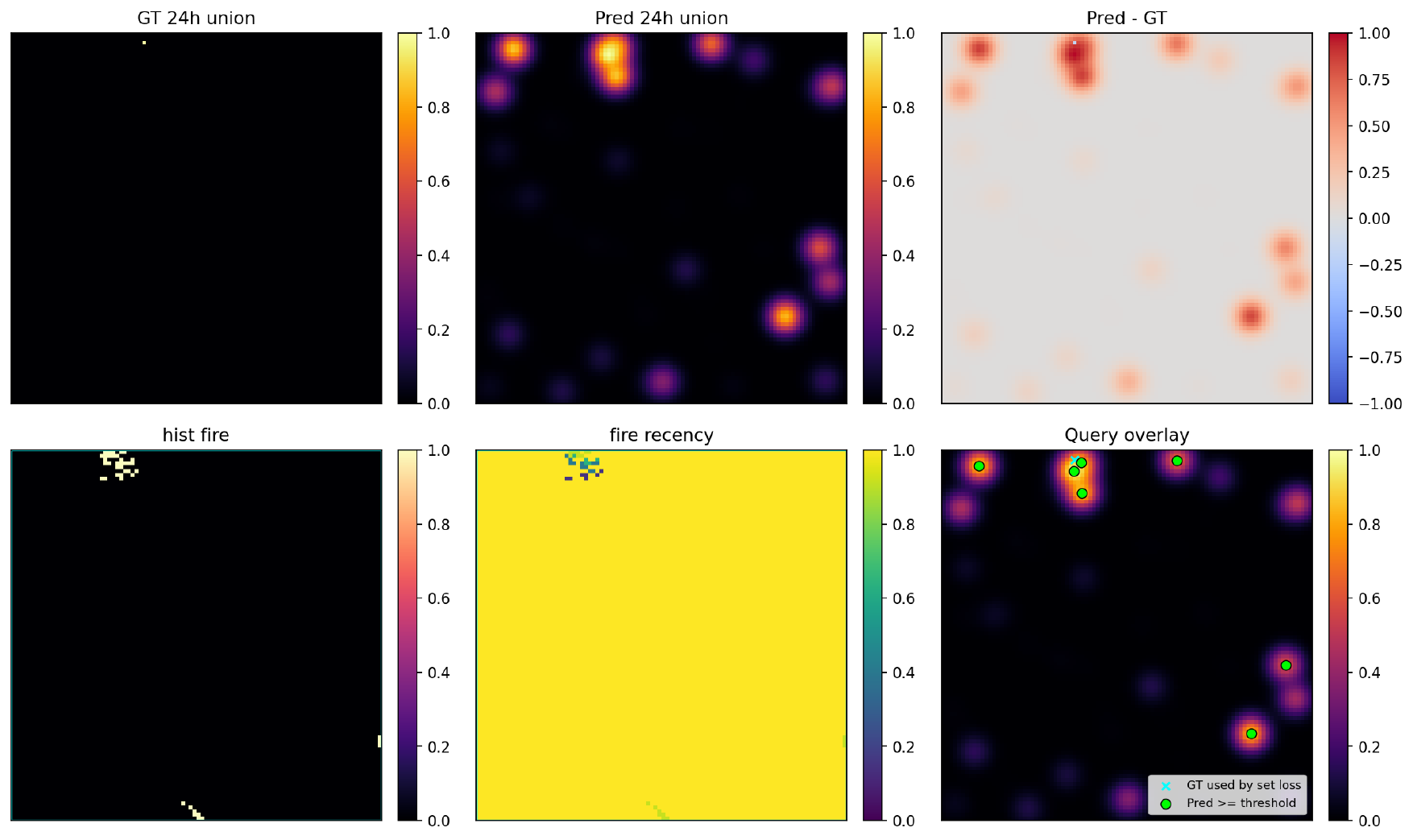} &
    \includegraphics[width=0.305\textwidth]{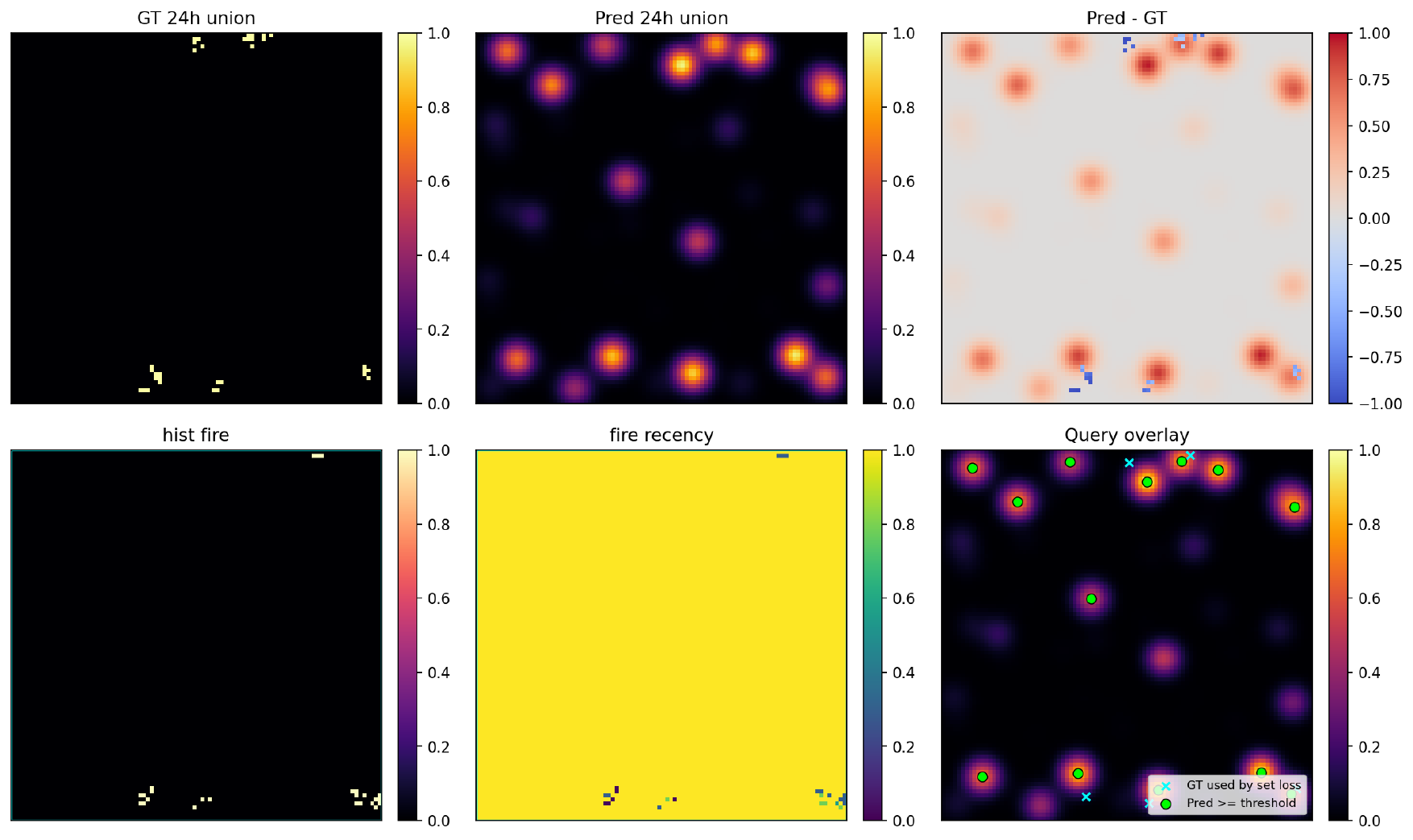} \\[-0.6em]

    \makebox[1.3em][c]{\raisebox{1.35\height}{\rotatebox{90}{\scriptsize\bfseries Median}}} &
    \includegraphics[width=0.305\textwidth]{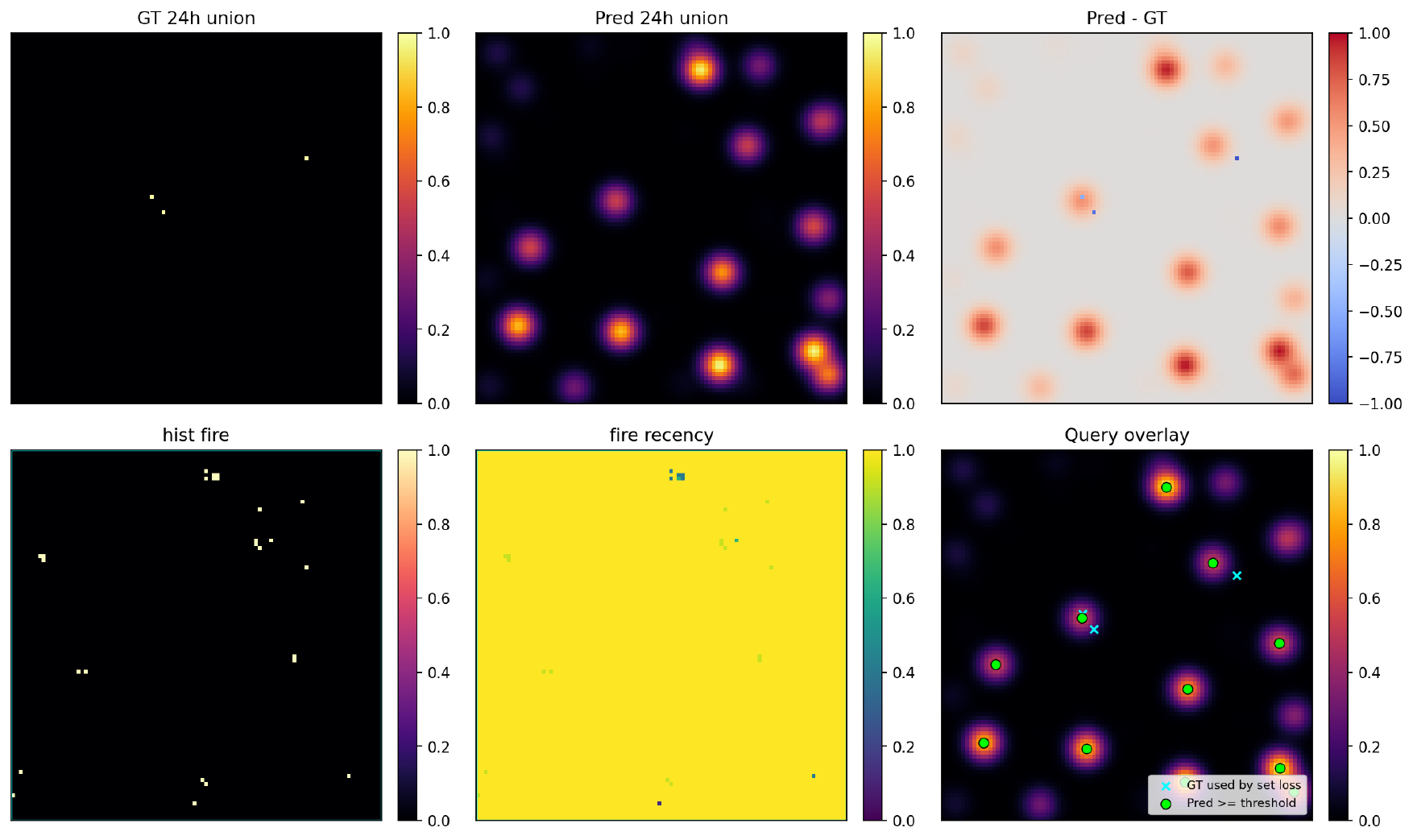} &
    \includegraphics[width=0.305\textwidth]{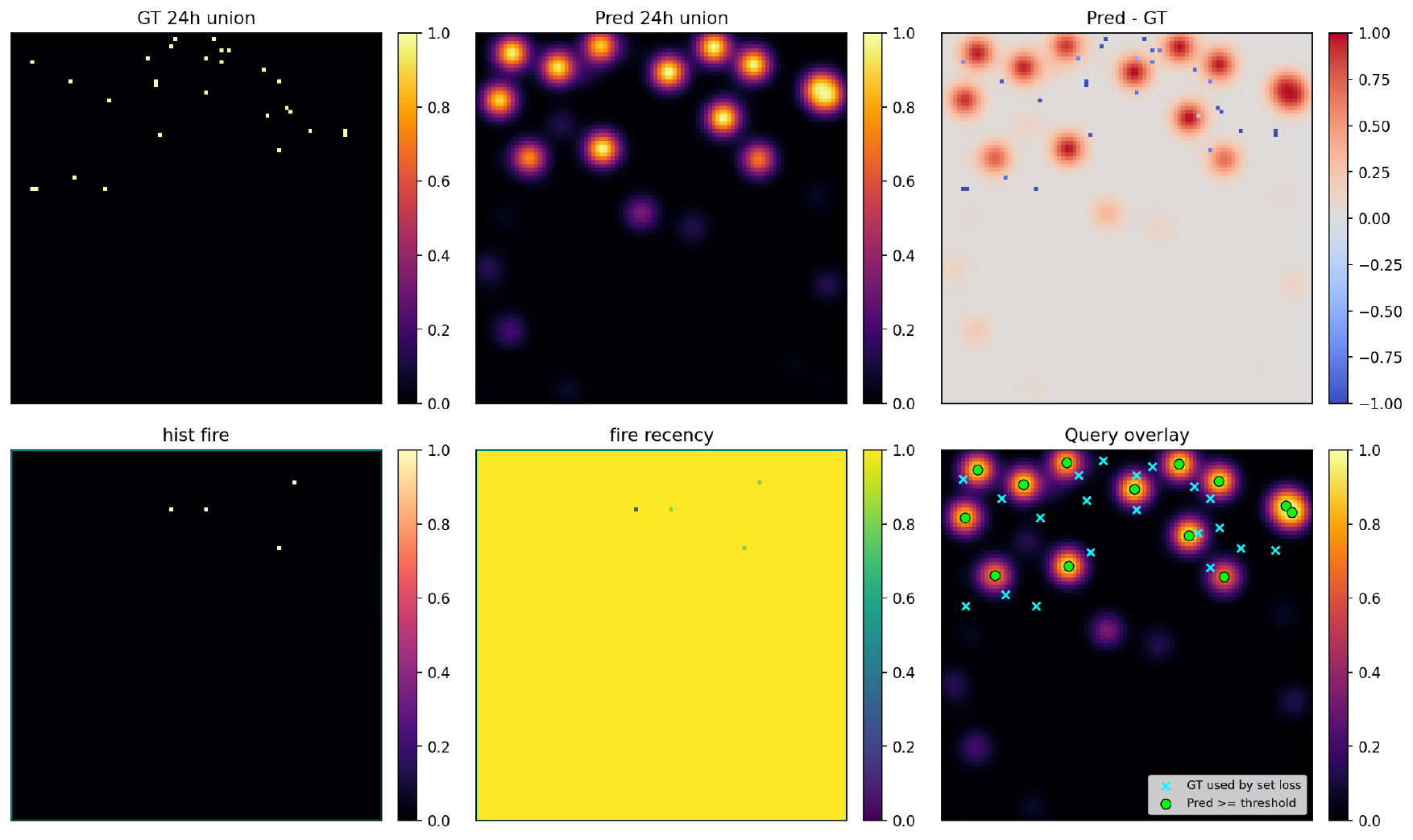} &
    \includegraphics[width=0.305\textwidth]{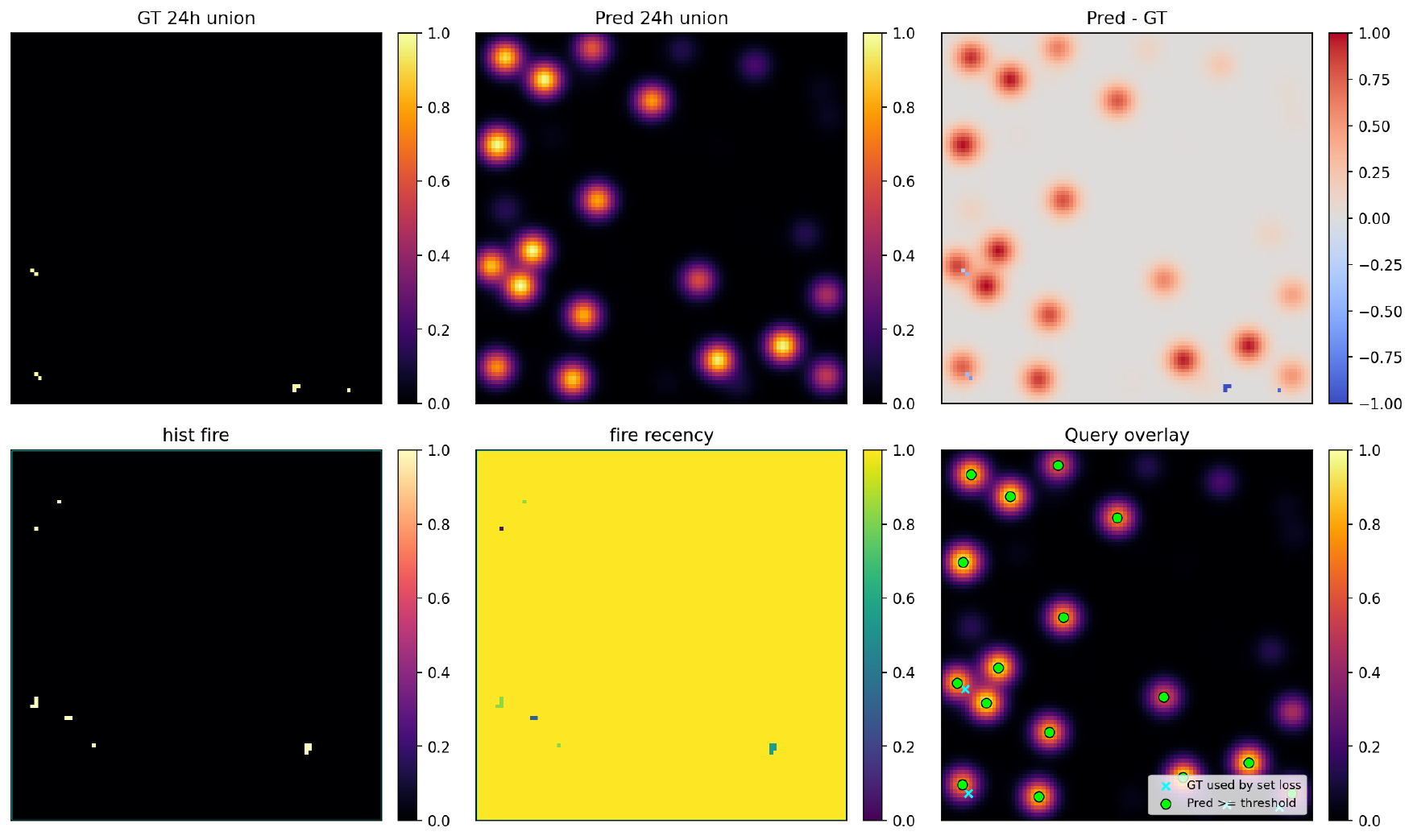} \\[-0.6em]

    \makebox[1.3em][c]{\raisebox{1.35\height}{\rotatebox{90}{\scriptsize\bfseries Failure}}} &
    \includegraphics[width=0.305\textwidth]{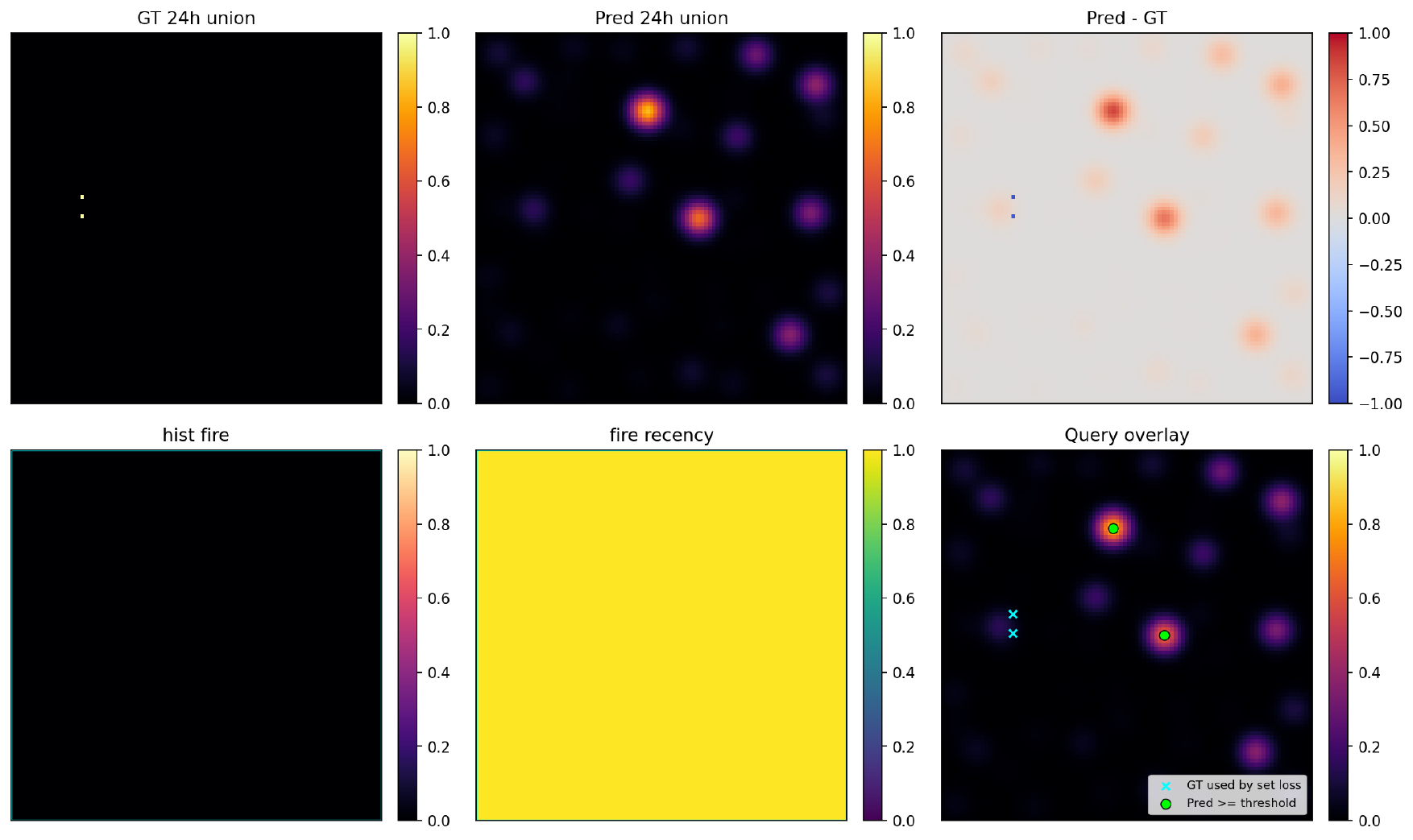} &
    \includegraphics[width=0.305\textwidth]{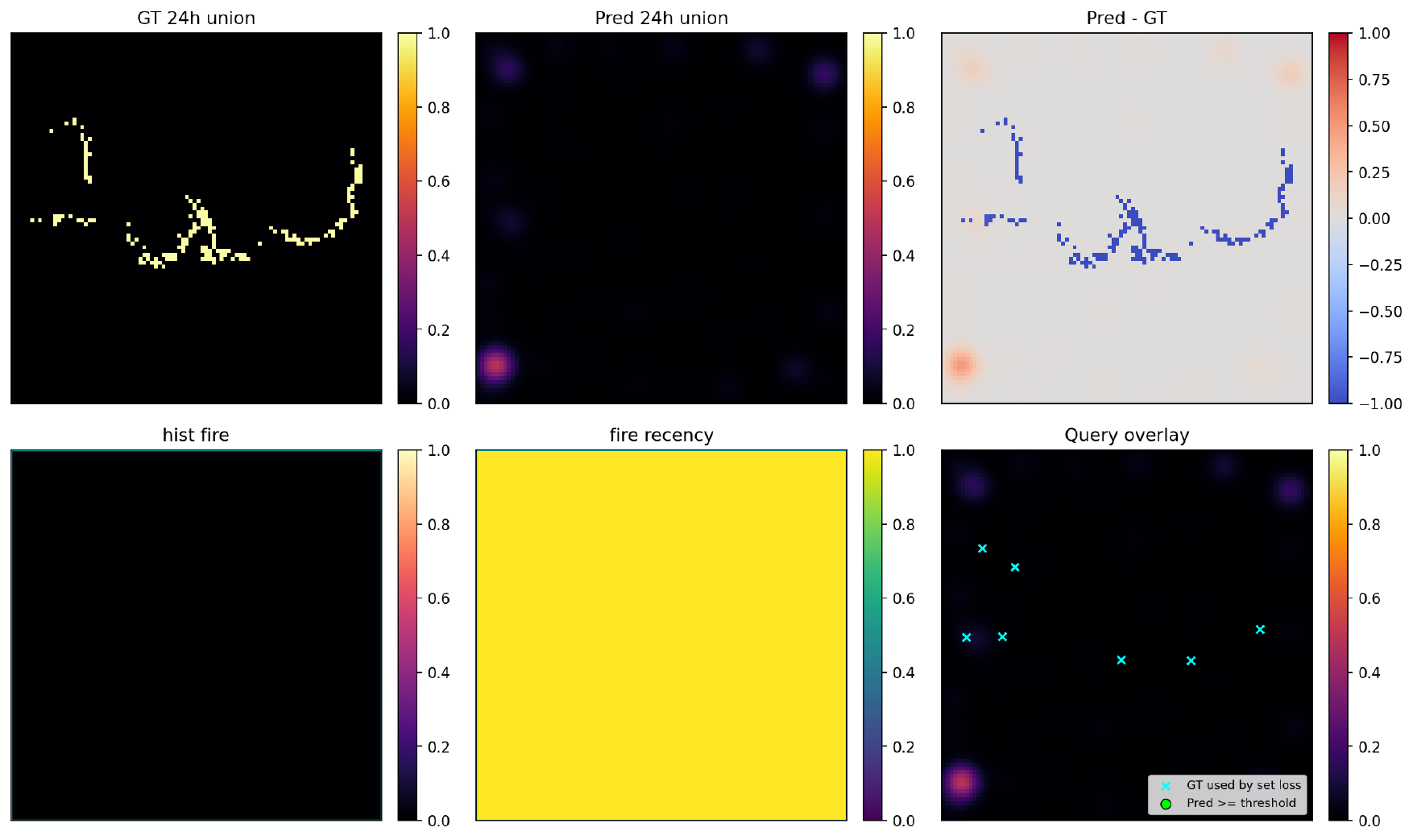} &
    \includegraphics[width=0.305\textwidth]{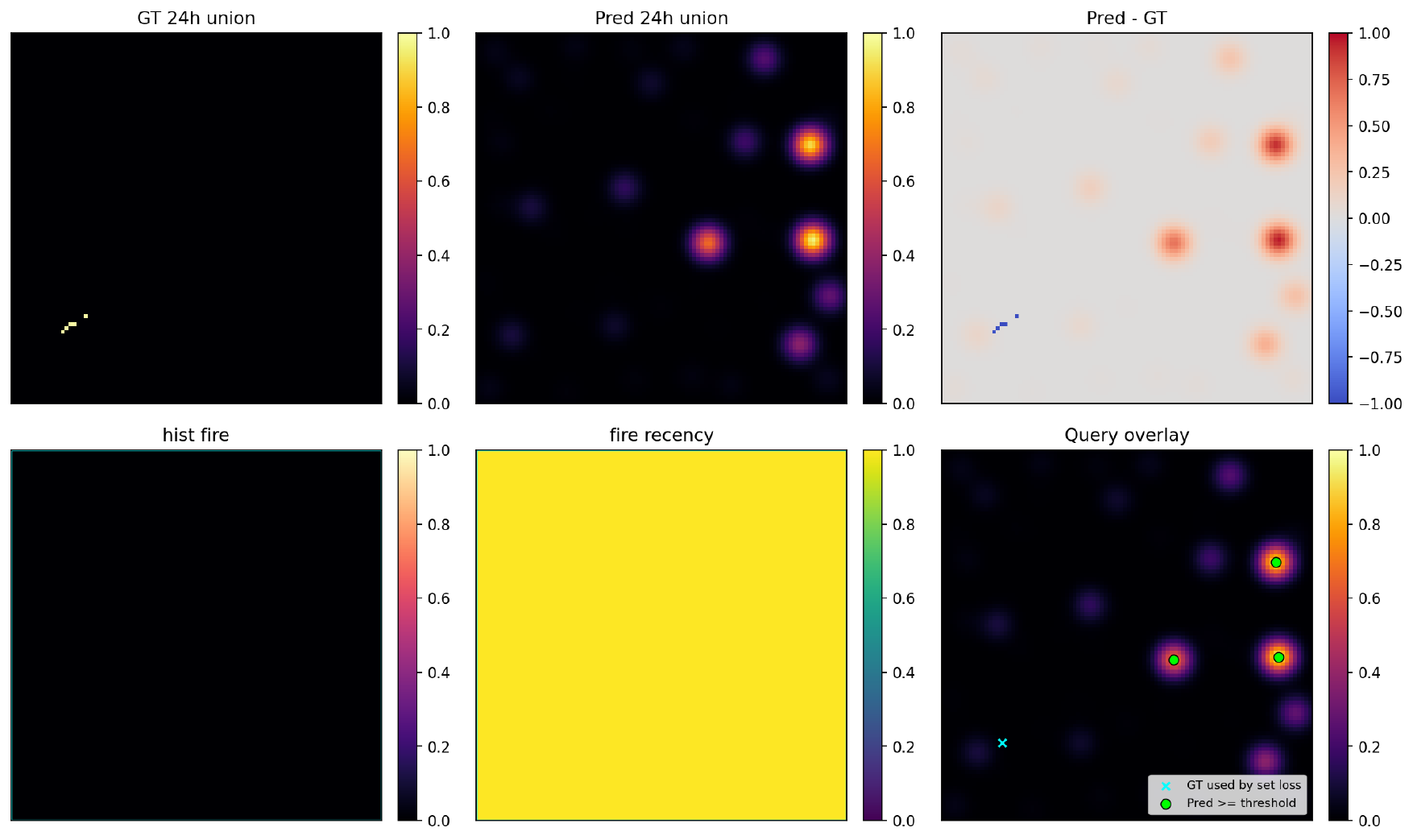}
\end{tabular}
\captionof{figure}{Additional qualitative gallery for WISP variant v6.}
\label{fig:app_qualitative_v6}
\end{center}

\clearpage


\end{document}